%% file: main.tex
\documentclass[letterpaper,journal]{IEEEtran}
\usepackage{graphicx}
\usepackage{float}
\usepackage{amsmath}
\usepackage{threeparttable, tablefootnote}
 
\let\vec\boldvec

\usepackage{epstopdf}
\usepackage{multicol}
\usepackage{booktabs}
\usepackage[hidelinks]{hyperref}
\usepackage{amssymb}
\usepackage{bm}
\usepackage{tabularx}
\usepackage{arydshln}
\usepackage[utf8]{inputenc}
\usepackage{blindtext}
\usepackage{subfig}
\usepackage{wrapfig}
\usepackage{dblfloatfix}
\usepackage{siunitx}
\usepackage{todonotes} 
\usepackage{cases}
\usepackage{nicefrac}
\makeatletter
\def\BState{\State\hskip-\ALG@thistlm}
\makeatother
\DeclareMathAlphabet{\mathcal}{OMS}{cmsy}{m}{n}
\usepackage{hyperref}
\usepackage{fancyhdr}
\usepackage{tikz}
\usepackage{pgfplots}
\pgfplotsset{compat=newest}
\usepgfplotslibrary{units}
\newlength{\figwidth}          
\newlength{\figheight}         
\usetikzlibrary{external} 
\tikzset{external/force  remake=false} 
\tikzsetexternalprefix{figures/externalized_figs/} 
\makeatletter
\renewcommand{\todo}[2][]{\tikzexternaldisable\@todo[#1]{#2}\tikzexternalenable}
\makeatother

\newcommand{\eq}[1]{Equation~(\ref{#1})}
\newcommand{\fig}[1]{Figure~\ref{#1}}

\newcommand{\sect}[1]{Section~\ref{#1}}
\newcommand{\tab}[1]{Table~\ref{#1}}
\newcommand{\alg}[1]{Algorithm~\ref{#1}}

\newcommand{\E}[1]{\mathbb{E}_{#1}}
\newcommand{\pdes}[1]{\vec{p}^\text{des}_{#1}}
\newcommand{\p}[1]{\vec{p}_{#1}}
\newcommand{\q}[1]{\vec{q}_{#1}}
\newcommand{\qvel}[1]{\dot{\vec{q}}_{#1}}
\newcommand{\R}[1]{\mathbb{R}^{#1}}
\newcommand{\sballi}[0]{\vec{s}^\text{b}_0}
\newcommand{\sball}[1]{\vec{s}^\text{b}_{#1}}
\newcommand{\sballrec}[1]{\vec{s}^\text{rec}_{#1}}
\newcommand{\sballsim}[1]{\vec{s}^\text{sim}_{#1}}
\newcommand{\srob}[1]{\vec{s}^\text{r}_{#1}}
\newcommand{\rpos}[1]{\vec{x}^\text{r}_{#1}}
\newcommand{\rvel}[1]{\dot{\vec{x}}^\text{r}_{#1}}
\newcommand{\s}[1]{\vec{s}_{#1}}
\newcommand{\ob}[1]{\vec{o}_{#1}}
\newcommand{\ac}[1]{\vec{a}_{#1}}
\newcommand{\bdes}[1]{\vec{b}^\text{des}_{#1}}
\newcommand{\bpos}[1]{\vec{b}_{#1}}
\newcommand{\bhit}[1]{\vec{b}^\text{h}_{#1}}
\newcommand{\bland}[1]{\vec{b}^\text{land}_{#1}}
\newcommand{\bvel}[1]{\dot{\vec{b}}_{#1}}
\newcommand{\trajrecargone}[1]{\vec{\tau}^\text{rec}_{#1}} 
\newcommand{\trajrec}[2]{\vec{\tau}^\text{rec,$#1$}_{#2}} 
\newcommand{\trajb}[1]{\vec{\tau}^\text{b}_{#1}} 
\newcommand{\trajr}[1]{\vec{\tau}^\text{r}_{#1}} 
\newcommand{\thit}[0]{t_\text{h}}
\newcommand{\tland}[0]{t_\text{land}}
\newcommand{\tnin}[0]{t_{\text{n}_\text{in}}}
\newcommand{\tnout}[0]{t_{\text{n}_\text{out}}}
\newcommand{\bnin}[1]{\vec{b}^{\text{n}_\text{in}}_{#1}}
\newcommand{\bnout}[1]{\vec{b}^{\text{n}_\text{out}}_{#1}}
\newcommand{\rhit}[1]{r^\text{hit}_{#1}}
\newcommand{\rtt}[1]{r^\text{tt}_{#1}}
\newcommand{\Nrec}[0]{N^\text{rec}}
\newcommand{\Drec}[1]{\mathcal{D}^\text{rec}_{#1}}
\usepackage{cite}
\ifCLASSINFOpdf
  \else
\fi
\usepackage{algorithmic}
\usepackage{algorithm}
\usepackage{url}
\begin{document}
\title{Learning to Play Table Tennis From Scratch\\ using Muscular Robots}
\author{Dieter~Büchler, 
        Simon~Guist, 
        Roberto~Calandra, 
        Vincent~Berenz, 
        Bernhard Schölkopf 
        and~Jan~Peters
\thanks{Dieter Büchler, Simon Guist, Vincent Berenz, Bernhard Schölkopf and Jan Peters are with the Department of Empirical Inference, MPI for Intelligent Systems, Tübingen~(Germany)~(dbuechler, sguist, vberenz, bs@tue.mpg.de, mail@jan-peters.net). 
Jan Peters is also with the Department of Computer Science, Technische Universität Darmstadt~(Germany).
Roberto Calandra is with Facebook AI Research~(USA)~(rcalandra@fb.com)}
}
\maketitle
\begin{abstract}
\input{0_abstract.tex}
\end{abstract}
\IEEEpeerreviewmaketitle

\thispagestyle{fancy}%
\section{Introduction}
\label{sec:intro}
\input{1_intro}


\section{Training of Muscular Robot Table Tennis}
\label{sec:methods}
\input{2_methods}

\section{Experiments and Evaluations}
\label{sec:experiment}
\input{3_experiments}

\section{Conclusion}
\label{sec:conclusion}
\input{4_conclusion}

\ifCLASSOPTIONcaptionsoff
  \newpage
\fi
\bibliographystyle{IEEEtran}
\bibliography{refs}
\newpage
\appendices
\label{sec:appendix}
\input{6_appendices}
\end{document}

%% file: 0_abstract.tex
Dynamic tasks like table tennis are relatively easy to learn for humans but pose significant challenges to robots.  
Such tasks require accurate control of fast movements and precise timing in the presence of imprecise state estimation of the flying ball and the robot.
Reinforcement Learning (RL) has shown promise in learning of complex control tasks from data.
However, applying step-based RL to dynamic tasks on real systems is safety-critical as RL requires exploring and failing safely for millions of time steps in high-speed and high-acceleration regimes.
In this paper, we demonstrate that safe learning of table tennis using model-free Reinforcement Learning can be achieved by using robot arms driven by pneumatic artificial muscles (PAMs).
Softness and back-drivability properties of PAMs prevent the system from leaving the safe region of its state space. 
In this manner, RL empowers the robot to return and \textit{smash} real balls with \SI{5}{\meter\per\second} and \SI{12}{\meter\per\second} on average respectively to a desired landing point.
Our setup allows the agent to learn this safety-critical task 
(i) without safety constraints in the algorithm,  
(ii) while maximizing the speed of returned balls directly in the reward function 
(iii) using a stochastic policy that acts directly on the low-level controls of the real system and  
(iv) trains for thousands of trials
(v) \textit{from scratch} without any prior knowledge.
Additionally, we present HYSR, a practical hybrid sim and real training procedure that avoids playing real balls during training by randomly replaying recorded ball trajectories in simulation and applying actions to the real robot.
To the best of our knowledge, this work pioneers (a) fail-safe learning of a safety-critical dynamic task using anthropomorphic robot arms, (b) learning a precision-demanding problem with a PAM-driven system that is inherently hard to control as well as (c) train robots to play table tennis without real balls.
Videos and datasets of the experiments can be found on \texttt{\url{muscularTT.embodied.ml}}.

%% file: 1_intro.tex
\IEEEPARstart{R}{einforcement} Learning~(RL) solves challenging tasks such as the complex game of Go~\cite{silver_mastering_2017-1}, full body locomotion tasks in simulation~\cite{heess_emergence_2017} or difficult continuous control tasks such as robot manipulation~\cite{gu_deep_2016,openai_learning_2018} to name just a few.
All of these challenging tasks are either realized in simulated environments or the real robot task is often engineered in a way to allow safe exploration: 
\begin{figure}[t]
    \centering\scriptsize%
    \includegraphics[width=0.98\linewidth]{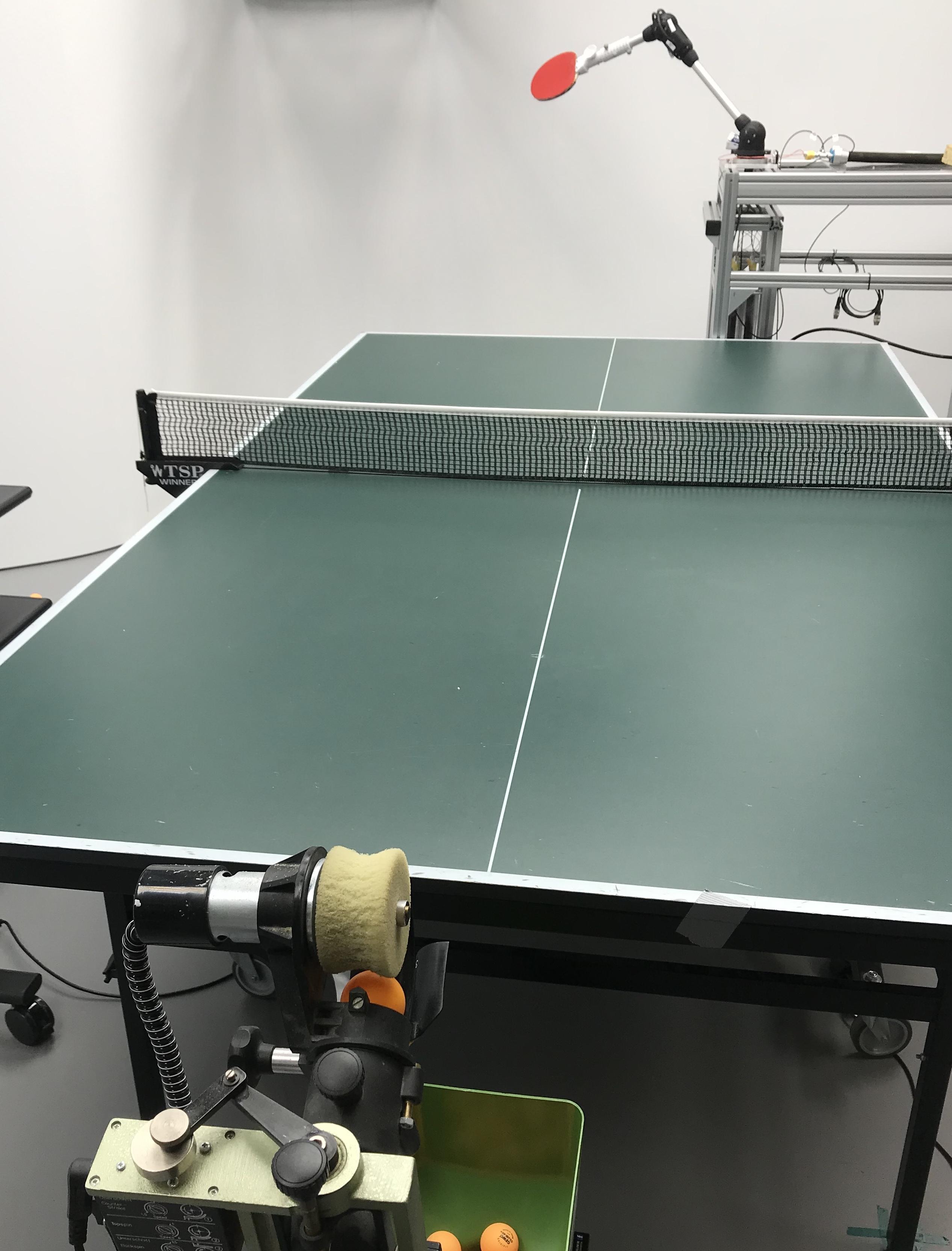}
    \caption{
        Our setup consists of a four DoF robot arm actuated by pneumatic muscles and a ballgun that launches balls towards the robot arm. 
        The robot successfully learns to return the ball to a desired landing point on the table. 
        The Cartesian positions of the ball are measured by a color-based camera detection system.
    }
    \vspace{-10pt}
    \label{fig:setup}
\end{figure}
In simulation, the agent is allowed to bump into objects, collide with itself, and the actions can act on low-level controls~(such as torques or activations to muscles) while being sampled from a stochastic policy.
Also, it is possible to employ step-based policies that permit the agent to react in every time step to changes in the state.
These changes might produce noisy action sequences as they can vary significantly from one time step to the next~(in contrast to episode-based RL where actions are executed in an open-loop fashion and are, hence, usually smooth). 
All of these points allow the agent to freely explore and learn from failures while interacting for millions of time steps with the environment.

To allow for similar behavior in real world applications of RL, current approaches often constrain robots to slow motion skills such as required for manipulation.
For such tasks, simple engineered checks assure robot safety.
These checks detect, for instance, collisions with objects~(usually based on heuristics) and the robot itself. 
Safety may be further ensured by limiting joint accelerations and velocities, stopping the robot before it reaches its joint limits, and filtering noisy actions generated by stochastic policies~\cite{schwab_simultaneously_2019,gu_deep_2016}. 
Also, multiple recent papers point out that robot safety is essential if RL should be reliably employed on real robots~\cite{bodnar_quantile_2019,majumdar_risk-sensitive_2017,dulac-arnold_challenges_2019}.

Cautious checks are sufficient for slow movements, e.g., in grasping or lifting objects.
However, in tasks such as table tennis, they are not applicable because it is essential to exert explosive motions. 
In a situation where an incoming ball is supposed to be hit at a point in space far away from the current racket position, the agent needs to rapidly accelerate the racket to gain momentum and reach the hitting point in time. 
Thus, the amount of accelerations that the robot can generate is proportional to the upper bound of the dexterity it can reach in dynamic tasks.
Cautious checks for such problems are disadvantageous because
1) empirically finding parameters for the safety heuristics is substantially harder at faster motions and
2) the safety checks heavily limit the capabilities of robots as they are usually conservatively chosen to avoid damage to the real system reliably.

It is challenging not to limit the performance of dynamic real robotic tasks too much by safety checks while letting the agent freely explore fast motions.
For this reason, learning approaches to robot table tennis using anthropomorphic human arm sized systems usually employ techniques such as imitation learning~\cite{gomez-gonzalez_adaptation_2018}, choosing or optimizing from safe demonstrations~\cite{muelling_learning_2010,mulling_learning_2013,huang_jointly_2016}, minimizing acceleration for optimized trajectories~\cite{koc_online_2018}, distributing torques over all degrees of freedom~(DoF)~\cite{kober_movement_2010} as well as cautious learning control approaches~\cite{koc_optimizing_2019}.

In~\cite{buchler_lightweight_2016,buchler_learning_2019-1}, it has been shown that systems actuated by pneumatic artificial muscles~(PAM) are suitable to execute explosive hitting motions safely. 
By adjusting the pressure range for each PAM, a fast motion generated by the high forces of the PAMs can be decelerated before a DoF exceeds its allowed joint angle range. 
Besides, PAM-driven robots are inherently backdrivable, which makes them exceptionally robust.
However, such actuators are substantially harder to control than traditional motor-driven systems~\cite{tondu_modelling_2012,buchler_control_2018}.
For this reason, the predominant use of PAM-driven systems is to slowly handle heavy objects, for disturbance rejection, or as a testbed for control approaches.

In this paper, we show that soft actuation of PAM-driven systems enables RL to be safely applicable to a dynamic task directly on real hardware and, thereby, enable RL to overcome the difficulties of a dynamic task as well as the control issues of soft robots.
By enabling RL to explore fast motions directly on the real system, we can leverage the benefits of such complex systems for dynamic tasks~(such as high power-to-weight ratio, and storage of energy and release) despite their control issues.
In particular, we show that, by using robots actuated by pneumatic artificial muscles~(PAM) from \cite{buchler_lightweight_2016,buchler_learning_2019-1}, the robot learns to return and even smash real table tennis balls using RL \textit{from scratch}.
Rather than avoiding fast motions, we leverage the inherent safety of the robot to favor highly accelerated strikes by maximizing the velocity of the returned ball directly in the reward function. 
Also, we 
1) apply noisy actions sampled from a Gaussian multi-layer perceptron~(MLP) policy directly on the low level controls~(desired pressures),
2) while running the RL algorithm for millions of time steps on the real hardware,
3) randomly initialize the policy at start and
4) introduce a hybrid sim and real training procedure to circumvent practical issues of long duration training such as collecting, removing and supplying table tennis balls from and to the system.
Using this training procedure, the agent learns to return and smash balls without touching any real ball during training.

It is worth mentioning that with our setup, it is possible to learn robot table tennis while adding as little priors on the solution as possible.
Neither do we have to add any constraints or regularizers, such as minimal accelerations or energy, nor do we have to use a higher abstraction level like task or joint space but instead learn directly on the low-level controls. 
Also, we avoid potentially suboptimal models or demonstrations.
Models, constraints, and regularizers steer the optimization of the policy to a solution with possibly degenerated performance.
Therefore, it is desirable to specify only task-related entities in the reward function and avoid such priors. 
In this manner, the solution emerges purely from the hardware and the desired behavior defined in the reward function. 

To the best of our knowledge, this work is novel in many aspects: 
First, it demonstrates the first results in learning the challenging task of playing table tennis with muscular robots.
Second, this paper pioneers learning smash hitting motions on real robots rather than only returning balls to the opponent's side of the table. 
Third, this work pioneers the safe application of model-free RL on safety-critical dynamic tasks. 
Fourth, we present the first approach to robot table tennis, where the training does not involve real balls.

The remaining of the paper is divide as follows:
In \sect{sec:methods}, we introduce the task and reward functions used to learn to return and smash as well as the hybrid sim and real training procedure~(HYSR). 
The return and smash experiment are described in detail in \sect{sec:experiment}.
We summarize the contributions and discuss the results in \sect{sec:conclusion}.

%% file: 2_methods.tex
Learning dynamic tasks using muscular robots allows the RL algorithm to run on real systems similarly to when applied to simulated robots.
In return, RL helps to overcome the inherent control difficulties PAM-driven systems and leverage its beneficial properties such as the powerful actuation to learn to return and smash real table tennis balls to desired landing points. 
To illustrate the details of this symbiosis, we introduce the task setup, the dense rewards used for the experiments, as well as the hybrid sim and real training (HYSR) that enables practical learning of these tasks.
%
\setlength{\figwidth }{0.55\columnwidth} 
\setlength{\figheight }{.618\figwidth} 
\begin{figure*}
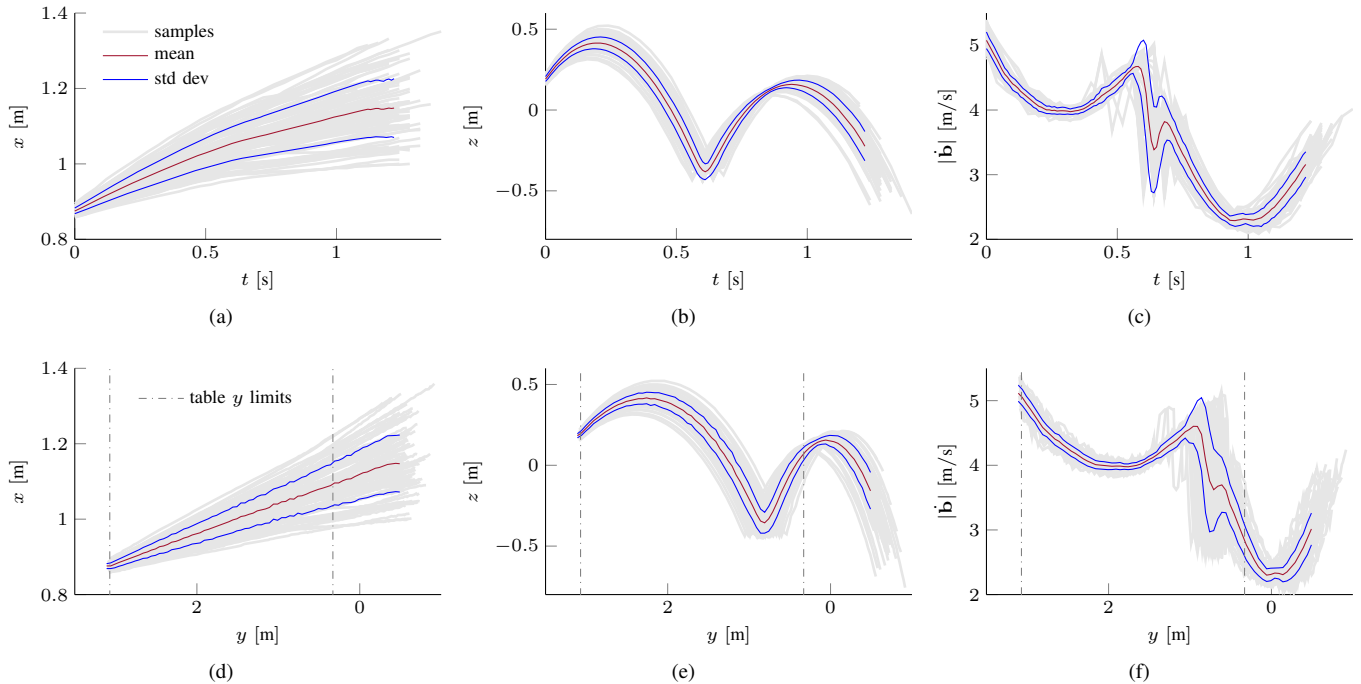

    \centering
    \textbf{Variability in Recorded Ball Trajectories}\par\medskip
    \vspace{-.5cm}
    \subfloat[]{
        \centering\scriptsize%
        \input{figures/recballvariability/xt}
    }
    \subfloat[]{
        \centering\scriptsize%
        \input{figures/recballvariability/zt}
    }
    \subfloat[]{
        \centering\scriptsize%
        \input{figures/recballvariability/velt}
    }
    \newline
    \subfloat[]{
        \scriptsize%
        \input{figures/recballvariability/xy}
    }
    \subfloat[]{
        \scriptsize%
        \input{figures/recballvariability/zy}
    }
    \subfloat[]{
        \scriptsize%
        \input{figures/recballvariability/vely}
    }
    \caption{
        Variability of the recorded ball trajectory dataset $\Drec{}$.
        A ball launcher with fixed settings launches table tennis balls towards the robot.
        A color-based camera vision system~\cite{gomez-gonzalez_reliable_2019} provided the position of the ball at \SI{180}{\hertz}.
        Variability is quantified by the sample mean and sample variance with respect to time a) to c) and along the long side of the table~($y$ coordinate) d) to f).
         The $x$ coordinate aligns with the shorter edge of the table and $z$ coordinate links to the normal of the table plane.
        Although the settings of the ball launcher are constant, the ball varies substantially: 
        1) variability increases throughout the trajectory, especially after the bounce, 
        2) the first bounce on the table~(incoming ball) varies around \SI{50}{\centi\meter} along $y$-coordinate~(subfigure e)).
        3) the agent has to handle deviation of around \SI{40}{\centi\meter} along the $x$-axis and $\sim$\SI{50}{\centi\meter} along the $z$-axis when the ball is in reach for the hit. 
        4) the variability with respect to time and the $y$-axis increases over time, which means that the agent cannot simply learn to start the hit after a particular fixed duration. 
        Additionally, the agent needs to learn to adjust the amount of energy it transfers to each ball as the ball velocity also varies at the end of the table.
    }
    \label{fig:ballvartime}
\end{figure*}
\subsection{Muscular Robot Table Tennis Task Setup}
\label{ssec:task}
The considered table tennis task consists of returning an incoming ball with the racket attached to the robot arm to a desired landing point on the table $\bdes{}\in\R{2}$.
We denote the ball trajectory $\trajb{}=[\sball{t}]_{t=0}^T$ consisting of a series of ball states $\sball{t}=[\bpos{t},\bvel{t}]$ that themselves contain the current ball position $\bpos{t}\in\R{3}$ and velocity $\bvel{t}\in\R{3}$.
In a successful stroke, the robot hits the ball at time $\thit$ and position $\bhit{}$ such that the ball lands on the table plane at position $\bland{}$ at time $\tland$. 
The ball crosses the plane aligned with the net on the incoming and outgoing ball trajectory at position $\bnin{}\in\R{2}$ at time $\tnin$ and $\bnout{}\in\R{2}$ at time $\tnout$ if the robot successfully returns the ball.

Table tennis falls into the general class of dynamic tasks such as baseball~\cite{senoo_ball_2006}, tennis or hockey~\cite{neumann_learning_2014}. 
Dynamic tasks represent a class of problems that are relatively easy to solve for humans but hard for robots. 
The features of dynamic tasks are 1) quick reaction times as adapting to changes in the environment (such as moving balls) must happen fast, 2) precise motions because objects are supposed to reach some goal state~(e.g. desired landing position on the table) and 3) fast and highly accelerated motions. 
The latter point is particularly important for two reasons:
First, a successful strategy can incorporate fast strikes such as a table tennis smash. 
Second, if the desired hitting position of the ball is far from the current racket position, highly accelerated motions help to reach this point in time. 
Thus, the maximum acceleration the system is capable of generating represents the upper limit to the dexterity the agent can develop at such tasks. 
The class of dynamic problems differs from manipulation, where the task itself can be richer than a dynamic task in the sense that the objects and setting can vary largely. 
Unlike in dynamic tasks, however, slow motions and small accelerations are usually sufficient.

Safely generating high accelerations is harder with anthropomorphic robots than with parallel~\cite{kawakami_omron_2016} or Cartesian systems~\cite{huang_adding_2013}.
Low inertia and force transmission without cables ease the control of these systems, making high acceleration and estimating potentially dangerous motions feasible.
Our work, in contrast, investigates table tennis with anthropomorphic robots.
Damages on such systems can occur by breaking cables due to fast-changing control commands or exceeding joint limits when the system cannot stop in time. 
Learning the solution to the task while assuring robot safety makes anthropomorphic table tennis especially challenging. 

Pneumatic artificial muscles~(PAMs) are a particularly useful actuation system for anthropomorphic robots when applied to dynamic tasks.
This actuators contract if the air pressure inside increases, hence at least two PAMs act antagonistically on one degree of freedom~(DoF) as a single PAM can only pull and not push.
In this paper, we leverage the PAM-driven robot arm developed in~\cite{buchler_lightweight_2016,buchler_learning_2019-1} which has four DoFs actuated by eight PAMs. 
Such robots are capable of generating high accelerations due to a high power-to-weight ratio.
At the same time, adjusting the allowed pressures ranges prevents exceeding joint limits despite fast motions. 
We use this property in \ref{ssec:return} and \ref{ssec:smash} to let the RL agent freely explore fast motions without any further safety considerations.
Another benefit of PAM-driven systems is the inherent robustness resulting from passive compliance. 
This property helps reducing damage at impact due to shock absorption~\cite{narioka_humanlike_2012} as well as the adjustable stiffness via co-contracting the PAMs in an antagonistic pair. 
In this paper, we leverage this robustness to apply stochastic policies directly on the desired pressures, which are the low-level actions in this system~(see \fig{fig:stpolicy}). 

These numerous beneficial properties come at the cost of control challenges.
PAMs are highly non-linear systems that change their dynamics with temperature as well as wear and are prone to hysteresis effects.
Thus, modeling such systems for better control is challenging~\cite{tondu_modelling_2012,buchler_control_2018}.
PAMs are, for this reason, predominately used as a testbed for control algorithms rather than dynamic tasks.
In this work, we show that it is possible to satisfy the precision demands of the table tennis task despite the control difficulties of PAM-driven systems by using RL~(see \ref{ssec:return} and \ref{ssec:smash}).

\subsection{Dense Reward Functions For Returning and Smashing}
\label{ssec:rewardfcns}
We formulate the learning problem as an episodic Markov Decision Process~(MDP) 
\begin{equation}
\mathcal{M}=(\mathcal{S},\mathcal{A},r,\mathcal{P},p_0,\gamma)\,
\label{eq:mdp}
\end{equation}
where $\mathcal{S}$ is the state space, $\mathcal{A}$ is the action space, $r:\mathcal{S}\times\mathcal{A}\mapsto\R{}$ is the immediate reward function, $\mathcal{P}:\mathcal{S}\times\mathcal{A}\mapsto\mathcal{S}$ is the transition probability or dynamics, $p_0$ is the initial state distribution and $\gamma\in[0,1]$ the discount factor.
The goal in RL is to find a policy $\pi:\mathcal{S}\mapsto\mathcal{A}$ that maximizes the expected return $J=\E{\vec{\tau}} \sum_{t=0}^T \gamma^t r(\s{t},\ac{t})$ where $\vec{\tau}=(\s{0},\ac{0},\s{1},\ac{1},\ldots,\s{T})$ is the state action trajectory, $\s{0}\sim p_0$, $\ac{t}\sim\pi(\s{t})$ and $\s{t+1}=\mathcal{P}(\s{t},\ac{t})$.

The state $\s{}=[\sball{},\srob{}]$ we use here is composed of the ball state $\sball{}$ and the robot state $\srob{}$.
The robot state $\srob{}=[\q{},\qvel{},\p{}]$ consists of the joint angles $\q{}\in\R{4}$, joint angle velocities $\qvel{}\in\R{4}$ and air pressures in each PAM $\p{}\in\R{8}$.
The ball state $\sball{}=[\bpos{},\bvel{}]$ has been already defined in \ref{ssec:task}.
The system we utilize here actuates each DoF with two PAMs. 
The actions $\ac{}$ are the change in desired pressures in each PAM $\Delta\pdes{}\in\R{8}$.

In practice, the true Markovian state $\s{}$ is not accessible in experiments with real robots.
Especially for PAM driven systems, the Markov state composition is unclear~\cite{buchler_control_2018} leading to a Partially Observable MDP~(POMDP) which assumes to receive observations $\ob{}$ of the true state $\s{}$.
For the sake of clarity, we continue using $\s{}$ instead of $\ob{}$ notation.
The immediate reward function $r(\s{t},\ac{t})$ defines the goal of the task.
The task of returning an incoming ball to a desired landing point can be divided into two stages: 1) manage to hit or touch the ball and 2) fine-tune the impact of the ball with the racket such that it flies in a desired manner. 
As the landing location of the ball changes only in case the robot manages to hit the ball, we introduce a conditional reward function 
\begin{equation}
r=
\begin{cases}
\rtt{}\, & \text{racket touches the ball}\\
\rhit{}              & \text{otherwise}\,,
\end{cases}
\label{eq:condreward}
\end{equation}
where $\rtt{}$ is the table tennis reward
\begin{equation}
\rtt{} =
\begin{cases}
1-c||\bland{}-\bdes{}||^\frac{3}{4}&\text{return task}\\
(1-c||\bland{}-\bdes{}||^\frac{3}{4})\max_{t>\thit}||\bvel{t}||&\text{smash task}
\end{cases}
\label{eq:rtt}
\end{equation}
that evaluates the stroke depending on the ball trajectory after the impact of the ball and racket.
In particular, it considers the distance of the actual landing point $\bland{}$ to the desired landing point $\bdes{}$ for the return task~(see \ref{ssec:return}).
The normalization constant $c=||\trajr{0}-\bdes{}||^{-1}$ is chosen such that $\rtt{}$ is usually within the range $[0,1]$ where $\trajr{0}$ is the initial racket position.
We also cap the table tennis reward $\rtt{}=\max(\rtt{},-0.2)$ in order to avoid too negative rewards in case the ball flies into a random direction with high velocity as happens if hit by an edge of the racket.
We also introduce an exponent to the components of $\rtt{}$ to cause the values of the rewards to be more different closer to the optimal value.
We found the exponent $\nicefrac{3}{4}$ to empirically work well. 
For the smashing task, the agent is supposed to maximize the ball velocity $\bvel{}$ simultaneously.
The product between these two goals forces the agent to be precise \emph{and} play fast balls as $\rtt{}$ is small overall if a single component has a low value.
The hitting reward
\begin{equation}
\rhit{} =-\min_t ||\trajb{t}-\trajr{t}||\,
\label{eq:rhit}
\end{equation}
is a dense reward function representing the minimal Euclidean distance in time between the ball trajectory $\trajb{}$ and Cartesian racket trajectory $\trajr{}$ where $\trajr{t}=\rpos{t}=\mathcal{T}(\q{t})\in\R{3}$ using the forward kinematics function $\mathcal{T}(\cdot)$, the Cartesian racket position $\rpos{t}$ and ignoring the racket orientation.
This reward function encourages the agent to get closer to the ball and finally hit it by providing feedback about how close it missed the racket. 

Note that we do not incorporate \textit{any} safety precautions such as state constraints like joint ranges, minimal accelerations $||\ddot{\vec{q}}_t||$ or change in actions $||\vec{a}_t-\vec{a}_{t-1}||$ into the reward function.
On the contrary, we add a term that favors faster hitting motions. 
By doing so, we let the solution emerge purely from the hardware and the reward function that - in our case - incorporates only task-related terms.  
\subsection{HYSR: Hybrid Sim and Real Training}
\label{ssec:simrealtraining}
Running experiments with real robots and real objects for millions of time steps is a tedious practical effort.
For instance, learning robot table tennis using model-free RL involves launching balls automatically, removing them from the scene after the stroke, and returning them to the ball reservoir. 
Automating this pipeline takes a substantial amount of work. 
Hence, we considered training with simulated balls. 
Training in simulation, however, can be problematic: 
Simulated balls might differ too much from the real ball so that the learned policy might not be useful when playing with real balls.
Additionally, the lack of good models of PAM-driven systems~\cite{tondu_modelling_2012,buchler_control_2018} renders simulating the real robot infeasible.

For this reason, we introduce a hybrid sim and real training~(HYSR) where the key idea is to use real data as much as possible and simulations wherever necessary. 
Specifically, actions $\ac{}$ sampled from the policy $\pi$ are applied to the real robot, but the ball exists only in simulation during training. 
In simulation, we replay a single ball trajectory per episode sampled uniformly from a prerecorded data set $\Drec{}=[\trajrec{i}{}]^{\Nrec}_{i=0}$.
Within the recorded data set $\Drec{}$ the $i$-th trajectory consist of a sequence of ball states $\trajrec{i}{t}=\sballrec{t}$.

The real robot is copied to the simulation by overwriting the simulated with the real robot state $\srob{}$ in every time step.
In this manner, the real robot moves in simulation in the same way as in reality.
In this manner, the simulation and the real scenery stay identical~(the only difference being that the ball has been launched before training) until contact of the ball with the racket.
At this point, it is impossible to predict the subsequent ball trajectory from $\Drec{}$.
For this reason, we start simulating the ball after the impact, which allows us to calculate the landing point of the ball $\bland{}$ more reliably as we simulate at the latest possible point.
\begin{algorithm}[t]
    \caption{HYSR: Hybrid Sim and Real Training}
    \label{alg:simrealtr}
    \begin{algorithmic}[1]
        \WHILE{max\_timesteps not reached}
        \STATE $i\sim\text{uniform}(0,\Nrec)$
        \STATE $\trajrecargone{}\gets\Drec{i}$
        \STATE $t\gets0$
        \STATE racketTouched$\gets$\emph{false}
        \WHILE{episode end not reached}
        \STATE $\srob{t}\gets$readSensors()
        \IF{racketTouched}
        \STATE $\sballsim{t}\gets \text{sim}(\sballsim{t-1})$
        \STATE $\s{t}\gets[\srob{t},\sballsim{t}]$
        \ELSE
        \STATE $\sballrec{t}\gets\trajrecargone{t}$
        \STATE $\s{t}\gets[\srob{t},\sballrec{t}]$
        \ENDIF
        \IF{$\textrm{racketTouchesBall}(\s{t})$}
        \STATE racketTouched$\gets$ \emph{true}
        \ENDIF
        \STATE $\ac{t}\sim\pi(\s{t})$
        \STATE $t\gets t+1$
        \ENDWHILE
        \STATE update $\pi$ using PPO
        \ENDWHILE
    \end{algorithmic}
\end{algorithm}
Essential for sufficiently accurate transfer from simulated to the real ball within HYSR is the rebound model of the ball with the racket.
We found the rebound model from~\cite{mulling_biomimetic_2011} to work well after optimizing its parameters empirically. 
The model
\begin{equation}
\bvel{\text{out}}-\rvel{\thit||}=\epsilon_\text{R}(-\bvel{\text{in}}+\rvel{\thit||})
\label{eq:rbrebound}
\end{equation}
calculates the outgoing velocity of the ball $\bvel{\text{out}}$ from the ball velocity $\bvel{\text{in}}$ before impact, the racket speed $\rvel{\thit||}$ at impact~(all measured along the racket normal) and the restitution coefficient of the racket $\epsilon_\text{R}$.
Note that this model assumes no spin.
\alg{alg:simrealtr} summarizes a single episode of the training procedure.

It is important to mention that, although the settings of the ball launcher have been kept fixed, the $\Nrec{}=100$ recorded ball trajectories in $\Drec{}$ largely vary as can be seen in \fig{fig:ballvartime}.
Variability arises through the noise in the color-based vision system that automatically detects balls and returns their Cartesian position.
Another reason is that the ball launcher adds little noise to each ball, accumulating to more significant deviations the farther the ball flies. 

HYSR allows us to leverage the simulation conveniently to estimate the landing position $\bland{}$ and other context entities such as distances or boolean contact indicators that are important for the reward~(see \eq{eq:rhit} and \eq{eq:rtt}).
Besides, we avoid collecting and launching real balls. 
Note that this way of training can serve as an entry point for sim2real techniques such as domain randomization or curriculum learning. 
For instance, the progress of the training could be used to update the ball's initial state $\sballi$, or the ball trajectory $\trajb{}$ can be perturbed.

%% file: figures/recballvariability/xt.tex
%
%
\definecolor{mycolor1}{rgb}{0.63500,0.07800,0.18400}%
\begin{tikzpicture}

\begin{axis}[%
width=\figwidth,
height=\figheight,
at={(0\figwidth,0\figheight)},
scale only axis,
xmin=0.000,
xmax=1.400,
xlabel style={font=\color{white!15!black}},
xlabel={$t~[\textrm{s}]$},
ymin=0.800,
ymax=1.400,
ylabel style={font=\color{white!15!black}},
ylabel={$x~[\textrm{m}]$},
axis background/.style={fill=white},
axis x line*=bottom,
axis y line*=left,
legend style={legend cell align=left, align=left, fill=none, draw=none},
ylabel near ticks,
xlabel near ticks,
legend style={at={(.05,1)},anchor=north west,legend cell align=left,align=left,fill=none,draw=none}
]
\addplot [color=white!90!black, line width=1.0pt]
  table[row sep=crcr]{%
0.000	0.878\\
0.040	0.891\\
0.080	0.905\\
0.120	0.917\\
0.160	0.930\\
0.240	0.952\\
0.280	0.966\\
0.320	0.977\\
0.400	0.996\\
0.520	1.024\\
0.560	1.034\\
0.600	1.040\\
0.640	1.046\\
0.720	1.055\\
0.760	1.058\\
0.880	1.070\\
0.920	1.072\\
0.960	1.076\\
1.000	1.079\\
1.040	1.085\\
1.080	1.087\\
1.120	1.093\\
1.160	1.093\\
1.200	1.097\\
1.240	1.102\\
1.280	1.105\\
};
\addlegendentry{samples}

\addplot [color=white!90!black, line width=1.0pt, forget plot]
  table[row sep=crcr]{%
0.000	0.864\\
0.080	0.883\\
0.120	0.891\\
0.160	0.899\\
0.200	0.906\\
0.240	0.915\\
0.280	0.922\\
0.320	0.931\\
0.360	0.939\\
0.400	0.945\\
0.440	0.953\\
0.480	0.961\\
0.560	0.974\\
0.600	0.979\\
0.640	0.982\\
0.680	0.987\\
0.800	0.997\\
0.840	1.002\\
0.880	1.006\\
0.920	1.010\\
0.960	1.016\\
1.080	1.027\\
1.120	1.031\\
1.160	1.037\\
1.200	1.041\\
};
\addplot [color=white!90!black, line width=1.0pt, forget plot]
  table[row sep=crcr]{%
0.000	0.876\\
0.040	0.892\\
0.080	0.909\\
0.120	0.924\\
0.160	0.937\\
0.200	0.952\\
0.240	0.966\\
0.280	0.980\\
0.320	0.995\\
0.360	1.008\\
0.400	1.023\\
0.440	1.035\\
0.560	1.076\\
0.600	1.083\\
0.680	1.099\\
0.720	1.109\\
0.840	1.134\\
0.880	1.143\\
0.920	1.151\\
0.960	1.163\\
1.000	1.169\\
1.040	1.177\\
1.080	1.185\\
1.160	1.200\\
1.200	1.210\\
1.240	1.216\\
};
\addplot [color=white!90!black, line width=1.0pt, forget plot]
  table[row sep=crcr]{%
0.000	0.866\\
0.080	0.887\\
0.120	0.899\\
0.160	0.909\\
0.200	0.921\\
0.240	0.932\\
0.280	0.943\\
0.320	0.953\\
0.360	0.964\\
0.440	0.982\\
0.480	0.993\\
0.560	1.012\\
0.640	1.028\\
0.680	1.032\\
0.720	1.037\\
0.760	1.044\\
0.800	1.049\\
0.880	1.063\\
0.960	1.074\\
1.000	1.083\\
1.040	1.086\\
1.120	1.100\\
1.160	1.105\\
1.200	1.112\\
1.240	1.115\\
1.280	1.123\\
};
\addplot [color=white!90!black, line width=1.0pt, forget plot]
  table[row sep=crcr]{%
0.000	0.876\\
0.040	0.893\\
0.080	0.906\\
0.200	0.948\\
0.280	0.977\\
0.480	1.040\\
0.520	1.051\\
0.560	1.061\\
0.600	1.067\\
0.640	1.075\\
0.880	1.117\\
0.960	1.129\\
1.000	1.139\\
1.120	1.160\\
1.160	1.163\\
1.200	1.170\\
1.240	1.174\\
};
\addplot [color=white!90!black, line width=1.0pt, forget plot]
  table[row sep=crcr]{%
0.000	0.876\\
0.040	0.886\\
0.080	0.898\\
0.120	0.909\\
0.160	0.920\\
0.200	0.933\\
0.240	0.944\\
0.280	0.953\\
0.360	0.975\\
0.400	0.985\\
0.520	1.013\\
0.560	1.021\\
0.600	1.031\\
0.640	1.038\\
0.680	1.043\\
0.920	1.084\\
0.960	1.090\\
1.000	1.101\\
1.040	1.106\\
1.080	1.114\\
1.120	1.120\\
1.160	1.127\\
1.200	1.133\\
1.240	1.141\\
1.320	1.154\\
1.360	1.158\\
};
\addplot [color=white!90!black, line width=1.0pt, forget plot]
  table[row sep=crcr]{%
0.000	0.864\\
0.080	0.885\\
0.160	0.903\\
0.360	0.948\\
0.440	0.965\\
0.480	0.974\\
0.600	0.997\\
0.640	1.002\\
0.680	1.004\\
0.800	1.011\\
0.880	1.016\\
0.920	1.018\\
0.960	1.021\\
1.000	1.026\\
1.040	1.028\\
1.080	1.030\\
1.120	1.030\\
1.200	1.039\\
1.280	1.041\\
};
\addplot [color=white!90!black, line width=1.0pt, forget plot]
  table[row sep=crcr]{%
0.000	0.878\\
0.040	0.892\\
0.080	0.902\\
0.120	0.916\\
0.160	0.928\\
0.200	0.941\\
0.240	0.953\\
0.320	0.974\\
0.400	0.997\\
0.520	1.023\\
0.560	1.033\\
0.600	1.040\\
0.640	1.052\\
0.680	1.061\\
0.720	1.072\\
0.760	1.086\\
0.800	1.098\\
0.840	1.109\\
0.880	1.120\\
0.920	1.131\\
0.960	1.143\\
1.000	1.157\\
1.040	1.167\\
1.080	1.180\\
1.120	1.191\\
1.160	1.199\\
1.200	1.215\\
1.280	1.237\\
1.320	1.248\\
};
\addplot [color=white!90!black, line width=1.0pt, forget plot]
  table[row sep=crcr]{%
0.000	0.878\\
0.040	0.889\\
0.080	0.904\\
0.200	0.941\\
0.240	0.956\\
0.280	0.967\\
0.360	0.992\\
0.400	1.003\\
0.440	1.016\\
0.520	1.039\\
0.560	1.049\\
0.640	1.066\\
0.680	1.072\\
0.840	1.103\\
0.920	1.117\\
0.960	1.126\\
1.000	1.132\\
1.040	1.143\\
1.080	1.150\\
1.120	1.155\\
1.200	1.169\\
};
\addplot [color=white!90!black, line width=1.0pt, forget plot]
  table[row sep=crcr]{%
0.000	0.879\\
0.120	0.923\\
0.160	0.936\\
0.240	0.965\\
0.280	0.977\\
0.320	0.991\\
0.360	1.005\\
0.440	1.029\\
0.480	1.036\\
0.520	1.049\\
0.600	1.069\\
0.680	1.082\\
0.760	1.095\\
0.800	1.100\\
0.880	1.114\\
0.960	1.124\\
1.000	1.131\\
1.040	1.136\\
1.080	1.142\\
1.120	1.147\\
1.160	1.151\\
1.200	1.157\\
1.280	1.166\\
};
\addplot [color=white!90!black, line width=1.0pt, forget plot]
  table[row sep=crcr]{%
0.000	0.872\\
0.040	0.885\\
0.120	0.909\\
0.160	0.923\\
0.240	0.948\\
0.320	0.971\\
0.440	1.008\\
0.480	1.020\\
0.560	1.042\\
0.600	1.054\\
0.680	1.062\\
0.720	1.066\\
0.800	1.079\\
0.880	1.090\\
0.920	1.094\\
0.960	1.102\\
1.000	1.113\\
1.040	1.115\\
1.160	1.134\\
1.200	1.137\\
1.240	1.144\\
};
\addplot [color=white!90!black, line width=1.0pt, forget plot]
  table[row sep=crcr]{%
0.000	0.874\\
0.040	0.887\\
0.080	0.902\\
0.120	0.915\\
0.160	0.927\\
0.240	0.948\\
0.280	0.960\\
0.400	0.992\\
0.440	1.001\\
0.480	1.012\\
0.600	1.035\\
0.640	1.044\\
0.680	1.057\\
0.720	1.062\\
0.760	1.070\\
0.800	1.079\\
0.840	1.087\\
0.880	1.095\\
0.920	1.103\\
0.960	1.113\\
1.000	1.118\\
1.080	1.136\\
1.120	1.144\\
1.160	1.149\\
1.240	1.162\\
1.280	1.166\\
};
\addplot [color=white!90!black, line width=1.0pt, forget plot]
  table[row sep=crcr]{%
0.000	0.863\\
0.080	0.880\\
0.120	0.889\\
0.160	0.898\\
0.240	0.916\\
0.280	0.923\\
0.400	0.949\\
0.440	0.957\\
0.560	0.979\\
0.600	0.988\\
0.640	0.993\\
0.800	1.011\\
0.880	1.020\\
0.920	1.023\\
0.960	1.028\\
1.000	1.035\\
1.040	1.038\\
1.080	1.043\\
1.120	1.046\\
1.160	1.049\\
1.200	1.053\\
1.240	1.060\\
1.280	1.063\\
1.320	1.068\\
};
\addplot [color=white!90!black, line width=1.0pt, forget plot]
  table[row sep=crcr]{%
0.000	0.879\\
0.040	0.894\\
0.080	0.910\\
0.120	0.924\\
0.160	0.942\\
0.200	0.954\\
0.240	0.969\\
0.280	0.981\\
0.320	0.995\\
0.360	1.008\\
0.440	1.032\\
0.480	1.041\\
0.560	1.063\\
0.600	1.072\\
0.640	1.078\\
0.720	1.091\\
0.800	1.104\\
0.920	1.122\\
0.960	1.129\\
1.000	1.136\\
1.040	1.141\\
1.080	1.149\\
1.120	1.152\\
};
\addplot [color=white!90!black, line width=1.0pt, forget plot]
  table[row sep=crcr]{%
0.000	0.883\\
0.040	0.899\\
0.080	0.916\\
0.120	0.935\\
0.160	0.946\\
0.200	0.966\\
0.280	0.994\\
0.320	1.009\\
0.360	1.025\\
0.480	1.061\\
0.520	1.072\\
0.560	1.085\\
0.640	1.103\\
0.680	1.111\\
0.720	1.118\\
0.800	1.136\\
0.880	1.150\\
0.960	1.167\\
1.000	1.179\\
1.040	1.182\\
1.080	1.191\\
1.160	1.208\\
1.200	1.213\\
1.240	1.222\\
};
\addplot [color=white!90!black, line width=1.0pt, forget plot]
  table[row sep=crcr]{%
0.000	0.869\\
0.040	0.879\\
0.120	0.902\\
0.160	0.912\\
0.200	0.923\\
0.240	0.934\\
0.280	0.944\\
0.320	0.955\\
0.360	0.967\\
0.400	0.976\\
0.440	0.987\\
0.480	0.996\\
0.560	1.015\\
0.600	1.023\\
0.640	1.027\\
0.680	1.033\\
0.720	1.038\\
0.760	1.043\\
0.800	1.050\\
0.880	1.060\\
0.920	1.066\\
0.960	1.073\\
1.000	1.077\\
1.040	1.086\\
1.080	1.088\\
1.120	1.092\\
1.160	1.097\\
1.200	1.104\\
};
\addplot [color=white!90!black, line width=1.0pt, forget plot]
  table[row sep=crcr]{%
0.000	0.877\\
0.040	0.893\\
0.120	0.921\\
0.160	0.936\\
0.240	0.963\\
0.280	0.975\\
0.320	0.986\\
0.360	1.002\\
0.440	1.025\\
0.520	1.047\\
0.560	1.058\\
0.600	1.067\\
0.640	1.078\\
0.680	1.089\\
0.720	1.097\\
0.800	1.116\\
0.840	1.126\\
0.880	1.135\\
0.920	1.146\\
0.960	1.156\\
1.000	1.169\\
1.040	1.176\\
1.080	1.185\\
1.120	1.191\\
1.160	1.203\\
1.280	1.227\\
};
\addplot [color=white!90!black, line width=1.0pt, forget plot]
  table[row sep=crcr]{%
0.000	0.871\\
0.040	0.882\\
0.160	0.921\\
0.320	0.964\\
0.400	0.983\\
0.440	0.992\\
0.480	1.002\\
0.520	1.009\\
0.560	1.017\\
0.600	1.023\\
0.680	1.026\\
0.720	1.027\\
0.760	1.027\\
0.800	1.029\\
0.840	1.029\\
0.880	1.030\\
0.920	1.030\\
0.960	1.031\\
1.040	1.036\\
1.120	1.034\\
1.160	1.036\\
1.200	1.034\\
1.240	1.039\\
1.280	1.036\\
};
\addplot [color=white!90!black, line width=1.0pt, forget plot]
  table[row sep=crcr]{%
0.000	0.872\\
0.040	0.882\\
0.120	0.907\\
0.160	0.921\\
0.240	0.944\\
0.320	0.966\\
0.360	0.978\\
0.480	1.008\\
0.520	1.018\\
0.560	1.027\\
0.640	1.041\\
0.680	1.047\\
0.720	1.055\\
0.760	1.058\\
0.800	1.064\\
0.840	1.068\\
0.880	1.074\\
0.920	1.079\\
1.000	1.093\\
1.040	1.097\\
1.080	1.103\\
1.120	1.107\\
1.160	1.112\\
1.200	1.115\\
};
\addplot [color=white!90!black, line width=1.0pt, forget plot]
  table[row sep=crcr]{%
0.000	0.878\\
0.040	0.893\\
0.080	0.911\\
0.200	0.955\\
0.240	0.972\\
0.280	0.986\\
0.320	1.001\\
0.440	1.041\\
0.480	1.055\\
0.520	1.066\\
0.560	1.079\\
0.600	1.089\\
0.640	1.097\\
0.720	1.116\\
0.800	1.131\\
0.840	1.141\\
0.920	1.158\\
0.960	1.168\\
1.000	1.177\\
1.040	1.181\\
1.080	1.191\\
1.160	1.208\\
1.240	1.225\\
1.280	1.231\\
};
\addplot [color=white!90!black, line width=1.0pt, forget plot]
  table[row sep=crcr]{%
0.000	0.879\\
0.040	0.888\\
0.080	0.908\\
0.120	0.924\\
0.160	0.938\\
0.200	0.952\\
0.360	1.004\\
0.400	1.015\\
0.440	1.028\\
0.560	1.059\\
0.600	1.067\\
0.680	1.080\\
0.720	1.091\\
0.800	1.105\\
0.920	1.127\\
0.960	1.134\\
1.000	1.142\\
1.040	1.148\\
1.080	1.158\\
1.120	1.166\\
1.160	1.172\\
1.200	1.178\\
1.240	1.183\\
1.280	1.192\\
};
\addplot [color=white!90!black, line width=1.0pt, forget plot]
  table[row sep=crcr]{%
0.000	0.863\\
0.040	0.872\\
0.120	0.894\\
0.160	0.904\\
0.200	0.914\\
0.240	0.922\\
0.280	0.931\\
0.320	0.942\\
0.400	0.959\\
0.440	0.970\\
0.520	0.985\\
0.560	0.995\\
0.600	1.003\\
0.680	1.017\\
0.720	1.021\\
0.840	1.037\\
0.880	1.041\\
0.920	1.048\\
0.960	1.057\\
1.000	1.061\\
1.040	1.067\\
1.080	1.074\\
1.120	1.079\\
1.160	1.083\\
1.280	1.101\\
1.320	1.107\\
};
\addplot [color=white!90!black, line width=1.0pt, forget plot]
  table[row sep=crcr]{%
0.000	0.877\\
0.040	0.891\\
0.160	0.929\\
0.200	0.942\\
0.240	0.954\\
0.280	0.967\\
0.320	0.977\\
0.400	1.002\\
0.480	1.025\\
0.520	1.034\\
0.560	1.045\\
0.600	1.056\\
0.640	1.064\\
0.680	1.070\\
0.800	1.096\\
0.840	1.101\\
0.920	1.120\\
1.080	1.151\\
1.120	1.158\\
1.160	1.168\\
};
\addplot [color=white!90!black, line width=1.0pt, forget plot]
  table[row sep=crcr]{%
0.000	0.864\\
0.040	0.873\\
0.120	0.891\\
0.160	0.900\\
0.240	0.916\\
0.280	0.922\\
0.320	0.931\\
0.360	0.939\\
0.400	0.945\\
0.440	0.953\\
0.480	0.959\\
0.520	0.965\\
0.560	0.972\\
0.600	0.977\\
0.640	0.979\\
0.800	0.980\\
0.840	0.982\\
0.880	0.982\\
0.920	0.984\\
0.960	0.986\\
1.000	0.989\\
1.040	0.989\\
1.080	0.994\\
1.120	0.994\\
1.160	0.997\\
1.200	0.998\\
1.240	0.998\\
1.280	1.000\\
};
\addplot [color=white!90!black, line width=1.0pt, forget plot]
  table[row sep=crcr]{%
0.000	0.874\\
0.040	0.885\\
0.120	0.913\\
0.160	0.922\\
0.200	0.937\\
0.240	0.949\\
0.280	0.961\\
0.320	0.973\\
0.360	0.982\\
0.400	0.996\\
0.440	1.005\\
0.480	1.016\\
0.600	1.047\\
0.640	1.055\\
0.680	1.060\\
0.720	1.072\\
0.880	1.100\\
0.920	1.108\\
0.960	1.119\\
1.000	1.124\\
1.080	1.141\\
1.120	1.148\\
1.200	1.160\\
};
\addplot [color=white!90!black, line width=1.0pt, forget plot]
  table[row sep=crcr]{%
0.000	0.874\\
0.040	0.883\\
0.080	0.899\\
0.160	0.926\\
0.240	0.949\\
0.280	0.963\\
0.320	0.974\\
0.360	0.986\\
0.400	0.999\\
0.440	1.011\\
0.520	1.031\\
0.560	1.041\\
0.600	1.050\\
0.640	1.055\\
0.680	1.057\\
0.720	1.064\\
0.840	1.080\\
0.920	1.090\\
0.960	1.098\\
1.040	1.106\\
1.120	1.118\\
1.160	1.121\\
1.200	1.127\\
1.240	1.130\\
};
\addplot [color=white!90!black, line width=1.0pt, forget plot]
  table[row sep=crcr]{%
0.000	0.873\\
0.040	0.884\\
0.080	0.897\\
0.120	0.910\\
0.160	0.924\\
0.200	0.932\\
0.240	0.947\\
0.280	0.958\\
0.360	0.982\\
0.400	0.992\\
0.480	1.014\\
0.520	1.026\\
0.600	1.043\\
0.640	1.050\\
0.680	1.053\\
0.720	1.062\\
0.760	1.067\\
0.800	1.074\\
0.840	1.079\\
0.920	1.092\\
0.960	1.102\\
1.000	1.107\\
1.040	1.114\\
1.080	1.118\\
1.120	1.125\\
1.160	1.129\\
1.240	1.141\\
1.280	1.146\\
};
\addplot [color=white!90!black, line width=1.0pt, forget plot]
  table[row sep=crcr]{%
0.000	0.872\\
0.080	0.898\\
0.120	0.909\\
0.160	0.921\\
0.320	0.964\\
0.360	0.972\\
0.440	0.993\\
0.480	1.003\\
0.560	1.021\\
0.600	1.029\\
0.640	1.037\\
0.720	1.050\\
0.840	1.072\\
0.960	1.093\\
1.000	1.101\\
1.040	1.107\\
1.120	1.123\\
1.160	1.126\\
1.200	1.135\\
1.240	1.141\\
};
\addplot [color=white!90!black, line width=1.0pt, forget plot]
  table[row sep=crcr]{%
0.000	0.869\\
0.080	0.891\\
0.160	0.910\\
0.280	0.940\\
0.320	0.948\\
0.360	0.958\\
0.400	0.967\\
0.440	0.977\\
0.520	0.994\\
0.600	1.009\\
0.640	1.015\\
0.680	1.022\\
0.720	1.026\\
0.760	1.031\\
0.840	1.043\\
0.880	1.046\\
0.920	1.054\\
0.960	1.059\\
1.000	1.066\\
1.040	1.071\\
1.080	1.079\\
1.120	1.084\\
1.160	1.088\\
1.200	1.091\\
1.240	1.098\\
1.280	1.102\\
};
\addplot [color=white!90!black, line width=1.0pt, forget plot]
  table[row sep=crcr]{%
0.000	0.883\\
0.120	0.937\\
0.160	0.957\\
0.240	0.991\\
0.400	1.058\\
0.440	1.075\\
0.480	1.089\\
0.520	1.106\\
0.560	1.122\\
0.600	1.137\\
0.640	1.151\\
0.720	1.176\\
0.760	1.180\\
0.800	1.186\\
0.840	1.195\\
0.880	1.202\\
1.000	1.228\\
1.040	1.235\\
1.080	1.243\\
1.120	1.249\\
1.160	1.258\\
1.200	1.266\\
};
\addplot [color=white!90!black, line width=1.0pt, forget plot]
  table[row sep=crcr]{%
0.000	0.880\\
0.040	0.893\\
0.080	0.905\\
0.120	0.920\\
0.160	0.934\\
0.360	0.991\\
0.520	1.031\\
0.560	1.038\\
0.600	1.046\\
0.640	1.057\\
0.680	1.065\\
0.720	1.082\\
0.760	1.091\\
0.800	1.102\\
0.840	1.112\\
0.880	1.125\\
0.920	1.135\\
1.000	1.160\\
1.040	1.170\\
1.080	1.183\\
1.160	1.203\\
1.200	1.212\\
};
\addplot [color=white!90!black, line width=1.0pt, forget plot]
  table[row sep=crcr]{%
0.000	0.880\\
0.080	0.906\\
0.120	0.919\\
0.160	0.934\\
0.200	0.943\\
0.240	0.959\\
0.280	0.970\\
0.400	1.007\\
0.480	1.029\\
0.520	1.038\\
0.600	1.058\\
0.640	1.062\\
0.680	1.059\\
0.720	1.072\\
0.760	1.074\\
0.840	1.083\\
0.880	1.089\\
0.920	1.092\\
0.960	1.098\\
1.000	1.102\\
1.040	1.107\\
1.080	1.112\\
1.120	1.115\\
1.160	1.118\\
1.240	1.127\\
1.280	1.128\\
};
\addplot [color=white!90!black, line width=1.0pt, forget plot]
  table[row sep=crcr]{%
0.000	0.872\\
0.040	0.885\\
0.160	0.924\\
0.200	0.935\\
0.240	0.950\\
0.320	0.972\\
0.360	0.985\\
0.400	0.996\\
0.440	1.010\\
0.520	1.031\\
0.560	1.042\\
0.600	1.052\\
0.640	1.059\\
0.680	1.063\\
0.720	1.075\\
0.800	1.088\\
0.840	1.096\\
0.920	1.110\\
0.960	1.119\\
1.000	1.126\\
1.040	1.134\\
1.080	1.142\\
1.120	1.148\\
1.160	1.155\\
};
\addplot [color=white!90!black, line width=1.0pt, forget plot]
  table[row sep=crcr]{%
0.000	0.885\\
0.040	0.904\\
0.080	0.920\\
0.120	0.935\\
0.160	0.958\\
0.240	0.990\\
0.280	1.007\\
0.320	1.022\\
0.360	1.040\\
0.440	1.068\\
0.480	1.081\\
0.520	1.092\\
0.560	1.109\\
0.600	1.121\\
0.680	1.142\\
0.720	1.151\\
0.760	1.162\\
0.840	1.183\\
0.880	1.193\\
0.920	1.204\\
1.040	1.236\\
1.080	1.245\\
1.120	1.256\\
1.200	1.276\\
1.240	1.282\\
};
\addplot [color=white!90!black, line width=1.0pt, forget plot]
  table[row sep=crcr]{%
0.000	0.871\\
0.040	0.884\\
0.080	0.898\\
0.280	0.962\\
0.360	0.985\\
0.400	1.000\\
0.440	1.013\\
0.520	1.032\\
0.560	1.042\\
0.600	1.051\\
0.640	1.061\\
0.680	1.067\\
0.720	1.075\\
0.760	1.079\\
0.960	1.110\\
1.000	1.118\\
1.040	1.123\\
1.080	1.130\\
1.120	1.138\\
1.160	1.143\\
1.200	1.149\\
};
\addplot [color=white!90!black, line width=1.0pt, forget plot]
  table[row sep=crcr]{%
0.000	0.876\\
0.040	0.888\\
0.080	0.900\\
0.120	0.914\\
0.200	0.938\\
0.240	0.952\\
0.280	0.960\\
0.320	0.974\\
0.480	1.020\\
0.520	1.028\\
0.560	1.039\\
0.600	1.047\\
0.640	1.054\\
0.680	1.057\\
0.720	1.070\\
0.760	1.072\\
0.800	1.080\\
0.840	1.085\\
0.920	1.098\\
0.960	1.103\\
1.000	1.112\\
1.040	1.117\\
1.080	1.124\\
1.120	1.132\\
1.160	1.139\\
1.200	1.145\\
1.240	1.148\\
};
\addplot [color=white!90!black, line width=1.0pt, forget plot]
  table[row sep=crcr]{%
0.000	0.878\\
0.040	0.893\\
0.160	0.932\\
0.200	0.947\\
0.280	0.972\\
0.360	0.995\\
0.400	1.010\\
0.440	1.018\\
0.520	1.039\\
0.560	1.049\\
0.600	1.056\\
0.640	1.066\\
0.680	1.074\\
0.720	1.084\\
0.760	1.089\\
0.840	1.107\\
0.880	1.116\\
0.960	1.133\\
1.000	1.144\\
1.040	1.150\\
1.080	1.159\\
1.120	1.168\\
1.160	1.176\\
1.200	1.184\\
1.240	1.191\\
1.280	1.199\\
};
\addplot [color=white!90!black, line width=1.0pt, forget plot]
  table[row sep=crcr]{%
0.000	0.885\\
0.040	0.899\\
0.080	0.917\\
0.160	0.950\\
0.200	0.968\\
0.240	0.984\\
0.280	1.001\\
0.320	1.014\\
0.400	1.042\\
0.480	1.072\\
0.520	1.080\\
0.560	1.098\\
0.600	1.113\\
0.640	1.126\\
0.720	1.135\\
0.800	1.149\\
0.840	1.157\\
0.920	1.169\\
0.960	1.177\\
1.000	1.182\\
1.040	1.191\\
1.080	1.197\\
};
\addplot [color=white!90!black, line width=1.0pt, forget plot]
  table[row sep=crcr]{%
0.000	0.876\\
0.040	0.892\\
0.080	0.907\\
0.120	0.919\\
0.160	0.936\\
0.240	0.962\\
0.440	1.025\\
0.520	1.048\\
0.600	1.069\\
0.640	1.076\\
0.680	1.081\\
0.720	1.091\\
0.760	1.096\\
0.800	1.105\\
0.920	1.126\\
0.960	1.136\\
1.000	1.141\\
1.040	1.151\\
1.080	1.153\\
1.120	1.160\\
1.160	1.168\\
};
\addplot [color=white!90!black, line width=1.0pt, forget plot]
  table[row sep=crcr]{%
0.000	0.888\\
0.040	0.905\\
0.080	0.923\\
0.200	0.973\\
0.240	0.993\\
0.320	1.023\\
0.360	1.040\\
0.400	1.055\\
0.440	1.072\\
0.560	1.115\\
0.640	1.139\\
0.680	1.150\\
0.720	1.160\\
0.760	1.171\\
0.840	1.189\\
0.880	1.202\\
0.920	1.213\\
0.960	1.225\\
1.000	1.233\\
1.080	1.257\\
1.120	1.266\\
};
\addplot [color=white!90!black, line width=1.0pt, forget plot]
  table[row sep=crcr]{%
0.000	0.880\\
0.080	0.915\\
0.200	0.959\\
0.240	0.976\\
0.280	0.990\\
0.320	1.002\\
0.440	1.043\\
0.480	1.055\\
0.560	1.078\\
0.600	1.088\\
0.640	1.097\\
0.760	1.119\\
0.800	1.129\\
0.840	1.136\\
0.880	1.146\\
0.920	1.155\\
0.960	1.162\\
1.000	1.173\\
1.040	1.181\\
1.080	1.191\\
1.120	1.199\\
1.160	1.206\\
1.200	1.215\\
};
\addplot [color=white!90!black, line width=1.0pt, forget plot]
  table[row sep=crcr]{%
0.000	0.869\\
0.040	0.881\\
0.080	0.893\\
0.120	0.902\\
0.160	0.916\\
0.280	0.949\\
0.320	0.957\\
0.360	0.974\\
0.440	0.995\\
0.480	1.002\\
0.520	1.015\\
0.560	1.026\\
0.640	1.043\\
0.680	1.046\\
0.760	1.056\\
0.840	1.063\\
0.960	1.075\\
1.040	1.086\\
1.120	1.091\\
1.160	1.098\\
1.200	1.101\\
1.240	1.103\\
};
\addplot [color=white!90!black, line width=1.0pt, forget plot]
  table[row sep=crcr]{%
0.000	0.870\\
0.040	0.883\\
0.080	0.892\\
0.160	0.916\\
0.240	0.936\\
0.320	0.956\\
0.400	0.973\\
0.440	1.011\\
0.480	0.998\\
0.520	0.998\\
0.600	1.012\\
0.640	1.019\\
0.680	1.023\\
0.760	1.033\\
0.800	1.036\\
0.880	1.047\\
0.920	1.050\\
0.960	1.056\\
1.000	1.061\\
1.080	1.067\\
1.160	1.076\\
1.200	1.079\\
1.240	1.084\\
1.280	1.089\\
1.320	1.090\\
};
\addplot [color=white!90!black, line width=1.0pt, forget plot]
  table[row sep=crcr]{%
0.000	0.872\\
0.040	0.885\\
0.200	0.932\\
0.360	0.969\\
0.400	0.978\\
0.440	1.015\\
0.480	0.993\\
0.520	1.001\\
0.560	1.011\\
0.600	1.019\\
0.640	1.026\\
0.680	1.031\\
0.720	1.038\\
0.760	1.045\\
0.880	1.065\\
0.960	1.079\\
1.000	1.088\\
1.040	1.096\\
1.080	1.100\\
1.120	1.108\\
1.160	1.115\\
1.200	1.120\\
};
\addplot [color=white!90!black, line width=1.0pt, forget plot]
  table[row sep=crcr]{%
0.000	0.880\\
0.240	0.970\\
0.400	1.025\\
0.440	1.034\\
0.480	1.049\\
0.520	1.060\\
0.560	1.072\\
0.600	1.080\\
0.640	1.091\\
0.680	1.097\\
0.720	1.108\\
0.760	1.115\\
0.800	1.124\\
0.840	1.133\\
0.880	1.140\\
0.960	1.156\\
1.000	1.166\\
1.040	1.172\\
1.080	1.181\\
1.120	1.186\\
1.160	1.196\\
1.200	1.204\\
};
\addplot [color=white!90!black, line width=1.0pt, forget plot]
  table[row sep=crcr]{%
0.000	0.867\\
0.040	0.880\\
0.160	0.916\\
0.200	0.928\\
0.240	0.938\\
0.280	0.950\\
0.360	0.971\\
0.400	0.991\\
0.440	1.000\\
0.480	1.004\\
0.520	1.013\\
0.560	1.023\\
0.640	1.042\\
0.680	1.047\\
0.720	1.056\\
0.760	1.064\\
0.800	1.070\\
0.960	1.103\\
1.000	1.110\\
1.040	1.119\\
1.120	1.130\\
1.160	1.140\\
1.200	1.147\\
1.240	1.155\\
1.280	1.161\\
1.320	1.168\\
};
\addplot [color=white!90!black, line width=1.0pt, forget plot]
  table[row sep=crcr]{%
0.000	0.863\\
0.040	0.872\\
0.080	0.881\\
0.280	0.918\\
0.320	0.924\\
0.360	0.934\\
0.400	0.939\\
0.440	0.947\\
0.480	0.954\\
0.520	0.958\\
0.560	0.966\\
0.600	0.970\\
0.640	0.973\\
0.680	0.973\\
0.720	0.976\\
0.760	0.978\\
0.800	0.978\\
0.840	0.980\\
0.880	0.983\\
0.960	0.986\\
1.000	0.991\\
1.040	0.991\\
1.080	0.993\\
1.120	0.996\\
1.160	0.996\\
1.240	1.000\\
};
\addplot [color=white!90!black, line width=1.0pt, forget plot]
  table[row sep=crcr]{%
0.000	0.886\\
0.040	0.901\\
0.120	0.942\\
0.160	0.959\\
0.280	1.009\\
0.320	1.028\\
0.400	1.051\\
0.440	1.072\\
0.480	1.086\\
0.520	1.101\\
0.600	1.127\\
0.640	1.136\\
0.680	1.147\\
0.720	1.154\\
0.760	1.165\\
0.800	1.175\\
0.880	1.193\\
0.920	1.204\\
1.000	1.223\\
1.040	1.231\\
1.080	1.243\\
1.160	1.258\\
1.200	1.267\\
};
\addplot [color=white!90!black, line width=1.0pt, forget plot]
  table[row sep=crcr]{%
0.000	0.859\\
0.040	0.866\\
0.080	0.877\\
0.120	0.885\\
0.160	0.893\\
0.200	0.901\\
0.240	0.910\\
0.280	0.917\\
0.360	0.934\\
0.400	0.941\\
0.440	0.949\\
0.640	0.979\\
0.680	0.986\\
0.800	1.002\\
0.840	1.009\\
0.880	1.014\\
0.920	1.020\\
0.960	1.028\\
1.000	1.033\\
1.120	1.052\\
1.160	1.057\\
1.200	1.060\\
1.240	1.069\\
};
\addplot [color=white!90!black, line width=1.0pt, forget plot]
  table[row sep=crcr]{%
0.000	0.888\\
0.040	0.907\\
0.080	0.923\\
0.120	0.943\\
0.320	1.022\\
0.360	1.035\\
0.400	1.045\\
0.440	1.065\\
0.480	1.081\\
0.560	1.104\\
0.600	1.116\\
0.640	1.125\\
0.680	1.139\\
0.720	1.147\\
0.760	1.157\\
0.800	1.167\\
0.920	1.200\\
0.960	1.209\\
1.000	1.220\\
1.040	1.231\\
1.080	1.242\\
1.200	1.272\\
1.240	1.279\\
};
\addplot [color=white!90!black, line width=1.0pt, forget plot]
  table[row sep=crcr]{%
0.000	0.889\\
0.040	0.909\\
0.120	0.944\\
0.200	0.978\\
0.240	0.988\\
0.280	1.010\\
0.360	1.041\\
0.400	1.058\\
0.480	1.084\\
0.520	1.099\\
0.680	1.153\\
0.720	1.161\\
0.800	1.188\\
0.840	1.198\\
0.880	1.211\\
0.960	1.233\\
1.000	1.247\\
1.040	1.260\\
1.080	1.269\\
1.120	1.280\\
1.160	1.291\\
1.200	1.302\\
};
\addplot [color=white!90!black, line width=1.0pt, forget plot]
  table[row sep=crcr]{%
0.000	0.872\\
0.040	0.886\\
0.160	0.922\\
0.200	0.933\\
0.280	0.959\\
0.320	0.970\\
0.360	0.982\\
0.400	0.993\\
0.440	1.006\\
0.480	1.016\\
0.520	1.027\\
0.600	1.049\\
0.640	1.056\\
0.680	1.056\\
0.720	1.065\\
0.960	1.103\\
1.000	1.110\\
1.080	1.122\\
1.120	1.125\\
1.160	1.133\\
1.200	1.137\\
};
\addplot [color=white!90!black, line width=1.0pt, forget plot]
  table[row sep=crcr]{%
0.000	0.878\\
0.040	0.892\\
0.080	0.905\\
0.120	0.919\\
0.160	0.934\\
0.200	0.947\\
0.240	0.959\\
0.280	0.970\\
0.320	0.980\\
0.360	1.000\\
0.400	1.013\\
0.440	1.014\\
0.480	1.024\\
0.520	1.035\\
0.560	1.045\\
0.600	1.054\\
0.640	1.061\\
0.800	1.082\\
0.920	1.095\\
0.960	1.101\\
1.000	1.108\\
1.160	1.126\\
1.240	1.137\\
1.280	1.139\\
};
\addplot [color=white!90!black, line width=1.0pt, forget plot]
  table[row sep=crcr]{%
0.000	0.858\\
0.040	0.866\\
0.080	0.874\\
0.160	0.889\\
0.240	0.902\\
0.320	0.915\\
0.360	0.921\\
0.400	0.927\\
0.440	0.935\\
0.480	0.941\\
0.520	0.945\\
0.560	0.952\\
0.640	0.961\\
0.680	0.965\\
0.800	0.974\\
0.840	0.977\\
0.880	0.981\\
1.000	0.990\\
1.040	0.995\\
1.080	1.000\\
1.120	1.003\\
1.160	1.008\\
1.200	1.011\\
1.240	1.011\\
};
\addplot [color=white!90!black, line width=1.0pt, forget plot]
  table[row sep=crcr]{%
0.000	0.872\\
0.040	0.883\\
0.080	0.897\\
0.120	0.909\\
0.160	0.921\\
0.200	0.936\\
0.400	0.993\\
0.480	1.015\\
0.520	1.026\\
0.560	1.037\\
0.600	1.047\\
0.640	1.054\\
0.680	1.060\\
0.880	1.090\\
0.960	1.100\\
1.000	1.108\\
1.040	1.114\\
1.080	1.119\\
1.120	1.125\\
1.160	1.129\\
1.240	1.142\\
1.280	1.147\\
};
\addplot [color=white!90!black, line width=1.0pt, forget plot]
  table[row sep=crcr]{%
0.000	0.868\\
0.040	0.878\\
0.080	0.891\\
0.120	0.902\\
0.200	0.928\\
0.240	0.937\\
0.280	0.952\\
0.320	0.963\\
0.360	0.977\\
0.600	1.044\\
0.640	1.052\\
0.680	1.059\\
0.720	1.064\\
0.760	1.072\\
0.840	1.085\\
0.880	1.090\\
0.920	1.097\\
0.960	1.103\\
1.000	1.112\\
1.040	1.116\\
1.080	1.123\\
1.120	1.129\\
1.160	1.133\\
1.200	1.139\\
};
\addplot [color=white!90!black, line width=1.0pt, forget plot]
  table[row sep=crcr]{%
0.000	0.884\\
0.040	0.900\\
0.080	0.916\\
0.120	0.934\\
0.280	0.993\\
0.320	1.008\\
0.360	1.024\\
0.440	1.050\\
0.480	1.062\\
0.520	1.076\\
0.560	1.088\\
0.640	1.107\\
0.720	1.124\\
0.760	1.136\\
0.840	1.155\\
0.880	1.162\\
0.920	1.171\\
0.960	1.182\\
1.040	1.198\\
1.080	1.208\\
1.160	1.223\\
};
\addplot [color=white!90!black, line width=1.0pt, forget plot]
  table[row sep=crcr]{%
0.000	0.864\\
0.040	0.876\\
0.120	0.893\\
0.160	0.904\\
0.320	0.940\\
0.360	0.948\\
0.400	0.957\\
0.480	0.972\\
0.560	0.987\\
0.600	0.992\\
0.680	0.997\\
0.720	1.002\\
0.760	1.005\\
0.800	1.008\\
0.840	1.009\\
0.880	1.013\\
0.960	1.021\\
1.040	1.030\\
1.080	1.032\\
1.120	1.037\\
1.160	1.038\\
1.200	1.042\\
1.240	1.047\\
};
\addplot [color=white!90!black, line width=1.0pt, forget plot]
  table[row sep=crcr]{%
0.000	0.871\\
0.040	0.884\\
0.080	0.896\\
0.160	0.916\\
0.200	0.928\\
0.240	0.937\\
0.280	0.947\\
0.400	0.972\\
0.440	0.982\\
0.520	0.997\\
0.640	1.018\\
0.680	1.023\\
0.720	1.030\\
0.760	1.036\\
0.800	1.044\\
0.840	1.051\\
0.920	1.064\\
1.160	1.108\\
1.200	1.109\\
1.240	1.121\\
1.280	1.127\\
};
\addplot [color=white!90!black, line width=1.0pt, forget plot]
  table[row sep=crcr]{%
0.000	0.884\\
0.040	0.900\\
0.080	0.915\\
0.120	0.933\\
0.160	0.949\\
0.200	0.964\\
0.240	0.978\\
0.280	0.996\\
0.320	1.010\\
0.360	1.025\\
0.440	1.052\\
0.480	1.067\\
0.520	1.080\\
0.680	1.130\\
0.720	1.141\\
0.760	1.154\\
0.800	1.165\\
0.840	1.177\\
0.880	1.188\\
0.920	1.200\\
0.960	1.213\\
1.000	1.225\\
1.040	1.234\\
1.080	1.247\\
1.120	1.256\\
1.160	1.267\\
};
\addplot [color=white!90!black, line width=1.0pt, forget plot]
  table[row sep=crcr]{%
0.000	0.891\\
0.080	0.928\\
0.160	0.963\\
0.200	0.980\\
0.240	0.997\\
0.360	1.040\\
0.400	1.055\\
0.440	1.069\\
0.520	1.093\\
0.560	1.104\\
0.600	1.118\\
0.640	1.132\\
0.680	1.147\\
0.720	1.161\\
0.760	1.174\\
0.800	1.192\\
0.920	1.235\\
0.960	1.251\\
1.040	1.277\\
1.120	1.307\\
1.160	1.318\\
1.200	1.332\\
};
\addplot [color=white!90!black, line width=1.0pt, forget plot]
  table[row sep=crcr]{%
0.000	0.872\\
0.080	0.897\\
0.200	0.930\\
0.240	0.943\\
0.360	0.974\\
0.400	0.983\\
0.440	0.993\\
0.480	1.004\\
0.520	1.014\\
0.640	1.039\\
0.680	1.044\\
0.720	1.052\\
0.760	1.058\\
0.800	1.063\\
0.840	1.070\\
0.920	1.082\\
0.960	1.089\\
1.000	1.093\\
1.040	1.100\\
1.080	1.105\\
1.120	1.110\\
1.160	1.119\\
1.200	1.124\\
};
\addplot [color=white!90!black, line width=1.0pt, forget plot]
  table[row sep=crcr]{%
0.000	0.865\\
0.040	0.873\\
0.080	0.881\\
0.120	0.892\\
0.160	0.899\\
0.200	0.907\\
0.240	0.914\\
0.360	0.936\\
0.440	0.949\\
0.480	0.955\\
0.520	0.962\\
0.560	0.967\\
0.640	0.973\\
0.680	0.977\\
0.720	0.982\\
0.800	0.990\\
0.920	1.000\\
1.000	1.008\\
1.080	1.018\\
1.120	1.020\\
1.160	1.024\\
1.200	1.026\\
1.240	1.030\\
};
\addplot [color=white!90!black, line width=1.0pt, forget plot]
  table[row sep=crcr]{%
0.000	0.872\\
0.040	0.885\\
0.080	0.898\\
0.120	0.909\\
0.160	0.921\\
0.200	0.933\\
0.240	0.946\\
0.320	0.967\\
0.360	0.978\\
0.440	1.000\\
0.480	1.012\\
0.520	1.021\\
0.560	1.031\\
0.640	1.048\\
0.680	1.056\\
0.720	1.066\\
0.800	1.078\\
0.960	1.108\\
1.000	1.117\\
1.040	1.123\\
1.120	1.140\\
1.160	1.146\\
1.200	1.151\\
};
\addplot [color=white!90!black, line width=1.0pt, forget plot]
  table[row sep=crcr]{%
0.000	0.887\\
0.040	0.905\\
0.080	0.922\\
0.120	0.940\\
0.160	0.960\\
0.280	1.009\\
0.320	1.024\\
0.360	1.041\\
0.560	1.114\\
0.600	1.127\\
0.640	1.136\\
0.720	1.162\\
0.760	1.175\\
0.840	1.197\\
0.920	1.216\\
1.000	1.241\\
1.040	1.250\\
1.080	1.259\\
1.120	1.270\\
1.160	1.282\\
1.200	1.290\\
1.240	1.301\\
};
\addplot [color=white!90!black, line width=1.0pt, forget plot]
  table[row sep=crcr]{%
0.000	0.879\\
0.040	0.890\\
0.160	0.929\\
0.200	0.942\\
0.280	0.965\\
0.360	0.987\\
0.400	0.997\\
0.440	1.009\\
0.480	1.018\\
0.520	1.028\\
0.600	1.045\\
0.680	1.058\\
0.720	1.068\\
0.760	1.074\\
0.800	1.082\\
0.840	1.089\\
0.880	1.097\\
0.920	1.103\\
0.960	1.112\\
1.000	1.121\\
1.040	1.127\\
1.080	1.132\\
1.120	1.140\\
1.160	1.147\\
1.200	1.154\\
1.240	1.163\\
};
\addplot [color=white!90!black, line width=1.0pt, forget plot]
  table[row sep=crcr]{%
0.000	0.870\\
0.040	0.882\\
0.120	0.903\\
0.240	0.941\\
0.320	0.963\\
0.360	0.974\\
0.400	0.987\\
0.440	0.998\\
0.600	1.040\\
0.680	1.056\\
0.720	1.062\\
0.760	1.067\\
0.840	1.080\\
0.920	1.092\\
0.960	1.101\\
1.000	1.106\\
1.040	1.117\\
1.080	1.119\\
1.160	1.134\\
1.200	1.140\\
1.240	1.150\\
};
\addplot [color=white!90!black, line width=1.0pt, forget plot]
  table[row sep=crcr]{%
0.000	0.878\\
0.080	0.901\\
0.120	0.914\\
0.200	0.938\\
0.240	0.949\\
0.360	0.978\\
0.400	0.989\\
0.440	0.998\\
0.480	1.005\\
0.520	1.015\\
0.560	1.022\\
0.600	1.029\\
0.640	1.038\\
0.680	1.044\\
0.720	1.053\\
0.760	1.057\\
0.840	1.072\\
0.880	1.078\\
0.960	1.092\\
1.000	1.102\\
1.040	1.109\\
1.120	1.121\\
1.200	1.138\\
};
\addplot [color=white!90!black, line width=1.0pt, forget plot]
  table[row sep=crcr]{%
0.000	0.873\\
0.040	0.884\\
0.080	0.894\\
0.120	0.911\\
0.160	0.926\\
0.240	0.951\\
0.280	0.962\\
0.320	0.975\\
0.360	0.987\\
0.440	1.010\\
0.480	1.019\\
0.520	1.033\\
0.640	1.064\\
0.680	1.069\\
0.720	1.078\\
0.760	1.085\\
0.880	1.105\\
0.920	1.113\\
0.960	1.123\\
1.000	1.128\\
1.040	1.137\\
1.080	1.142\\
1.120	1.152\\
1.160	1.160\\
};
\addplot [color=white!90!black, line width=1.0pt, forget plot]
  table[row sep=crcr]{%
0.000	0.888\\
0.040	0.907\\
0.080	0.927\\
0.200	0.983\\
0.240	1.004\\
0.280	1.021\\
0.320	1.038\\
0.360	1.055\\
0.400	1.075\\
0.440	1.092\\
0.480	1.108\\
0.520	1.122\\
0.560	1.138\\
0.600	1.152\\
0.640	1.169\\
0.680	1.179\\
0.720	1.191\\
0.760	1.204\\
0.800	1.215\\
0.840	1.229\\
0.960	1.266\\
1.000	1.275\\
1.040	1.290\\
1.080	1.302\\
1.120	1.312\\
1.160	1.321\\
};
\addplot [color=white!90!black, line width=1.0pt, forget plot]
  table[row sep=crcr]{%
0.000	0.876\\
0.040	0.891\\
0.080	0.901\\
0.120	0.914\\
0.160	0.924\\
0.240	0.950\\
0.280	0.959\\
0.400	0.992\\
0.440	1.000\\
0.480	1.009\\
0.560	1.027\\
0.680	1.050\\
0.720	1.060\\
0.920	1.098\\
1.160	1.150\\
1.200	1.158\\
};
\addplot [color=white!90!black, line width=1.0pt, forget plot]
  table[row sep=crcr]{%
0.000	0.883\\
0.080	0.911\\
0.120	0.928\\
0.160	0.943\\
0.200	0.957\\
0.280	0.981\\
0.320	0.995\\
0.440	1.030\\
0.480	1.038\\
0.520	1.052\\
0.560	1.062\\
0.680	1.085\\
0.720	1.096\\
0.760	1.105\\
0.880	1.129\\
0.920	1.139\\
0.960	1.146\\
1.040	1.165\\
1.120	1.181\\
1.160	1.188\\
1.200	1.199\\
};
\addplot [color=white!90!black, line width=1.0pt, forget plot]
  table[row sep=crcr]{%
0.000	0.879\\
0.040	0.892\\
0.080	0.908\\
0.120	0.923\\
0.200	0.951\\
0.240	0.966\\
0.280	0.979\\
0.320	0.991\\
0.360	1.005\\
0.520	1.052\\
0.560	1.063\\
0.600	1.073\\
0.640	1.086\\
0.680	1.091\\
0.720	1.098\\
0.840	1.120\\
0.880	1.127\\
0.960	1.143\\
1.000	1.151\\
1.040	1.157\\
1.080	1.164\\
1.120	1.171\\
};
\addplot [color=white!90!black, line width=1.0pt, forget plot]
  table[row sep=crcr]{%
0.000	0.870\\
0.040	0.881\\
0.120	0.905\\
0.160	0.918\\
0.240	0.941\\
0.280	0.950\\
0.320	0.961\\
0.360	0.974\\
0.400	0.984\\
0.440	0.995\\
0.480	1.005\\
0.520	1.014\\
0.560	1.025\\
0.640	1.042\\
0.680	1.052\\
0.720	1.055\\
0.800	1.065\\
0.840	1.070\\
0.880	1.074\\
1.000	1.089\\
1.040	1.093\\
1.160	1.107\\
1.200	1.110\\
};
\addplot [color=white!90!black, line width=1.0pt, forget plot]
  table[row sep=crcr]{%
0.000	0.876\\
0.040	0.894\\
0.160	0.939\\
0.200	0.953\\
0.240	0.966\\
0.280	0.984\\
0.320	0.998\\
0.400	1.024\\
0.480	1.052\\
0.520	1.062\\
0.560	1.077\\
0.600	1.089\\
0.640	1.103\\
0.680	1.111\\
0.720	1.117\\
0.760	1.128\\
0.800	1.138\\
0.840	1.148\\
0.960	1.175\\
1.000	1.185\\
1.080	1.203\\
1.120	1.212\\
1.160	1.220\\
1.240	1.239\\
};
\addplot [color=white!90!black, line width=1.0pt, forget plot]
  table[row sep=crcr]{%
0.000	0.866\\
0.040	0.874\\
0.120	0.894\\
0.240	0.921\\
0.360	0.945\\
0.400	0.952\\
0.440	0.960\\
0.600	0.989\\
0.680	0.995\\
0.720	1.000\\
0.760	1.001\\
0.800	1.005\\
0.920	1.015\\
1.040	1.027\\
1.080	1.027\\
1.120	1.033\\
1.160	1.035\\
1.280	1.046\\
1.320	1.048\\
};
\addplot [color=white!90!black, line width=1.0pt, forget plot]
  table[row sep=crcr]{%
0.000	0.865\\
0.040	0.874\\
0.160	0.907\\
0.200	0.917\\
0.440	0.970\\
0.480	0.979\\
0.520	0.985\\
0.560	0.994\\
0.640	1.008\\
0.720	1.018\\
0.760	1.024\\
0.800	1.029\\
0.880	1.038\\
1.000	1.056\\
1.040	1.060\\
1.160	1.076\\
1.200	1.079\\
1.240	1.084\\
1.320	1.093\\
};
\addplot [color=white!90!black, line width=1.0pt, forget plot]
  table[row sep=crcr]{%
0.000	0.865\\
0.040	0.872\\
0.120	0.890\\
0.160	0.901\\
0.240	0.919\\
0.280	0.926\\
0.320	0.935\\
0.360	0.944\\
0.520	0.974\\
0.560	0.979\\
0.600	0.988\\
0.640	0.989\\
0.720	0.993\\
0.760	0.998\\
0.800	1.001\\
0.840	1.003\\
0.880	1.007\\
0.920	1.010\\
1.000	1.019\\
1.040	1.021\\
1.080	1.025\\
1.120	1.025\\
1.160	1.030\\
1.200	1.032\\
1.240	1.035\\
};
\addplot [color=white!90!black, line width=1.0pt, forget plot]
  table[row sep=crcr]{%
0.000	0.888\\
0.040	0.906\\
0.080	0.927\\
0.200	0.982\\
0.240	0.999\\
0.280	1.018\\
0.320	1.036\\
0.360	1.052\\
0.440	1.081\\
0.520	1.112\\
0.600	1.139\\
0.640	1.149\\
0.680	1.160\\
0.720	1.168\\
0.760	1.179\\
0.800	1.186\\
0.840	1.195\\
0.880	1.205\\
0.920	1.214\\
0.960	1.225\\
1.000	1.233\\
1.040	1.244\\
1.080	1.251\\
1.120	1.261\\
1.280	1.293\\
};
\addplot [color=white!90!black, line width=1.0pt, forget plot]
  table[row sep=crcr]{%
0.000	0.869\\
0.040	0.879\\
0.080	0.892\\
0.120	0.900\\
0.240	0.930\\
0.320	0.946\\
0.520	0.983\\
0.600	0.995\\
0.680	1.001\\
0.720	1.006\\
0.760	1.008\\
0.840	1.016\\
0.920	1.022\\
0.960	1.024\\
1.080	1.036\\
1.120	1.041\\
1.160	1.042\\
1.200	1.045\\
};
\addplot [color=white!90!black, line width=1.0pt, forget plot]
  table[row sep=crcr]{%
0.000	0.882\\
0.040	0.900\\
0.120	0.931\\
0.160	0.950\\
0.280	0.995\\
0.360	1.029\\
0.400	1.042\\
0.440	1.051\\
0.480	1.062\\
0.520	1.077\\
0.600	1.099\\
0.640	1.110\\
0.680	1.121\\
0.760	1.142\\
0.800	1.154\\
0.840	1.163\\
0.960	1.198\\
1.000	1.209\\
1.040	1.213\\
1.080	1.229\\
1.120	1.241\\
1.200	1.259\\
};
\addplot [color=white!90!black, line width=1.0pt, forget plot]
  table[row sep=crcr]{%
0.000	0.885\\
0.040	0.900\\
0.080	0.918\\
0.120	0.935\\
0.200	0.971\\
0.360	1.035\\
0.400	1.051\\
0.480	1.076\\
0.520	1.093\\
0.560	1.106\\
0.600	1.120\\
0.640	1.131\\
0.680	1.138\\
0.720	1.146\\
0.760	1.156\\
0.840	1.172\\
0.920	1.184\\
0.960	1.193\\
1.080	1.213\\
1.120	1.218\\
};
\addplot [color=white!90!black, line width=1.0pt, forget plot]
  table[row sep=crcr]{%
0.000	0.876\\
0.040	0.886\\
0.080	0.902\\
0.240	0.949\\
0.280	0.959\\
0.320	0.969\\
0.360	0.983\\
0.400	0.992\\
0.440	1.002\\
0.480	1.010\\
0.520	1.019\\
0.560	1.029\\
0.640	1.044\\
0.680	1.050\\
0.720	1.057\\
0.800	1.065\\
0.840	1.070\\
0.880	1.076\\
0.960	1.086\\
1.000	1.092\\
1.040	1.096\\
1.080	1.105\\
1.120	1.109\\
1.160	1.114\\
1.200	1.119\\
1.280	1.128\\
};
\addplot [color=white!90!black, line width=1.0pt, forget plot]
  table[row sep=crcr]{%
0.000	0.882\\
0.080	0.913\\
0.120	0.928\\
0.160	0.945\\
0.200	0.960\\
0.240	0.978\\
0.320	1.006\\
0.360	1.023\\
0.400	1.035\\
0.440	1.051\\
0.560	1.087\\
0.600	1.101\\
0.680	1.122\\
0.720	1.131\\
0.760	1.143\\
0.800	1.155\\
0.840	1.164\\
0.880	1.175\\
0.920	1.185\\
0.960	1.198\\
1.000	1.208\\
1.040	1.220\\
1.080	1.231\\
1.120	1.240\\
1.160	1.248\\
1.200	1.258\\
1.240	1.270\\
1.280	1.279\\
};
\addplot [color=white!90!black, line width=1.0pt, forget plot]
  table[row sep=crcr]{%
0.000	0.877\\
0.040	0.892\\
0.120	0.918\\
0.160	0.934\\
0.240	0.960\\
0.320	0.985\\
0.360	0.999\\
0.560	1.056\\
0.600	1.066\\
0.640	1.074\\
0.680	1.078\\
0.760	1.090\\
0.800	1.094\\
0.960	1.114\\
1.000	1.121\\
1.040	1.126\\
1.080	1.128\\
1.120	1.136\\
1.160	1.139\\
};
\addplot [color=white!90!black, line width=1.0pt, forget plot]
  table[row sep=crcr]{%
0.000	0.875\\
0.040	0.886\\
0.080	0.898\\
0.120	0.913\\
0.160	0.925\\
0.200	0.937\\
0.280	0.957\\
0.320	0.969\\
0.360	0.979\\
0.440	0.997\\
0.480	1.003\\
0.520	1.012\\
0.600	1.026\\
0.720	1.038\\
0.760	1.039\\
1.000	1.059\\
1.040	1.063\\
1.080	1.066\\
1.160	1.073\\
1.200	1.074\\
};
\addplot [color=white!90!black, line width=1.0pt, forget plot]
  table[row sep=crcr]{%
0.000	0.881\\
0.080	0.907\\
0.120	0.926\\
0.280	0.980\\
0.320	0.991\\
0.400	1.014\\
0.440	1.026\\
0.480	1.035\\
0.520	1.047\\
0.560	1.056\\
0.600	1.065\\
0.680	1.080\\
0.760	1.100\\
0.880	1.124\\
0.960	1.143\\
1.000	1.151\\
1.040	1.157\\
1.080	1.166\\
1.160	1.183\\
1.240	1.199\\
};
\addplot [color=white!90!black, line width=1.0pt, forget plot]
  table[row sep=crcr]{%
0.000	0.871\\
0.040	0.884\\
0.080	0.898\\
0.120	0.914\\
0.200	0.941\\
0.240	0.954\\
0.280	0.968\\
0.320	0.980\\
0.360	0.992\\
0.400	1.004\\
0.440	1.018\\
0.480	1.028\\
0.520	1.039\\
0.560	1.049\\
0.640	1.070\\
0.680	1.076\\
0.720	1.084\\
0.760	1.090\\
0.840	1.104\\
0.880	1.109\\
0.920	1.115\\
1.000	1.131\\
1.040	1.137\\
1.080	1.145\\
1.120	1.152\\
1.160	1.160\\
};
\addplot [color=white!90!black, line width=1.0pt, forget plot]
  table[row sep=crcr]{%
0.000	0.867\\
0.040	0.878\\
0.080	0.891\\
0.120	0.903\\
0.200	0.925\\
0.240	0.938\\
0.320	0.959\\
0.360	0.968\\
0.400	0.980\\
0.520	1.010\\
0.600	1.030\\
0.640	1.035\\
0.880	1.078\\
0.960	1.091\\
1.080	1.114\\
1.120	1.120\\
1.160	1.128\\
1.200	1.133\\
};
\addplot [color=white!90!black, line width=1.0pt, forget plot]
  table[row sep=crcr]{%
0.000	0.883\\
0.040	0.896\\
0.080	0.916\\
0.120	0.932\\
0.200	0.966\\
0.240	0.981\\
0.320	1.011\\
0.360	1.025\\
0.400	1.042\\
0.560	1.094\\
0.600	1.104\\
0.640	1.112\\
0.680	1.120\\
0.720	1.128\\
0.800	1.148\\
0.880	1.165\\
1.040	1.204\\
1.080	1.210\\
1.120	1.220\\
1.160	1.229\\
1.200	1.239\\
};
\addplot [color=white!90!black, line width=1.0pt, forget plot]
  table[row sep=crcr]{%
0.000	0.868\\
0.040	0.880\\
0.080	0.892\\
0.120	0.901\\
0.160	0.913\\
0.200	0.924\\
0.240	0.933\\
0.280	0.941\\
0.320	0.952\\
0.440	0.978\\
0.520	0.994\\
0.560	1.001\\
0.640	1.015\\
0.680	1.020\\
0.760	1.032\\
0.840	1.041\\
0.880	1.047\\
0.920	1.051\\
0.960	1.059\\
1.000	1.063\\
1.080	1.073\\
1.120	1.076\\
1.160	1.084\\
1.240	1.092\\
1.280	1.096\\
};
\addplot [color=white!90!black, line width=1.0pt, forget plot]
  table[row sep=crcr]{%
0.000	0.880\\
0.040	0.893\\
0.120	0.922\\
0.160	0.936\\
0.200	0.948\\
0.240	0.958\\
0.280	0.974\\
0.320	0.986\\
0.400	1.008\\
0.520	1.039\\
0.600	1.055\\
0.640	1.065\\
0.680	1.074\\
0.720	1.088\\
0.880	1.127\\
0.960	1.150\\
1.000	1.160\\
1.040	1.172\\
1.120	1.193\\
1.160	1.201\\
1.200	1.209\\
};
\addplot [color=white!90!black, line width=1.0pt, forget plot]
  table[row sep=crcr]{%
0.000	0.887\\
0.040	0.905\\
0.080	0.921\\
0.120	0.939\\
0.200	0.969\\
0.240	0.982\\
0.280	0.996\\
0.320	1.011\\
0.360	1.025\\
0.440	1.050\\
0.480	1.063\\
0.560	1.085\\
0.600	1.094\\
0.640	1.101\\
0.720	1.116\\
0.760	1.126\\
0.920	1.153\\
0.960	1.162\\
1.000	1.170\\
1.040	1.177\\
1.080	1.183\\
1.120	1.190\\
1.160	1.195\\
1.200	1.204\\
1.240	1.209\\
};
\addplot [color=white!90!black, line width=1.0pt, forget plot]
  table[row sep=crcr]{%
0.000	0.878\\
0.040	0.889\\
0.080	0.903\\
0.120	0.917\\
0.240	0.952\\
0.280	0.962\\
0.360	0.985\\
0.400	0.994\\
0.440	1.004\\
0.520	1.021\\
0.560	1.030\\
0.600	1.037\\
0.640	1.043\\
0.720	1.049\\
0.760	1.052\\
0.800	1.056\\
0.920	1.066\\
0.960	1.070\\
1.080	1.085\\
1.120	1.085\\
1.160	1.091\\
1.200	1.092\\
1.240	1.098\\
1.280	1.101\\
};
\addplot [color=white!90!black, line width=1.0pt, forget plot]
  table[row sep=crcr]{%
0.000	0.887\\
0.040	0.906\\
0.120	0.941\\
0.160	0.962\\
0.200	0.973\\
0.240	0.995\\
0.280	1.012\\
0.320	1.027\\
0.360	1.043\\
0.400	1.060\\
0.480	1.088\\
0.520	1.102\\
0.560	1.114\\
0.640	1.140\\
0.720	1.161\\
0.760	1.173\\
0.800	1.186\\
0.880	1.209\\
0.920	1.221\\
0.960	1.234\\
1.000	1.244\\
1.040	1.257\\
1.080	1.268\\
1.120	1.278\\
1.160	1.292\\
1.200	1.300\\
1.240	1.311\\
1.280	1.321\\
1.320	1.332\\
1.360	1.343\\
1.400	1.351\\
};
\addplot [color=white!90!black, line width=1.0pt, forget plot]
  table[row sep=crcr]{%
0.000	0.872\\
0.040	0.885\\
0.160	0.924\\
0.200	0.937\\
0.240	0.948\\
0.280	0.959\\
0.320	0.970\\
0.360	0.983\\
0.440	1.005\\
0.480	1.016\\
0.600	1.045\\
0.640	1.052\\
0.680	1.055\\
0.760	1.065\\
0.800	1.068\\
0.880	1.076\\
0.920	1.080\\
0.960	1.085\\
1.000	1.092\\
1.040	1.093\\
1.080	1.097\\
1.120	1.100\\
1.160	1.104\\
1.200	1.107\\
};
\addplot [color=white!90!black, line width=1.0pt, forget plot]
  table[row sep=crcr]{%
0.000	0.895\\
0.040	0.916\\
0.080	0.935\\
0.120	0.959\\
0.160	0.978\\
0.240	1.014\\
0.280	1.033\\
0.320	1.050\\
0.360	1.069\\
0.440	1.100\\
0.480	1.117\\
0.560	1.145\\
0.600	1.157\\
0.640	1.171\\
0.680	1.183\\
0.760	1.210\\
0.840	1.233\\
0.920	1.257\\
0.960	1.270\\
1.040	1.292\\
1.080	1.307\\
1.120	1.315\\
};
\addplot [color=mycolor1]
  table[row sep=crcr]{%
0.000	0.875\\
0.030	0.885\\
0.110	0.913\\
0.180	0.936\\
0.220	0.949\\
0.240	0.955\\
0.260	0.961\\
0.330	0.982\\
0.420	1.008\\
0.470	1.022\\
0.500	1.029\\
0.600	1.054\\
0.640	1.062\\
0.710	1.074\\
0.730	1.077\\
0.790	1.088\\
0.900	1.107\\
0.940	1.114\\
0.990	1.124\\
1.030	1.130\\
1.070	1.137\\
1.100	1.142\\
1.120	1.145\\
1.130	1.144\\
1.150	1.147\\
1.160	1.147\\
1.170	1.145\\
1.180	1.145\\
1.200	1.148\\
1.210	1.147\\
1.220	1.148\\
};
\addlegendentry{mean}

\addplot [color=blue, forget plot]
  table[row sep=crcr]{%
0.000	0.883\\
0.020	0.891\\
0.040	0.899\\
0.150	0.944\\
0.220	0.971\\
0.240	0.979\\
0.260	0.986\\
0.410	1.039\\
0.530	1.078\\
0.600	1.098\\
0.650	1.111\\
0.710	1.124\\
0.720	1.127\\
0.930	1.176\\
0.990	1.191\\
1.090	1.214\\
1.120	1.220\\
1.130	1.218\\
1.150	1.222\\
1.160	1.222\\
1.170	1.220\\
1.180	1.219\\
1.190	1.223\\
1.200	1.225\\
1.210	1.222\\
1.220	1.226\\
};
\addplot [color=blue]
  table[row sep=crcr]{%
0.000	0.868\\
0.050	0.881\\
0.090	0.892\\
0.200	0.921\\
0.460	0.982\\
0.600	1.009\\
0.640	1.015\\
0.960	1.052\\
0.980	1.055\\
1.030	1.061\\
1.110	1.069\\
1.120	1.070\\
1.130	1.070\\
1.150	1.072\\
1.180	1.071\\
1.190	1.070\\
1.210	1.072\\
1.220	1.070\\
};
\addlegendentry{std dev}

\end{axis}
\end{tikzpicture}%

%% file: 3_experiments.tex
\setlength{\figwidth }{0.85\columnwidth} 
\setlength{\figheight }{.309\figwidth} 
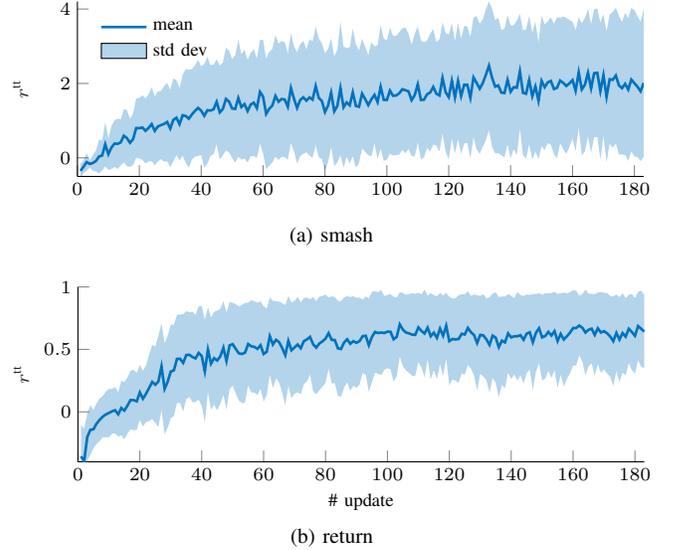
\begin{figure}
    \centering
    \textbf{Learning Curves of Return and Smash Experiments}\par\medskip
    \scriptsize%
    \vspace{-.5cm}
    \subfloat[smash]{
        \centering\scriptsize%
        \input{figures/learncurve/lcsmash}
    }
    \newline
    \subfloat[return]{
        \centering\scriptsize%
        \input{figures/learncurve/lcreturn}
    }
    \caption[Learning curves]{
        Sample mean and standard deviation of the rewards of each episode for the a) smash and b) return task from \sect{ssec:return} and \sect{ssec:smash}. 
        The policies for both experiments were updated 183 times.
        The number of episodes differs for each experiment as we define the end of the training by the number of time steps~(1.5 million for both experiments at \SI{10}{\milli\second}), but the actual episode length varies.
        The return task required 15676 and the smash task 15161 strokes or episodes. 
        Each episode takes approximately \SI{1}{\second} with additional initialization of the robot~($2$ to \SI{4}{\second}, see \fig{sfig:videoprep})
        The total moving time is $\SI{14}{\hour}$ and $\SI{10}{\minute}$ for the return task and $\SI{14}{\hour}$ and $\SI{18}{\minute}$ for the smash experiment.
    }
    \label{fig:learncurve}
\end{figure}
The key contributions of this paper are to 1) enable RL to explore fast motions of soft robots without safety precautions, by doing so, 2) learn a difficult dynamic task using RL with a complex real system.
To show 2), we learn to return and smash a table tennis ball with a PAM-driven robot arm using the HYSR training described in \sect{ssec:simrealtraining} and the reward functions discussed in \sect{ssec:rewardfcns}.
We highlight 1) by quantifying the robustness of the system during the training.
In particular, we illustrate the speed of the returned ball and depict the noisy actions on the low-level controls of the real system due to the application of a stochastic policy.
Results are best seen in the supplemental videos at \texttt{\url{muscularTT.embodied.ml}}.
\subsection{Learning to Return}
\label{ssec:return}
Returning table tennis balls with PAM-driven systems is a challenging problem due to PAMs being hard to model and control~\cite{tondu_modelling_2012,buchler_control_2018} and table tennis requiring precise control of the racket at impact with the ball.
In this task, the robot's task is to return balls to a desired landing position $\bdes{}$~(see \fig{fig:precision}) on the opposite side of the table shot by a ball launcher as shown in \fig{fig:setup} and described in \sect{ssec:task}.
We demonstrate that by enabling RL to explore freely at fast paces, the agent can learn this task well.

We let the robot train for $1.5$ million time-steps using a stochastic policy. 
The policy has been randomly initialized and the actions are changes in target pressures $\Delta\pdes{}$.
One strike corresponds to one episode.
At the end of each episode, the agent receives a reward according to the dense return reward function from \eq{eq:rtt}.
We use PPO~\cite{schulman_proximal_2017} as a backbone RL algorithm.
In particular, we leverage the ppo2 implementation of PPO from OpenAI baseline~\cite{dhariwal_openai_2017}.
\tab{tab:hyps} lists the hyperparameters used for this experiment.

After training, the final policy has been evaluated with real balls.
The agent hits 96\% and returns 77\% of the 107 real balls shot by the ball launcher, as indicated in \tab{tab:rates}.
\fig{fig:precision} illustrates that the landing points spread in a circle around the desired landing point.
This circle overlaps with the opponent's side but is not fully contained by it. 
Thus, the return rate to the table would have been higher if $\bdes{}$ would be moved towards the center of the table half.

\begin{figure*}
    \centering
    \textbf{Visualization of Learned Two Stage Hitting Motion}\par\medskip
    \vspace{-.5cm}
    \subfloat[Initial posture]{
        \centering\scriptsize%
        \includegraphics[scale=0.085]{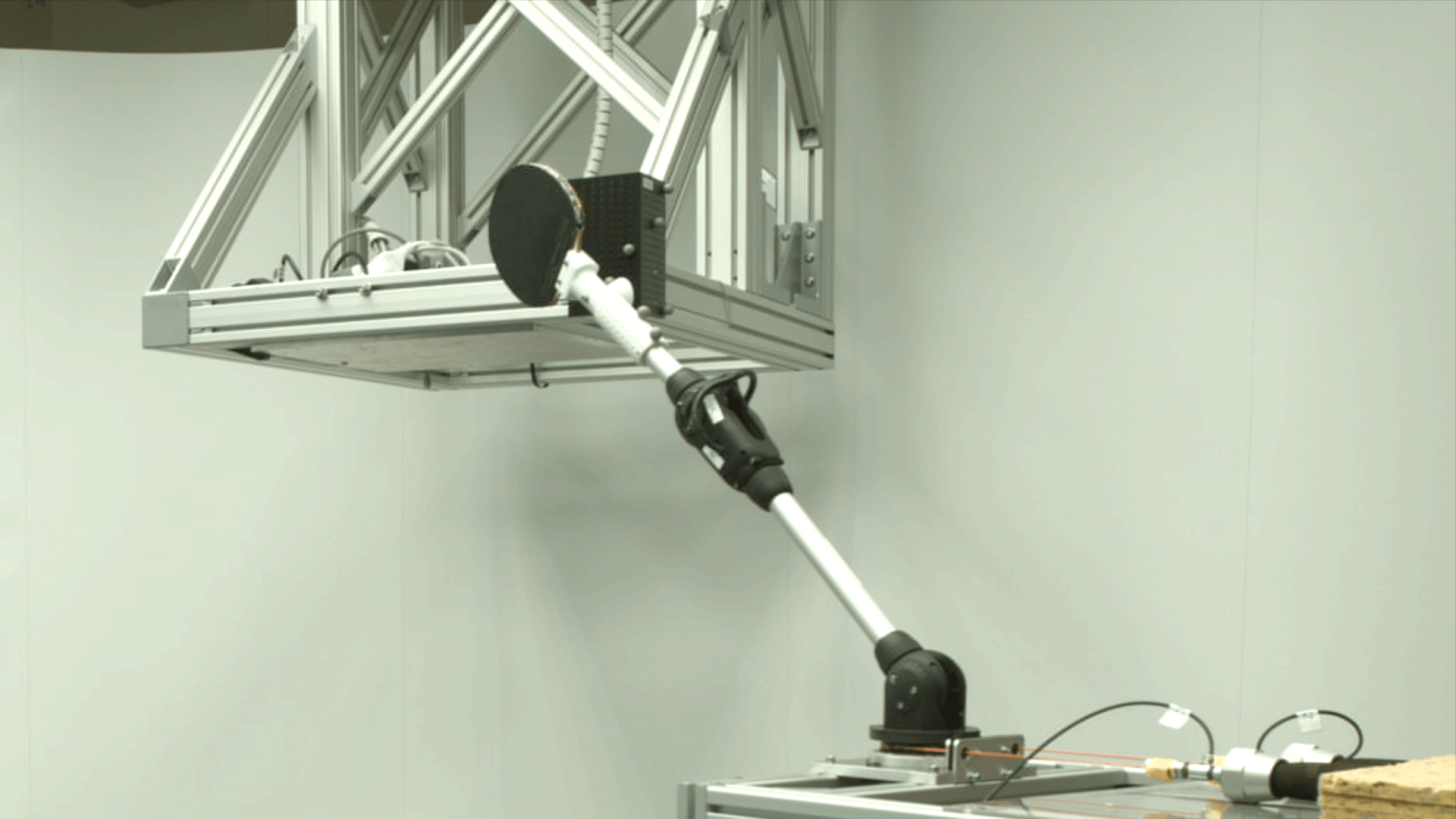}
        \label{sfig:videoprep}
    }
    \subfloat[Preparing for hit]{
        \centering\scriptsize%
        \includegraphics[scale=0.085]{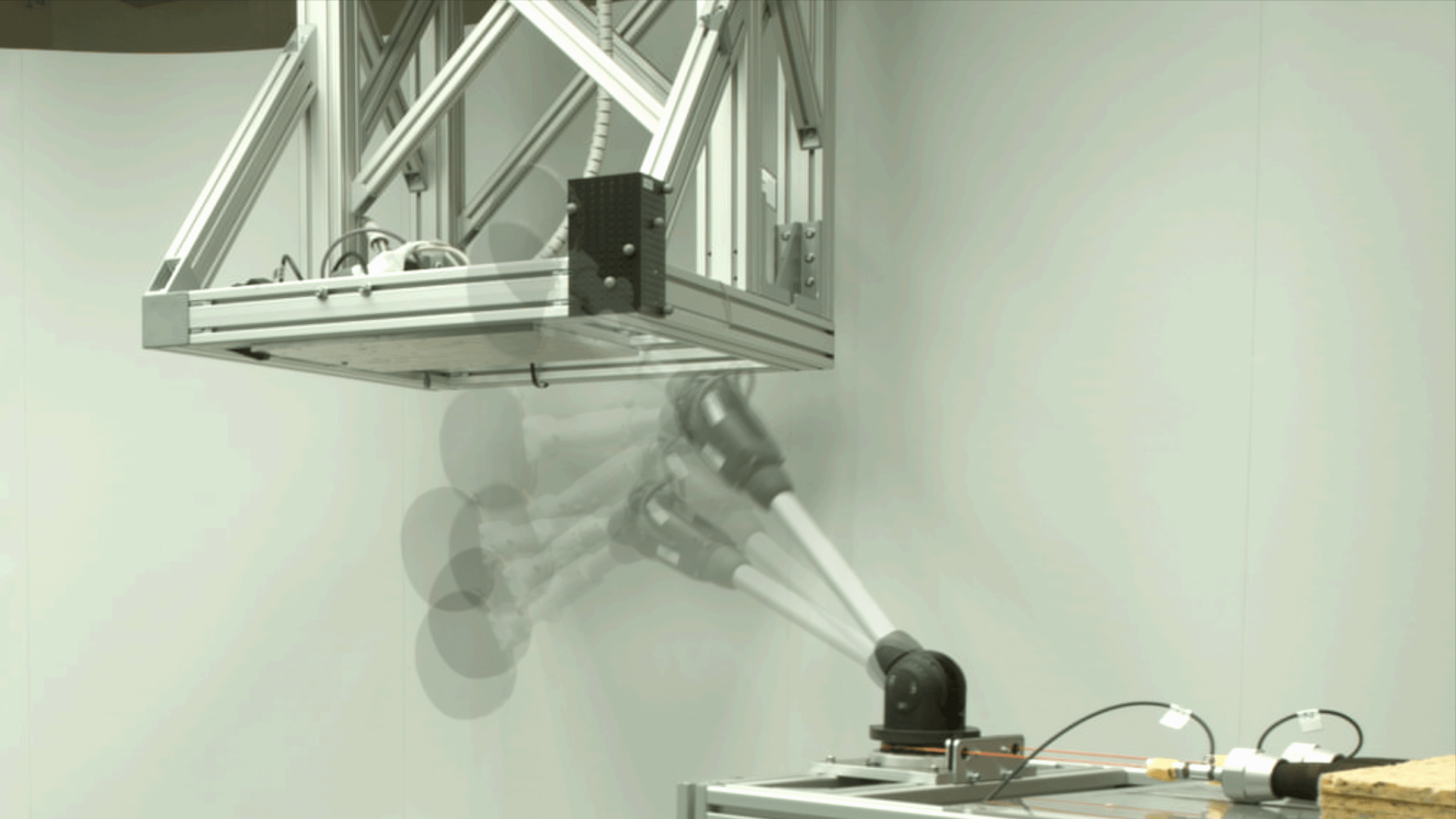}
    }
    \subfloat[Actual hit]{
        \centering\scriptsize%
        \includegraphics[scale=0.085]{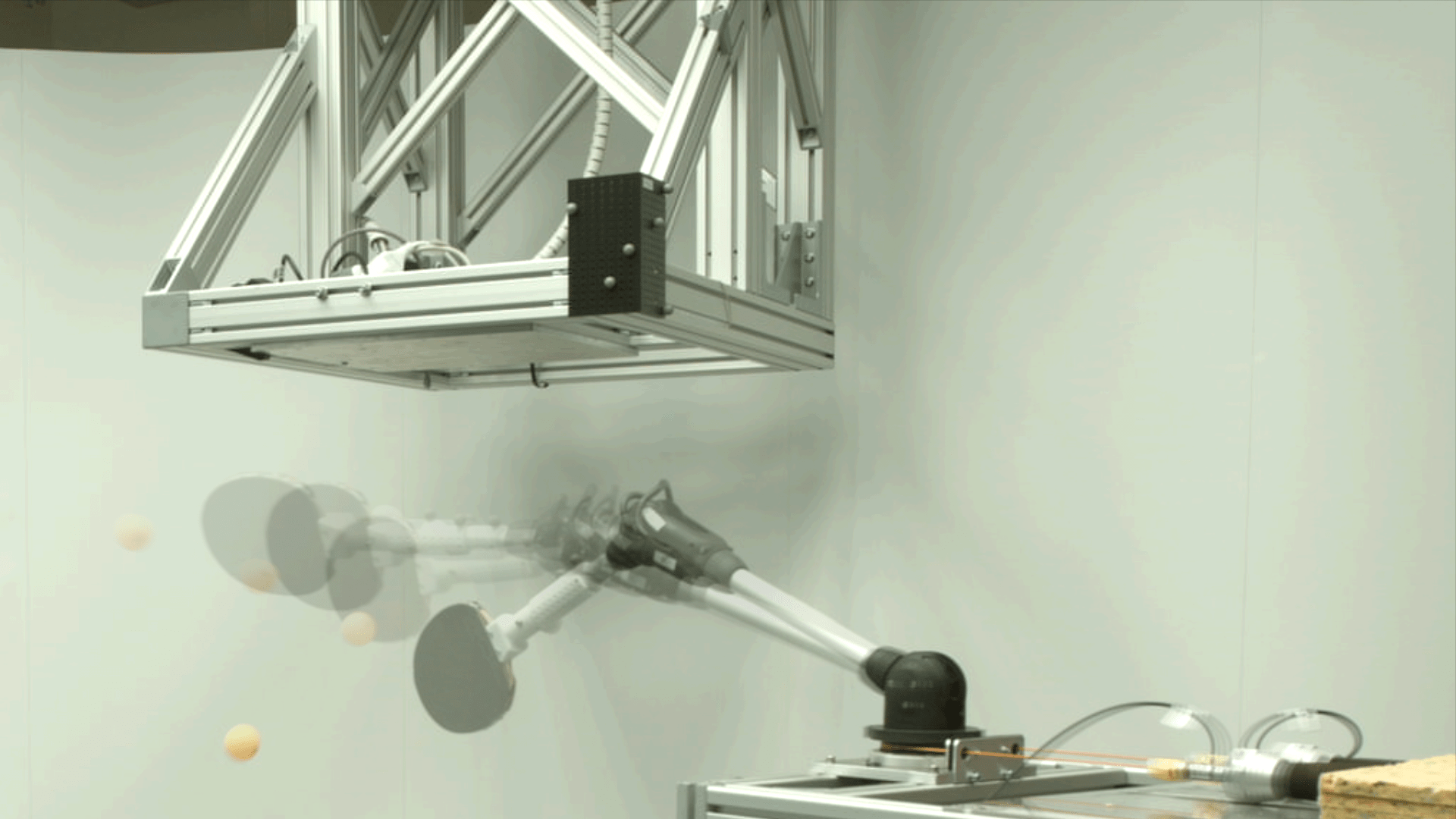}
        \label{sfig:videohit}
    }
    \caption{
        Images extracted from a video showing the learned hitting motion of the return experiment. 
        After an episode, the robot initializes to an initial position shown in a), which takes \SI{2}{\second} to \SI{4}{\second}.
        The agent automatically learned two distinctive hitting stages consisting of b) preparing for a hit and c) the actual hit. 
    }\label{fig:video}
\end{figure*}

Interestingly, the agent did not only learn to intercept the ball but also to prepare for the hit, as can be seen in \fig{fig:video}.
These two stages are part of the four stages of a table tennis game introduced in~\cite{ramanantsoa_towards_1994} and recorded from play in~\cite{mulling_biomimetic_2011}.
This behavior emerged, although the goal was only to return the ball to a desired landing point.
Specifying the same behavior within the classical pipeline of 1) planning a reference trajectory and 2) tracking with an existing (model-based) controller appears to be more difficult:
Such approaches have the disadvantage of relying on human expert knowledge, which is likely to lead to non-optimal solutions considering the challenges of the control problem being solved.
Hence, this work represents a type of end-to-end approach to dynamic tasks where we learn a mapping from sensor information to low-level controls directly.
This kind of approaches are only possible if the hardware is sufficiently robust.

\setlength{\figwidth }{0.9\columnwidth} 
\setlength{\figheight }{.1545\figwidth} 
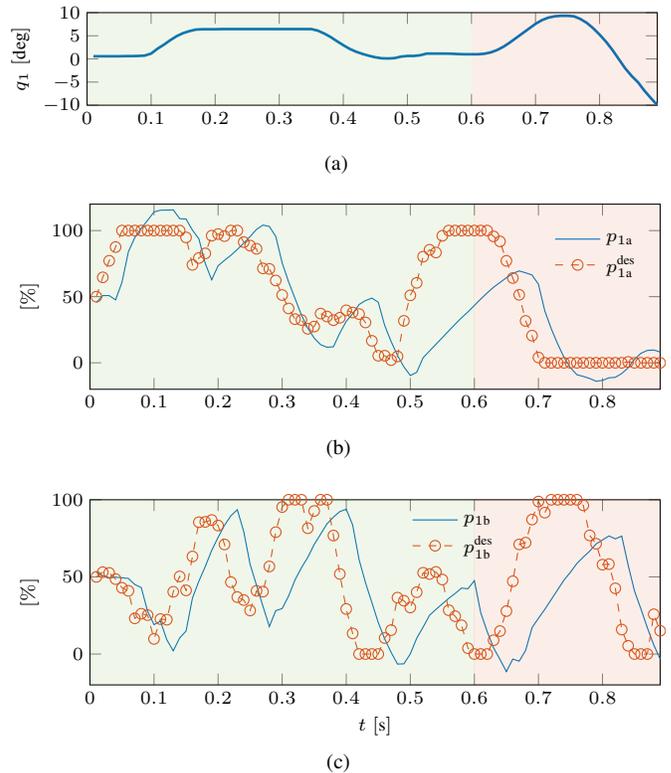
\begin{figure}
    \centering
    \textbf{Noisy Actions of Stochastic Policy Applied Directly to Low Level Controls}\par\medskip
    \vspace{-.25cm}
    \subfloat[]{
        \centering\scriptsize%
        \hspace{-.25cm}
        \input{figures/noisyactions/q1}
        \label{sfig:q1}
    }
    \newline
    \setlength{\figwidth }{0.9\columnwidth} 
    \setlength{\figheight }{.309\figwidth} 
    \subfloat[]{
        \centering\scriptsize%
        \input{figures/noisyactions/pago}
    }
    \newline
    \subfloat[]{
        \centering\scriptsize%
        \input{figures/noisyactions/pantago}
        \label{sfig:antago}
    }
    \caption{
        Noisy actions due to sampling of a stochastic policy in every time step exemplified on the first DoF with both antagonistic PAMs $p_{1\text{a}}$ and $p_{1\text{b}}$ of one episode. 
        The pressures are normalized between $[0\%\ldots100\%]$ to indicate that pressure ranges which map to $[\SI{0}{\bar}\ldots\SI{3}{\bar}]$.
        The low-level controls of the system are the desired pressures $p^\text{des}$ which are set by the RL policy.
        $p_{1\text{a}}$ and $p_{1\text{b}}$ switch along the whole pressure range multiple times during the episode. 
        The impact with the ball happens at $t=\SI{0.9}{\s}$.
        At $t=\SI{0.6}{\s}$ the agent switches the pressures from minimum to maximum and vice versa to hit the ball.
        The green and red background color indicate the two distinct phases of preparation and hit from \fig{fig:video}.
        Applying such action sequences to the low level controls of traditional robots of similar dimensions~(e.g. link lengths) likely breaks the system.
        \fig{fig:detailmotion} visualizes this return and a smash motion in more detail. 
    }\label{fig:stpolicy}
\end{figure}
\setlength{\figwidth }{0.9\columnwidth} 
\setlength{\figheight }{.618\figwidth} 
\begin{figure}
    \centering
    \textbf{Histogram of Maximal Speeds of Returned Balls}\par\medskip
    \scriptsize%
    \input{figures/ballspeed/speedhist}
    \caption{
        Histograms depicting the speeds of the balls for the return and smash experiment. 
        Considered are the maximum velocities after impact with the racket.
        The final policy of each experiment generated the data for the particular histogram.
        One can see that the speeds significantly increase for the smashing experiment.
    }
    \label{fig:speedhist}
\end{figure}
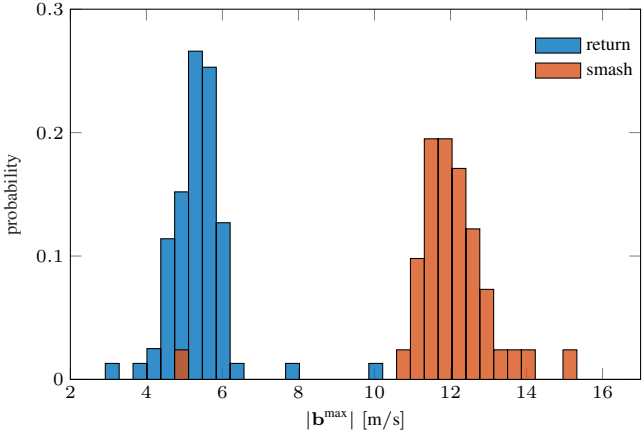
\begin{table}[b]
    \caption{Hyperparameters used for RL experiments}
    \label{tab:hyps}
    \begin{center}
        \begin{tabular}{c|c}
            hyperparameter & value     \\
            \hline
            algorithm & ppo2 \\
            network & mlp \\
            num\_layers & 1 \\
            num\_hidden & 512 \\
            activation & tanh \\
            nsteps & 4096 \\
            ent\_coef & 0.001 \\
            learning\_rate & lambda f:1e-3*f \\
            vf\_coefs & 0.66023 \\
            max\_grad\_norm & 0.05 \\
            gamma & 0.9999 \\
            lam & 0.98438 \\
            nminibatches & 8 \\
            noptepochs & 32 \\
            cliprange & 0.4 
        \end{tabular}
    \end{center}
\end{table}
\setlength{\figwidth }{0.4\columnwidth} 
\setlength{\figheight }{.618\figwidth} 
%
\subsection{Learning to Smash}
\label{ssec:smash}
In table tennis, smashing is a means of maximizing the ball's velocity, such that the opponent has a hard time returning it.
The motion needs to be very fast and, at the same time, precise enough to return the ball to the opponent's side.
Smashing is harder to learn than just returning balls as imprecise motions might lead to the ball not being returned on the table entirely. 
Here we show that, by using soft robots, we can learn this skill using RL only by defining a reward function that maximizes the ball speed and minimizes the distance to the desired landing point~(\eq{eq:rtt}) concurrently.
Rather than taking safety into account algorithmically, we - on the contrary - favor aggressive and explosive motions.

\begin{figure}
    \centering
    \textbf{Precision of Returned Balls}\par\medskip
    \scriptsize%
    \input{figures/return/prec}
    \caption{
        Landing positions of the returned balls for the return as well as for the smash experiment from \sect{ssec:return} and \sect{ssec:smash}.
        The landing points of the smash experiment are further apart from the desired landing point compared to the return experiment as can be seen by the fitted Gaussian distributions. 
        The mean of this distribution for the return experiment coincides with the desired landing point $\bdes{}$ but is shifted for the smash experiment.
        At the same time the smashed balls are faster, hence, corresponding to a more aggressive but risky game play strategy. 
        At such speeds, small error in the racket orientation leads to bigger deviations in landing point.
    }
    \label{fig:precision}
\end{figure}
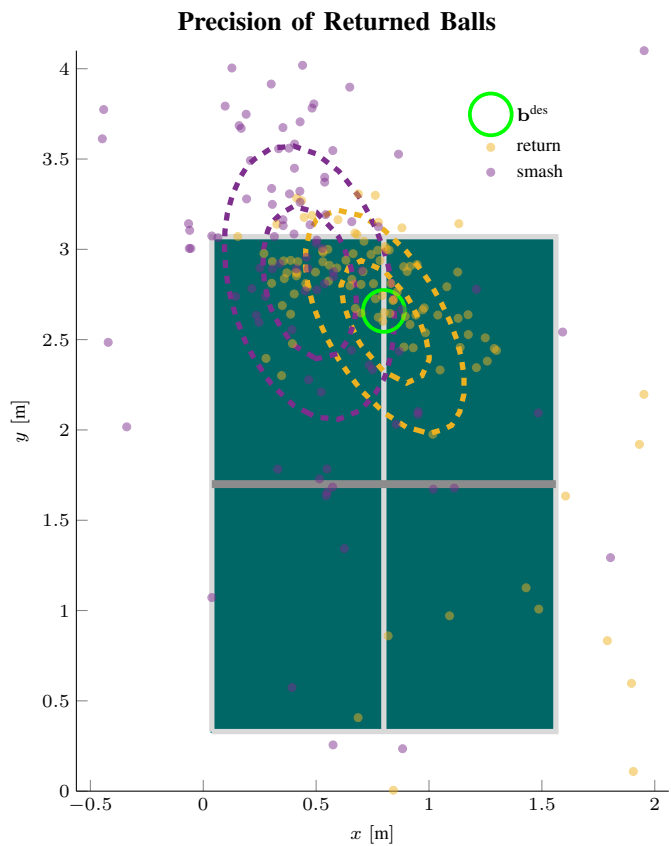

We chose to repeat the experiment from \sect{ssec:return} with the same hyperparameters as in \sect{ssec:return} but updated the reward function with the speed bonus from \eq{eq:rtt}. 
The learning curve is depicted in \fig{fig:learncurve}.
In comparison with the learning curve from \sect{ssec:return}, the smashing task is harder to learn for the agent. 
The standard deviation of the reward for the smash experiment is higher than in the return task. 
Also, the precision of the returned balls is lower in the smash experiment, as shown in \fig{fig:precision}.

\fig{fig:speedhist} shows histograms of the maximum ball velocities after the hit for reward function with and without speed bonus.
The ball speed increases when the reward contains a speed bonus.
For the return task, the robot returns balls at \SI{5}{\meter\per\second} on average, whereas in the smash experiment, this number increases to \SI{12}{\meter\per\second}.
Note that \cite{muelling_learning_2014} considers balls faster than \SI{10}{\meter\per\second} to be smashes in human play. 

The higher return speed comes at a cost: 
The return and hitting rates indicated in \tab{tab:rates} show that the faster the hit, the robot returns the ball less precisely. 
Hence, the more energy is transferred to the ball, the higher the chance of failing.
\begin{table}[b]
    \caption{Return and hitting rates of return~(107 trails) and smash~(128 trails) experiments}
    \label{tab:rates}
    \begin{center}
        \begin{tabular}{c|c|c}
            task & hitting rate & rate of returning to opponents side\\
            \hline
            return & 0.96 & 0.75 \\
            smash & 0.77 & 0.29
        \end{tabular}
    \end{center}
\end{table}
\setlength{\figwidth }{0.9\columnwidth} 
\setlength{\figheight }{1.236\figwidth} 
%
\subsection{Robustness}
The robustness of the PAM-driven robot arm enables the RL algorithm to explore fast motions while executing a stochastic policy directly on the low-level controls.
We quantify the robustness of this PAM-driven system in multiple manners.

First, the sheer number of running a real system for 1.5 million training time-steps stresses that soft robots are indeed robust. 
1.5 million time steps at $\SI{100}{\hertz}$ are equivalent to $\SI{4}{\hour}$ and $\SI{16}{\minute}$ of actual training time.
Also, the robot initializes after each episode, which takes further $2$ to $\SI{4}{\second}$ per episode.
In total, we train the return task for $\SI{14}{\hour}$ and $\SI{18}{\minute}$ and the smash task for $\SI{14}{\hour}$ and $\SI{10}{\minute}$.
Within these durations, the policies of both experiments are updated 183 times, and we perform 15676~(return) and 15161~(smash) strokes, each corresponding to one episode.
Note that the training stopped because the algorithm converged and not due to hardware issues. 

Second, each episode is carried out by sampling actions from a Gaussian multi-layer perceptron~(MLP) at every time-step.
For this reason, the signals are noisy and can vary substantially with each time step.
\fig{fig:stpolicy} depicts the actions $\Delta\pdes{1\text{a}}$ and $\Delta\pdes{1\text{b}}$ of the first DoF and the measured pressures $p_{1\text{a}}$ and $p_{1\text{b}}$ in percent of the allowed pressure ranges alongside with the corresponding joint angle $q_1$.
The hit of the ball happens at $\SI{0.9}{\second}$.
The agent learned to switch the actions from minimum to maximum pressure and vice versa around $t=\SI{0.6}{\second}$ right before the hit.  
In the preparation phase~(see \fig{sfig:videoprep}) before the hit~(indicated by the green background color in \ref{fig:stpolicy}), the agent used the whole pressure range to bring the robot into a beneficial initial state for the hit.
Applying such signals to the low-level controls of a traditional motor-driven system with the same dimensions as our robot arm presumably causes damage or at least severe wear.

Third, in both experiments, we learn from scratch, where the initial policy receives random weights.
Nevertheless, the motions during training did not exceed the joint limits because the allowed pressure ranges have been set such that one of the muscles in the antagonistic pair is stretched close to the respective joint limit~\cite{buchler_lightweight_2016,buchler_learning_2019-1}.
In this manner, the robot can train \emph{without} human supervision.
To achieve safety for dynamic motions on traditional robotic systems, a filter on the actions would be required, such as in~\cite{schwab_simultaneously_2019}.
However, there are multiple downsides to this approach: 
1) Adding a filter makes the state non-Markovian if the internal filter state is not part of the RL state, which would, in turn, increase the RL state's dimensions.
2) Tuning the filter can be tedious: On the one hand, setting the parameters too conservatively leads to slower motions, although the task requires fast motions.
On the other hand, too optimistic parameters might damage the robot for some configurations. 
3) Filtering the stochastic output of the policy counteracts its intended use. 

%% file: figures/learncurve/lcsmash.tex
%
%
\definecolor{mycolor1}{rgb}{0.00000,0.44700,0.74100}%
\begin{tikzpicture}

\begin{axis}[%
width=\figwidth,
height=\figheight,
at={(0\figwidth,0\figheight)},
scale only axis,
xmin=0.000,
xmax=183.000,
ymin=-0.500,
ymax=4.200,
ylabel style={font=\color{white!15!black}},
ylabel={$r^\textrm{tt}$},
axis background/.style={fill=white},
axis x line*=bottom,
axis y line*=left,
legend style={legend cell align=left, align=left, fill=none, draw=none},
ylabel style={at={(-0.075,0.5)}},
xlabel near ticks,
legend style={at={(0.025,.95)},anchor=north west,legend cell align=left,align=left,fill=none,draw=none}
]
\addplot [color=mycolor1, line width=1.0pt]
  table[row sep=crcr]{%
1.000	-0.350\\
2.000	-0.237\\
3.000	-0.113\\
4.000	-0.164\\
5.000	-0.137\\
6.000	-0.083\\
7.000	0.036\\
8.000	0.062\\
9.000	0.363\\
10.000	0.098\\
11.000	0.277\\
12.000	0.384\\
13.000	0.377\\
14.000	0.415\\
15.000	0.595\\
16.000	0.531\\
17.000	0.404\\
18.000	0.503\\
19.000	0.801\\
20.000	0.797\\
21.000	0.817\\
22.000	0.726\\
23.000	0.805\\
24.000	0.907\\
25.000	0.725\\
26.000	0.844\\
27.000	0.879\\
28.000	0.911\\
29.000	0.964\\
30.000	0.787\\
31.000	1.021\\
32.000	1.079\\
33.000	0.905\\
34.000	1.144\\
35.000	1.126\\
36.000	1.052\\
37.000	1.167\\
38.000	1.261\\
39.000	1.344\\
40.000	1.268\\
41.000	1.130\\
42.000	1.280\\
43.000	1.271\\
44.000	1.304\\
45.000	1.506\\
46.000	1.221\\
47.000	1.315\\
48.000	1.583\\
49.000	1.512\\
50.000	1.326\\
51.000	1.315\\
52.000	1.608\\
53.000	1.365\\
54.000	1.445\\
55.000	1.256\\
56.000	1.434\\
57.000	1.482\\
58.000	1.337\\
59.000	1.773\\
60.000	1.646\\
61.000	1.181\\
62.000	1.321\\
63.000	1.405\\
64.000	1.664\\
65.000	1.508\\
66.000	1.582\\
67.000	1.568\\
68.000	1.375\\
69.000	1.877\\
70.000	1.585\\
71.000	1.342\\
72.000	1.555\\
73.000	1.628\\
74.000	1.537\\
75.000	1.624\\
76.000	1.566\\
77.000	1.930\\
78.000	1.611\\
79.000	1.527\\
80.000	1.233\\
81.000	1.405\\
82.000	1.749\\
83.000	1.252\\
84.000	1.396\\
85.000	1.869\\
86.000	1.359\\
87.000	1.491\\
88.000	1.411\\
89.000	1.760\\
90.000	1.394\\
91.000	1.373\\
92.000	1.725\\
93.000	1.578\\
94.000	1.417\\
95.000	1.574\\
96.000	1.693\\
97.000	1.500\\
98.000	1.909\\
99.000	1.558\\
100.000	1.556\\
101.000	1.808\\
102.000	1.669\\
103.000	1.676\\
104.000	1.736\\
105.000	1.844\\
106.000	1.818\\
107.000	1.788\\
108.000	1.550\\
109.000	1.685\\
110.000	1.974\\
111.000	1.556\\
112.000	1.563\\
113.000	1.709\\
114.000	1.756\\
115.000	1.907\\
116.000	1.724\\
117.000	1.677\\
118.000	2.143\\
119.000	1.764\\
120.000	1.632\\
121.000	2.068\\
122.000	1.627\\
123.000	2.065\\
124.000	1.662\\
125.000	1.967\\
126.000	1.966\\
127.000	1.616\\
128.000	2.161\\
129.000	2.018\\
130.000	1.857\\
131.000	2.009\\
132.000	2.234\\
133.000	2.476\\
134.000	2.154\\
135.000	1.915\\
136.000	1.947\\
137.000	1.754\\
138.000	1.743\\
139.000	2.096\\
140.000	1.777\\
141.000	1.899\\
142.000	1.885\\
143.000	2.322\\
144.000	1.748\\
145.000	1.723\\
146.000	2.092\\
147.000	1.657\\
148.000	1.917\\
149.000	1.538\\
150.000	2.070\\
151.000	1.899\\
152.000	2.030\\
153.000	2.068\\
154.000	1.849\\
155.000	1.883\\
156.000	2.078\\
157.000	2.017\\
158.000	1.932\\
159.000	2.002\\
160.000	2.090\\
161.000	1.804\\
162.000	2.297\\
163.000	1.839\\
164.000	1.966\\
165.000	2.285\\
166.000	1.871\\
167.000	2.276\\
168.000	2.287\\
169.000	1.795\\
170.000	2.221\\
171.000	1.642\\
172.000	2.113\\
173.000	2.097\\
174.000	2.213\\
175.000	1.948\\
176.000	1.873\\
177.000	1.989\\
178.000	1.836\\
179.000	2.103\\
180.000	2.006\\
181.000	1.911\\
182.000	1.790\\
183.000	2.014\\
};
\addlegendentry{mean}

\addplot[area legend, draw=none, fill=mycolor1, fill opacity=0.300]
table[row sep=crcr] {%
x	y\\
1.000	-0.540\\
2.000	-0.410\\
3.000	-0.352\\
4.000	-0.264\\
5.000	-0.352\\
6.000	-0.363\\
7.000	-0.421\\
8.000	-0.339\\
9.000	-0.233\\
10.000	-0.323\\
11.000	-0.305\\
12.000	-0.243\\
13.000	-0.301\\
14.000	-0.202\\
15.000	-0.117\\
16.000	-0.227\\
17.000	-0.242\\
18.000	-0.196\\
19.000	0.003\\
20.000	-0.016\\
21.000	-0.052\\
22.000	-0.151\\
23.000	-0.097\\
24.000	-0.087\\
25.000	-0.204\\
26.000	-0.098\\
27.000	-0.038\\
28.000	-0.103\\
29.000	-0.094\\
30.000	-0.252\\
31.000	-0.094\\
32.000	0.068\\
33.000	-0.070\\
34.000	0.057\\
35.000	-0.136\\
36.000	-0.264\\
37.000	-0.050\\
38.000	0.031\\
39.000	0.049\\
40.000	0.074\\
41.000	-0.266\\
42.000	-0.159\\
43.000	0.049\\
44.000	0.099\\
45.000	0.365\\
46.000	-0.043\\
47.000	0.067\\
48.000	0.197\\
49.000	0.144\\
50.000	-0.004\\
51.000	-0.130\\
52.000	0.029\\
53.000	-0.050\\
54.000	-0.021\\
55.000	-0.126\\
56.000	0.042\\
57.000	0.048\\
58.000	-0.001\\
59.000	0.378\\
60.000	0.117\\
61.000	-0.305\\
62.000	-0.214\\
63.000	-0.122\\
64.000	0.027\\
65.000	-0.056\\
66.000	0.026\\
67.000	0.019\\
68.000	-0.057\\
69.000	0.403\\
70.000	0.172\\
71.000	-0.155\\
72.000	0.047\\
73.000	0.110\\
74.000	0.153\\
75.000	0.061\\
76.000	0.002\\
77.000	0.333\\
78.000	0.015\\
79.000	-0.033\\
80.000	-0.353\\
81.000	-0.123\\
82.000	0.291\\
83.000	-0.224\\
84.000	-0.120\\
85.000	0.069\\
86.000	-0.265\\
87.000	0.026\\
88.000	-0.151\\
89.000	0.041\\
90.000	-0.267\\
91.000	-0.144\\
92.000	0.250\\
93.000	0.162\\
94.000	0.070\\
95.000	0.038\\
96.000	0.388\\
97.000	-0.029\\
98.000	0.239\\
99.000	-0.084\\
100.000	-0.049\\
101.000	0.114\\
102.000	0.059\\
103.000	0.074\\
104.000	0.090\\
105.000	0.208\\
106.000	0.123\\
107.000	0.241\\
108.000	0.052\\
109.000	0.035\\
110.000	0.412\\
111.000	-0.045\\
112.000	0.015\\
113.000	0.264\\
114.000	0.171\\
115.000	0.351\\
116.000	0.331\\
117.000	0.164\\
118.000	0.643\\
119.000	0.167\\
120.000	-0.045\\
121.000	0.513\\
122.000	0.113\\
123.000	0.499\\
124.000	0.104\\
125.000	0.386\\
126.000	0.400\\
127.000	-0.020\\
128.000	0.547\\
129.000	0.229\\
130.000	0.105\\
131.000	0.297\\
132.000	0.472\\
133.000	0.743\\
134.000	0.293\\
135.000	0.123\\
136.000	0.126\\
137.000	0.072\\
138.000	0.012\\
139.000	0.358\\
140.000	-0.012\\
141.000	0.232\\
142.000	0.076\\
143.000	0.770\\
144.000	0.069\\
145.000	-0.016\\
146.000	0.361\\
147.000	-0.052\\
148.000	0.173\\
149.000	-0.039\\
150.000	0.314\\
151.000	0.211\\
152.000	0.391\\
153.000	0.355\\
154.000	0.095\\
155.000	-0.011\\
156.000	0.393\\
157.000	0.236\\
158.000	0.102\\
159.000	0.296\\
160.000	0.516\\
161.000	0.169\\
162.000	0.813\\
163.000	0.064\\
164.000	0.174\\
165.000	0.562\\
166.000	0.137\\
167.000	0.555\\
168.000	0.557\\
169.000	-0.048\\
170.000	0.579\\
171.000	-0.124\\
172.000	0.374\\
173.000	0.270\\
174.000	0.386\\
175.000	0.082\\
176.000	0.091\\
177.000	0.079\\
178.000	-0.083\\
179.000	0.176\\
180.000	0.035\\
181.000	0.019\\
182.000	-0.088\\
183.000	0.034\\
183.000	3.994\\
182.000	3.669\\
181.000	3.803\\
180.000	3.977\\
179.000	4.030\\
178.000	3.756\\
177.000	3.899\\
176.000	3.655\\
175.000	3.814\\
174.000	4.041\\
173.000	3.923\\
172.000	3.852\\
171.000	3.407\\
170.000	3.863\\
169.000	3.639\\
168.000	4.017\\
167.000	3.997\\
166.000	3.604\\
165.000	4.008\\
164.000	3.758\\
163.000	3.615\\
162.000	3.780\\
161.000	3.440\\
160.000	3.663\\
159.000	3.708\\
158.000	3.762\\
157.000	3.799\\
156.000	3.762\\
155.000	3.777\\
154.000	3.603\\
153.000	3.780\\
152.000	3.669\\
151.000	3.587\\
150.000	3.827\\
149.000	3.114\\
148.000	3.662\\
147.000	3.366\\
146.000	3.822\\
145.000	3.463\\
144.000	3.426\\
143.000	3.874\\
142.000	3.693\\
141.000	3.565\\
140.000	3.566\\
139.000	3.834\\
138.000	3.475\\
137.000	3.437\\
136.000	3.768\\
135.000	3.707\\
134.000	4.015\\
133.000	4.208\\
132.000	3.996\\
131.000	3.722\\
130.000	3.609\\
129.000	3.808\\
128.000	3.776\\
127.000	3.252\\
126.000	3.531\\
125.000	3.548\\
124.000	3.221\\
123.000	3.631\\
122.000	3.142\\
121.000	3.624\\
120.000	3.310\\
119.000	3.361\\
118.000	3.642\\
117.000	3.189\\
116.000	3.117\\
115.000	3.463\\
114.000	3.342\\
113.000	3.154\\
112.000	3.112\\
111.000	3.156\\
110.000	3.536\\
109.000	3.336\\
108.000	3.049\\
107.000	3.334\\
106.000	3.513\\
105.000	3.481\\
104.000	3.383\\
103.000	3.279\\
102.000	3.279\\
101.000	3.501\\
100.000	3.161\\
99.000	3.199\\
98.000	3.579\\
97.000	3.028\\
96.000	2.997\\
95.000	3.111\\
94.000	2.763\\
93.000	2.994\\
92.000	3.201\\
91.000	2.890\\
90.000	3.054\\
89.000	3.479\\
88.000	2.972\\
87.000	2.955\\
86.000	2.984\\
85.000	3.668\\
84.000	2.912\\
83.000	2.728\\
82.000	3.208\\
81.000	2.933\\
80.000	2.819\\
79.000	3.088\\
78.000	3.207\\
77.000	3.526\\
76.000	3.131\\
75.000	3.186\\
74.000	2.920\\
73.000	3.146\\
72.000	3.063\\
71.000	2.840\\
70.000	2.998\\
69.000	3.351\\
68.000	2.806\\
67.000	3.116\\
66.000	3.138\\
65.000	3.072\\
64.000	3.302\\
63.000	2.932\\
62.000	2.855\\
61.000	2.667\\
60.000	3.174\\
59.000	3.168\\
58.000	2.675\\
57.000	2.917\\
56.000	2.826\\
55.000	2.638\\
54.000	2.910\\
53.000	2.781\\
52.000	3.187\\
51.000	2.760\\
50.000	2.657\\
49.000	2.881\\
48.000	2.969\\
47.000	2.563\\
46.000	2.484\\
45.000	2.647\\
44.000	2.508\\
43.000	2.492\\
42.000	2.719\\
41.000	2.526\\
40.000	2.462\\
39.000	2.638\\
38.000	2.491\\
37.000	2.385\\
36.000	2.368\\
35.000	2.387\\
34.000	2.230\\
33.000	1.880\\
32.000	2.090\\
31.000	2.135\\
30.000	1.827\\
29.000	2.022\\
28.000	1.926\\
27.000	1.797\\
26.000	1.787\\
25.000	1.653\\
24.000	1.902\\
23.000	1.706\\
22.000	1.603\\
21.000	1.686\\
20.000	1.611\\
19.000	1.598\\
18.000	1.201\\
17.000	1.050\\
16.000	1.289\\
15.000	1.308\\
14.000	1.033\\
13.000	1.054\\
12.000	1.010\\
11.000	0.860\\
10.000	0.520\\
9.000	0.958\\
8.000	0.463\\
7.000	0.492\\
6.000	0.198\\
5.000	0.077\\
4.000	-0.065\\
3.000	0.126\\
2.000	-0.063\\
1.000	-0.161\\
}--cycle;
\addlegendentry{std dev}

\end{axis}
\end{tikzpicture}%

%% file: figures/learncurve/lcreturn.tex
%
%
\definecolor{mycolor1}{rgb}{0.00000,0.44700,0.74100}%
\begin{tikzpicture}

\begin{axis}[%
width=\figwidth,
height=\figheight,
at={(0\figwidth,0\figheight)},
scale only axis,
xmin=0.000,
xmax=183.000,
xlabel style={font=\color{white!15!black}},
xlabel={$\textrm{\# update}$},
ymin=-0.400,
ymax=1.000,
ylabel style={font=\color{white!15!black}},
ylabel={$r^\textrm{tt}$},
axis background/.style={fill=white},
axis x line*=bottom,
axis y line*=left,
ylabel style={at={(-0.075,0.5)}},
xlabel near ticks,
legend style={at={(0.1,.9)},anchor=north west,legend cell align=left,align=left,fill=none,draw=none}
]

\addplot[area legend, draw=none, fill=mycolor1, fill opacity=0.300, forget plot]
table[row sep=crcr] {%
x	y\\
1.000	-0.608\\
2.000	-0.655\\
3.000	-0.391\\
4.000	-0.344\\
5.000	-0.294\\
6.000	-0.258\\
7.000	-0.235\\
8.000	-0.204\\
9.000	-0.204\\
10.000	-0.204\\
11.000	-0.175\\
12.000	-0.173\\
13.000	-0.198\\
14.000	-0.188\\
15.000	-0.220\\
16.000	-0.174\\
17.000	-0.140\\
18.000	-0.148\\
19.000	-0.144\\
20.000	-0.065\\
21.000	-0.119\\
22.000	-0.083\\
23.000	-0.074\\
24.000	-0.057\\
25.000	-0.089\\
26.000	-0.122\\
27.000	0.014\\
28.000	-0.181\\
29.000	-0.127\\
30.000	-0.051\\
31.000	-0.034\\
32.000	0.071\\
33.000	0.071\\
34.000	0.002\\
35.000	0.103\\
36.000	0.118\\
37.000	0.077\\
38.000	0.079\\
39.000	0.098\\
40.000	0.047\\
41.000	-0.087\\
42.000	0.158\\
43.000	0.017\\
44.000	0.059\\
45.000	0.012\\
46.000	0.064\\
47.000	0.137\\
48.000	0.017\\
49.000	0.182\\
50.000	0.209\\
51.000	0.144\\
52.000	0.090\\
53.000	0.069\\
54.000	0.076\\
55.000	0.212\\
56.000	0.155\\
57.000	0.154\\
58.000	0.134\\
59.000	0.183\\
60.000	0.081\\
61.000	0.135\\
62.000	0.294\\
63.000	0.189\\
64.000	0.268\\
65.000	0.234\\
66.000	0.153\\
67.000	0.052\\
68.000	0.209\\
69.000	0.108\\
70.000	0.145\\
71.000	0.156\\
72.000	0.140\\
73.000	0.222\\
74.000	0.245\\
75.000	0.298\\
76.000	0.163\\
77.000	0.141\\
78.000	0.171\\
79.000	0.201\\
80.000	0.231\\
81.000	0.343\\
82.000	0.227\\
83.000	0.230\\
84.000	0.128\\
85.000	0.120\\
86.000	0.193\\
87.000	0.236\\
88.000	0.169\\
89.000	0.207\\
90.000	0.193\\
91.000	0.217\\
92.000	0.313\\
93.000	0.296\\
94.000	0.159\\
95.000	0.237\\
96.000	0.314\\
97.000	0.304\\
98.000	0.294\\
99.000	0.335\\
100.000	0.349\\
101.000	0.340\\
102.000	0.241\\
103.000	0.284\\
104.000	0.441\\
105.000	0.364\\
106.000	0.332\\
107.000	0.311\\
108.000	0.288\\
109.000	0.286\\
110.000	0.399\\
111.000	0.355\\
112.000	0.274\\
113.000	0.419\\
114.000	0.258\\
115.000	0.333\\
116.000	0.261\\
117.000	0.381\\
118.000	0.285\\
119.000	0.426\\
120.000	0.211\\
121.000	0.260\\
122.000	0.250\\
123.000	0.232\\
124.000	0.307\\
125.000	0.298\\
126.000	0.376\\
127.000	0.346\\
128.000	0.290\\
129.000	0.169\\
130.000	0.303\\
131.000	0.283\\
132.000	0.253\\
133.000	0.153\\
134.000	0.272\\
135.000	0.231\\
136.000	0.116\\
137.000	0.215\\
138.000	0.190\\
139.000	0.312\\
140.000	0.262\\
141.000	0.299\\
142.000	0.250\\
143.000	0.294\\
144.000	0.226\\
145.000	0.311\\
146.000	0.338\\
147.000	0.171\\
148.000	0.323\\
149.000	0.388\\
150.000	0.252\\
151.000	0.265\\
152.000	0.329\\
153.000	0.270\\
154.000	0.332\\
155.000	0.191\\
156.000	0.298\\
157.000	0.202\\
158.000	0.262\\
159.000	0.308\\
160.000	0.369\\
161.000	0.371\\
162.000	0.452\\
163.000	0.403\\
164.000	0.291\\
165.000	0.365\\
166.000	0.388\\
167.000	0.289\\
168.000	0.381\\
169.000	0.311\\
170.000	0.339\\
171.000	0.307\\
172.000	0.279\\
173.000	0.205\\
174.000	0.326\\
175.000	0.239\\
176.000	0.312\\
177.000	0.292\\
178.000	0.392\\
179.000	0.333\\
180.000	0.307\\
181.000	0.414\\
182.000	0.361\\
183.000	0.348\\
183.000	0.938\\
182.000	0.970\\
181.000	0.961\\
180.000	0.911\\
179.000	0.944\\
178.000	0.937\\
177.000	0.911\\
176.000	0.950\\
175.000	0.919\\
174.000	0.939\\
173.000	0.926\\
172.000	0.927\\
171.000	0.952\\
170.000	0.953\\
169.000	0.960\\
168.000	0.956\\
167.000	0.943\\
166.000	0.944\\
165.000	0.948\\
164.000	0.979\\
163.000	0.943\\
162.000	0.930\\
161.000	0.970\\
160.000	0.974\\
159.000	0.959\\
158.000	0.949\\
157.000	0.955\\
156.000	0.949\\
155.000	0.944\\
154.000	0.961\\
153.000	0.952\\
152.000	0.951\\
151.000	0.949\\
150.000	0.939\\
149.000	0.950\\
148.000	0.942\\
147.000	0.943\\
146.000	0.936\\
145.000	0.941\\
144.000	0.928\\
143.000	0.935\\
142.000	0.960\\
141.000	0.912\\
140.000	0.930\\
139.000	0.944\\
138.000	0.923\\
137.000	0.924\\
136.000	0.920\\
135.000	0.957\\
134.000	0.947\\
133.000	0.918\\
132.000	0.907\\
131.000	0.942\\
130.000	0.929\\
129.000	0.934\\
128.000	0.940\\
127.000	0.929\\
126.000	0.967\\
125.000	0.941\\
124.000	0.944\\
123.000	0.933\\
122.000	0.911\\
121.000	0.907\\
120.000	0.921\\
119.000	0.953\\
118.000	0.932\\
117.000	0.953\\
116.000	0.924\\
115.000	0.935\\
114.000	0.951\\
113.000	0.930\\
112.000	0.979\\
111.000	0.941\\
110.000	0.953\\
109.000	0.946\\
108.000	0.949\\
107.000	0.944\\
106.000	0.937\\
105.000	0.952\\
104.000	0.959\\
103.000	0.916\\
102.000	0.922\\
101.000	0.943\\
100.000	0.936\\
99.000	0.926\\
98.000	0.976\\
97.000	0.958\\
96.000	0.966\\
95.000	0.962\\
94.000	0.910\\
93.000	0.923\\
92.000	0.899\\
91.000	0.915\\
90.000	0.913\\
89.000	0.926\\
88.000	0.920\\
87.000	0.926\\
86.000	0.917\\
85.000	0.886\\
84.000	0.889\\
83.000	0.920\\
82.000	0.925\\
81.000	0.917\\
80.000	0.940\\
79.000	0.927\\
78.000	0.923\\
77.000	0.887\\
76.000	0.919\\
75.000	0.902\\
74.000	0.915\\
73.000	0.911\\
72.000	0.878\\
71.000	0.922\\
70.000	0.887\\
69.000	0.877\\
68.000	0.927\\
67.000	0.856\\
66.000	0.869\\
65.000	0.882\\
64.000	0.887\\
63.000	0.889\\
62.000	0.917\\
61.000	0.827\\
60.000	0.846\\
59.000	0.885\\
58.000	0.875\\
57.000	0.886\\
56.000	0.894\\
55.000	0.874\\
54.000	0.872\\
53.000	0.857\\
52.000	0.845\\
51.000	0.862\\
50.000	0.874\\
49.000	0.881\\
48.000	0.797\\
47.000	0.888\\
46.000	0.824\\
45.000	0.811\\
44.000	0.801\\
43.000	0.756\\
42.000	0.839\\
41.000	0.737\\
40.000	0.847\\
39.000	0.846\\
38.000	0.777\\
37.000	0.802\\
36.000	0.797\\
35.000	0.807\\
34.000	0.782\\
33.000	0.779\\
32.000	0.777\\
31.000	0.698\\
30.000	0.698\\
29.000	0.611\\
28.000	0.549\\
27.000	0.712\\
26.000	0.610\\
25.000	0.523\\
24.000	0.537\\
23.000	0.459\\
22.000	0.372\\
21.000	0.328\\
20.000	0.378\\
19.000	0.313\\
18.000	0.336\\
17.000	0.330\\
16.000	0.275\\
15.000	0.239\\
14.000	0.249\\
13.000	0.158\\
12.000	0.200\\
11.000	0.182\\
10.000	0.186\\
9.000	0.161\\
8.000	0.126\\
7.000	0.108\\
6.000	0.068\\
5.000	0.015\\
4.000	0.056\\
3.000	-0.015\\
2.000	-0.134\\
1.000	-0.105\\
}--cycle;
\addplot [color=mycolor1, line width=1.0pt, forget plot]
  table[row sep=crcr]{%
1.000	-0.356\\
2.000	-0.394\\
3.000	-0.203\\
4.000	-0.144\\
5.000	-0.140\\
6.000	-0.095\\
7.000	-0.064\\
8.000	-0.039\\
9.000	-0.021\\
10.000	-0.009\\
11.000	0.004\\
12.000	0.013\\
13.000	-0.020\\
14.000	0.031\\
15.000	0.010\\
16.000	0.050\\
17.000	0.095\\
18.000	0.094\\
19.000	0.084\\
20.000	0.156\\
21.000	0.105\\
22.000	0.145\\
23.000	0.192\\
24.000	0.240\\
25.000	0.217\\
26.000	0.244\\
27.000	0.363\\
28.000	0.184\\
29.000	0.242\\
30.000	0.323\\
31.000	0.332\\
32.000	0.424\\
33.000	0.425\\
34.000	0.392\\
35.000	0.455\\
36.000	0.457\\
37.000	0.439\\
38.000	0.428\\
39.000	0.472\\
40.000	0.447\\
41.000	0.325\\
42.000	0.498\\
43.000	0.386\\
44.000	0.430\\
45.000	0.411\\
46.000	0.444\\
47.000	0.513\\
48.000	0.407\\
49.000	0.532\\
50.000	0.542\\
51.000	0.503\\
52.000	0.468\\
53.000	0.463\\
54.000	0.474\\
55.000	0.543\\
56.000	0.525\\
57.000	0.520\\
58.000	0.504\\
59.000	0.534\\
60.000	0.463\\
61.000	0.481\\
62.000	0.606\\
63.000	0.539\\
64.000	0.577\\
65.000	0.558\\
66.000	0.511\\
67.000	0.454\\
68.000	0.568\\
69.000	0.493\\
70.000	0.516\\
71.000	0.539\\
72.000	0.509\\
73.000	0.566\\
74.000	0.580\\
75.000	0.600\\
76.000	0.541\\
77.000	0.514\\
78.000	0.547\\
79.000	0.564\\
80.000	0.586\\
81.000	0.630\\
82.000	0.576\\
83.000	0.575\\
84.000	0.508\\
85.000	0.503\\
86.000	0.555\\
87.000	0.581\\
88.000	0.545\\
89.000	0.566\\
90.000	0.553\\
91.000	0.566\\
92.000	0.606\\
93.000	0.609\\
94.000	0.534\\
95.000	0.600\\
96.000	0.640\\
97.000	0.631\\
98.000	0.635\\
99.000	0.631\\
100.000	0.642\\
101.000	0.641\\
102.000	0.581\\
103.000	0.600\\
104.000	0.700\\
105.000	0.658\\
106.000	0.635\\
107.000	0.628\\
108.000	0.618\\
109.000	0.616\\
110.000	0.676\\
111.000	0.648\\
112.000	0.627\\
113.000	0.674\\
114.000	0.605\\
115.000	0.634\\
116.000	0.592\\
117.000	0.667\\
118.000	0.609\\
119.000	0.689\\
120.000	0.566\\
121.000	0.583\\
122.000	0.580\\
123.000	0.583\\
124.000	0.625\\
125.000	0.619\\
126.000	0.672\\
127.000	0.637\\
128.000	0.615\\
129.000	0.551\\
130.000	0.616\\
131.000	0.612\\
132.000	0.580\\
133.000	0.536\\
134.000	0.610\\
135.000	0.594\\
136.000	0.518\\
137.000	0.570\\
138.000	0.556\\
139.000	0.628\\
140.000	0.596\\
141.000	0.605\\
142.000	0.605\\
143.000	0.614\\
144.000	0.577\\
145.000	0.626\\
146.000	0.637\\
147.000	0.557\\
148.000	0.633\\
149.000	0.669\\
150.000	0.596\\
151.000	0.607\\
152.000	0.640\\
153.000	0.611\\
154.000	0.646\\
155.000	0.567\\
156.000	0.624\\
157.000	0.579\\
158.000	0.605\\
159.000	0.634\\
160.000	0.671\\
161.000	0.671\\
162.000	0.691\\
163.000	0.673\\
164.000	0.635\\
165.000	0.657\\
166.000	0.666\\
167.000	0.616\\
168.000	0.669\\
169.000	0.635\\
170.000	0.646\\
171.000	0.629\\
172.000	0.603\\
173.000	0.566\\
174.000	0.632\\
175.000	0.579\\
176.000	0.631\\
177.000	0.602\\
178.000	0.665\\
179.000	0.639\\
180.000	0.609\\
181.000	0.687\\
182.000	0.665\\
183.000	0.643\\
};
\end{axis}
\end{tikzpicture}%

%% file: figures/noisyactions/q1.tex
%
%
\definecolor{mycolor1}{rgb}{0.00000,0.44700,0.74100}%
\definecolor{mycolor2}{rgb}{0.85000,0.32500,0.09800}%
\definecolor{mycolor3}{rgb}{0.46600,0.67400,0.18800}%
\begin{tikzpicture}

\begin{axis}[%
width=0.951\figwidth,
height=\figheight,
at={(0\figwidth,0\figheight)},
scale only axis,
xmin=0,
xmax=0.89,
ymin=-10,
ymax=10,
ylabel style={font=\color{white!15!black}},
ylabel={$q_1~[\textrm{deg}]$},
axis background/.style={fill=white},
ylabel near ticks,
xlabel near ticks,
legend style={at={(1,1)},anchor=north west,legend cell align=left,align=left,fill=none,draw=none}
]
\addplot [color=mycolor1, line width=1.0pt, forget plot]
  table[row sep=crcr]{%
0.00999999999999979	0.57403175967297\\
0.0500000000000007	0.57403175967297\\
0.0600000000000005	0.610322079652295\\
0.0700000000000003	0.610322079652295\\
0.0800000000000001	0.64661239963162\\
0.0899999999999999	0.719193039590269\\
0.0999999999999996	1.15467687934217\\
0.109999999999999	2.2070961587426\\
0.119999999999999	3.0780638382464\\
0.130000000000001	4.05790247768818\\
0.140000000000001	4.92887015719198\\
0.15	5.58209591681983\\
0.16	6.01757975657173\\
0.17	6.34419263638566\\
0.18	6.41677327634431\\
0.199999999999999	6.41677327634431\\
0.210000000000001	6.45306359632363\\
0.35	6.45306359632363\\
0.359999999999999	6.12645071650971\\
0.369999999999999	5.50951527686118\\
0.380000000000001	4.74741855729535\\
0.390000000000001	3.84016055781223\\
0.4	2.96919287830843\\
0.41	2.17080583876327\\
0.42	1.55387039911475\\
0.43	1.08209623938352\\
0.44	0.682902719610945\\
0.449999999999999	0.35628983979702\\
0.460000000000001	0.138547919921068\\
0.470000000000001	0.102257599941744\\
0.48	0.174838239900394\\
0.49	0.50145111971432\\
0.5	0.64661239963162\\
0.51	0.64661239963162\\
0.52	0.791773679548919\\
0.529999999999999	1.15467687934217\\
0.56	1.15467687934217\\
0.57	1.11838655936285\\
0.58	1.0458059194042\\
0.59	1.00951559942487\\
0.609999999999999	1.00951559942487\\
0.619999999999999	1.11838655936285\\
0.630000000000001	1.40870911919745\\
0.640000000000001	1.916773598908\\
0.65	2.6425799984945\\
0.66	3.54983799797763\\
0.699999999999999	7.57806351568271\\
0.710000000000001	8.37645055522786\\
0.720000000000001	8.92080535491774\\
0.73	9.21112791475234\\
0.74	9.31999887469031\\
0.75	9.31999887469031\\
0.76	9.21112791475234\\
0.77	8.63048279508314\\
0.779999999999999	7.79580543555866\\
0.790000000000001	6.63451519622026\\
0.800000000000001	5.25548303700591\\
0.81	3.73128959787425\\
0.82	1.95306391888732\\
0.84	-2.03887127883844\\
0.85	-3.67193567790807\\
0.859999999999999	-5.05096783712242\\
0.869999999999999	-6.86548383608867\\
0.880000000000001	-8.46225791517898\\
0.890000000000001	-9.9501610343313\\
};

\addplot[area legend, draw=none, fill=mycolor2, fill opacity=0.1, forget plot]
table[row sep=crcr] {%
x	y\\
0.6	-10\\
2	-10\\
2	10\\
0.6	10\\
}--cycle;

\addplot[area legend, draw=none, fill=mycolor3, fill opacity=0.1, forget plot]
table[row sep=crcr] {%
x	y\\
0	-10\\
0.6	-10\\
0.6	10\\
0	10\\
}--cycle;
\end{axis}
\end{tikzpicture}%

%% file: figures/noisyactions/pago.tex
%
%
\definecolor{mycolor1}{rgb}{0.00000,0.44700,0.74100}%
\definecolor{mycolor2}{rgb}{0.85000,0.32500,0.09800}%
\definecolor{mycolor3}{rgb}{0.46600,0.67400,0.18800}%
\begin{tikzpicture}

\begin{axis}[%
width=0.951\figwidth,
height=\figheight,
at={(0\figwidth,0\figheight)},
scale only axis,
xmin=0,
xmax=0.89,
ymin=-20,
ymax=120,
ylabel style={font=\color{white!15!black}},
ylabel={$[\%]$},
axis background/.style={fill=white},
legend style={legend cell align=left, align=left, fill=none, draw=none},
ylabel near ticks,
xlabel near ticks,
legend style={at={(.8,.9)},anchor=north west,legend cell align=left,align=left,fill=none,draw=none}
]
\addplot [color=mycolor1]
  table[row sep=crcr]{%
0.0100000000000051	45.6555555555556\\
0.019999999999996	50.8111111111111\\
0.0300000000000011	50.8111111111111\\
0.0400000000000063	47.6\\
0.0499999999999972	60.8444444444444\\
0.0600000000000023	84.0111111111111\\
0.0699999999999932	94.6333333333333\\
0.0799999999999983	101.666666666667\\
0.0900000000000034	107.744444444444\\
0.0999999999999943	114.088888888889\\
0.109999999999999	115.411111111111\\
0.120000000000005	115.466666666667\\
0.129999999999995	115.688888888889\\
0.140000000000001	109.022222222222\\
0.150000000000006	108.677777777778\\
0.159999999999997	100.288888888889\\
0.170000000000002	93.7666666666667\\
0.180000000000007	75.0777777777778\\
0.189999999999998	62.7777777777778\\
0.200000000000003	73.3555555555556\\
0.209999999999994	77.5111111111111\\
0.219999999999999	81.3111111111111\\
0.230000000000004	85.7333333333333\\
0.239999999999995	90.4333333333333\\
0.25	95.5333333333333\\
0.260000000000005	100.911111111111\\
0.269999999999996	104.155555555556\\
0.280000000000001	103.122222222222\\
0.290000000000006	95.5444444444445\\
0.299999999999997	75.5555555555556\\
0.310000000000002	61.2444444444445\\
0.329999999999998	36.2222222222222\\
0.340000000000003	26.1444444444444\\
0.349999999999994	18.7222222222222\\
0.359999999999999	13.7222222222222\\
0.370000000000005	11.7666666666667\\
0.379999999999995	12.1111111111111\\
0.390000000000001	22.5\\
0.400000000000006	31.3777777777778\\
0.409999999999997	37.0555555555556\\
0.420000000000002	44.0888888888889\\
0.430000000000007	47.0333333333333\\
0.439999999999998	48.9333333333333\\
0.450000000000003	45.8\\
0.459999999999994	28.2888888888889\\
0.469999999999999	16.1777777777778\\
0.480000000000004	5.58888888888889\\
0.489999999999995	-3.07777777777778\\
0.5	-9.61111111111111\\
0.510000000000005	-7.2\\
0.519999999999996	3.86666666666666\\
0.569999999999993	29.2111111111111\\
0.579999999999998	34.0888888888889\\
0.590000000000003	38.5\\
0.609999999999999	48.2777777777778\\
0.620000000000005	52.8555555555556\\
0.629999999999995	57.1666666666667\\
0.650000000000006	64.3444444444444\\
0.659999999999997	67.5888888888889\\
0.670000000000002	69.4777777777778\\
0.680000000000007	68.0555555555556\\
0.689999999999998	66.4222222222222\\
0.700000000000003	59.4222222222222\\
0.709999999999994	39.1666666666667\\
0.719999999999999	25.8\\
0.730000000000004	14.3111111111111\\
0.739999999999995	4.06666666666666\\
0.75	-3.77777777777777\\
0.760000000000005	-8.97777777777777\\
0.769999999999996	-10.3\\
0.780000000000001	-11.8555555555556\\
0.790000000000006	-13.9444444444444\\
0.799999999999997	-13.2777777777778\\
0.810000000000002	-11.3555555555556\\
0.819999999999993	-10.9888888888889\\
0.829999999999998	-8.11111111111111\\
0.840000000000003	-3.74444444444444\\
0.849999999999994	2.64444444444445\\
0.859999999999999	7.2\\
0.870000000000005	9.37777777777778\\
0.879999999999995	9.68888888888888\\
0.890000000000001	8.17777777777778\\
};
\addlegendentry{$p_{1\textrm{a}} $}

\addplot [color=mycolor2, dashed, mark=o, mark options={solid, mycolor2}]
  table[row sep=crcr]{%
0.0100000000000051	50\\
0.019999999999996	64.6222222222222\\
0.0300000000000011	77.1444444444444\\
0.0400000000000063	87.5222222222222\\
0.0499999999999972	100\\
0.0600000000000023	100\\
0.0699999999999932	100\\
0.0799999999999983	100\\
0.0900000000000034	100\\
0.0999999999999943	100\\
0.109999999999999	100\\
0.120000000000005	100\\
0.129999999999995	100\\
0.140000000000001	100\\
0.150000000000006	94.8222222222222\\
0.159999999999997	74.0888888888889\\
0.170000000000002	79.2444444444444\\
0.180000000000007	82.7333333333333\\
0.189999999999998	96.1111111111111\\
0.200000000000003	97.3444444444444\\
0.209999999999994	95.8111111111111\\
0.219999999999999	100\\
0.230000000000004	100\\
0.239999999999995	91.2666666666667\\
0.25	88.6\\
0.260000000000005	86.2222222222222\\
0.269999999999996	71.4\\
0.280000000000001	70.9888888888889\\
0.290000000000006	62.2222222222222\\
0.299999999999997	51.2555555555556\\
0.310000000000002	41.1111111111111\\
0.319999999999993	33.0333333333333\\
0.329999999999998	32.3777777777778\\
0.340000000000003	25.7222222222222\\
0.349999999999994	27.4111111111111\\
0.359999999999999	37.2333333333333\\
0.370000000000005	35.0222222222222\\
0.379999999999995	32.2333333333333\\
0.390000000000001	34.0777777777778\\
0.400000000000006	39.6666666666667\\
0.409999999999997	38.0222222222222\\
0.420000000000002	36.8777777777778\\
0.430000000000007	30.5111111111111\\
0.439999999999998	16.6333333333333\\
0.450000000000003	5.35555555555555\\
0.459999999999994	5.36666666666666\\
0.469999999999999	2.05555555555556\\
0.480000000000004	4.83333333333333\\
0.489999999999995	31.7222222222222\\
0.5	51\\
0.510000000000005	60.4777777777778\\
0.519999999999996	80.2666666666667\\
0.530000000000001	85.4666666666667\\
0.540000000000006	83.4111111111111\\
0.549999999999997	96.0555555555556\\
0.560000000000002	100\\
0.569999999999993	100\\
0.579999999999998	100\\
0.590000000000003	100\\
0.599999999999994	100\\
0.609999999999999	100\\
0.620000000000005	100\\
0.629999999999995	95.6\\
0.640000000000001	91.8\\
0.650000000000006	77.1\\
0.659999999999997	64.1111111111111\\
0.670000000000002	51.4222222222222\\
0.680000000000007	31.6666666666667\\
0.689999999999998	20.5444444444445\\
0.700000000000003	4.02222222222223\\
0.709999999999994	0\\
0.719999999999999	0\\
0.730000000000004	0\\
0.739999999999995	0\\
0.75	0\\
0.760000000000005	0\\
0.769999999999996	0\\
0.780000000000001	0\\
0.790000000000006	0\\
0.799999999999997	0\\
0.810000000000002	0\\
0.819999999999993	0\\
0.829999999999998	0\\
0.840000000000003	0.688888888888883\\
0.849999999999994	0\\
0.859999999999999	0\\
0.870000000000005	0\\
0.879999999999995	0\\
0.890000000000001	0\\
};
\addlegendentry{$p^\textrm{des}_{1\textrm{a}}$}

\addplot[area legend, draw=none, fill=mycolor2, fill opacity=0.1, forget plot]
table[row sep=crcr] {%
x	y\\
0.6	-20\\
2	-20\\
2	120\\
0.6	120\\
}--cycle;

\addplot[area legend, draw=none, fill=mycolor3, fill opacity=0.1, forget plot]
table[row sep=crcr] {%
x	y\\
0	-20\\
0.6	-20\\
0.6	120\\
0	120\\
}--cycle;
\end{axis}
\end{tikzpicture}%

%% file: figures/noisyactions/pantago.tex
%
%
\definecolor{mycolor1}{rgb}{0.00000,0.44700,0.74100}%
\definecolor{mycolor2}{rgb}{0.85000,0.32500,0.09800}%
\definecolor{mycolor3}{rgb}{0.46600,0.67400,0.18800}%
\begin{tikzpicture}

\begin{axis}[%
width=0.951\figwidth,
height=\figheight,
at={(0\figwidth,0\figheight)},
scale only axis,
xmin=0,
xmax=0.89,
xlabel style={font=\color{white!15!black}},
xlabel={$t~[\textrm{s}]$},
ymin=-20,
ymax=100,
ylabel style={font=\color{white!15!black}},
ylabel={$[\%]$},
axis background/.style={fill=white},
legend style={legend cell align=left, align=left, fill=none, draw=none},
ylabel near ticks,
xlabel near ticks,
legend style={at={(.55,.98)},anchor=north west,legend cell align=left,align=left,fill=none,draw=none}
]
\addplot [color=mycolor1]
  table[row sep=crcr]{%
0.0100000000000051	53.0555555555556\\
0.019999999999996	50.7333333333333\\
0.0300000000000011	49.4888888888889\\
0.0400000000000063	49.8222222222222\\
0.0600000000000023	49\\
0.0699999999999932	45.0333333333333\\
0.0799999999999983	43.1444444444444\\
0.0900000000000034	27.7\\
0.0999999999999943	19.0222222222222\\
0.109999999999999	21.3666666666667\\
0.120000000000005	9.21111111111111\\
0.129999999999995	2.07777777777778\\
0.140000000000001	10.4111111111111\\
0.150000000000006	14.9\\
0.159999999999997	36.3111111111111\\
0.170000000000002	47.1888888888889\\
0.180000000000007	56.7555555555556\\
0.189999999999998	67.1666666666667\\
0.200000000000003	75.6333333333333\\
0.209999999999994	83.4555555555555\\
0.219999999999999	89.6444444444444\\
0.230000000000004	93.5555555555556\\
0.239999999999995	79.3777777777778\\
0.25	58.0888888888889\\
0.269999999999996	30.9555555555556\\
0.280000000000001	17.8222222222222\\
0.290000000000006	28.1444444444444\\
0.299999999999997	29.4666666666667\\
0.310000000000002	37.5444444444445\\
0.319999999999993	48.0111111111111\\
0.329999999999998	55.8888888888889\\
0.340000000000003	62.6888888888889\\
0.349999999999994	71.0333333333333\\
0.370000000000005	84.3777777777778\\
0.379999999999995	88.7333333333333\\
0.390000000000001	92.3666666666667\\
0.400000000000006	93.8333333333333\\
0.409999999999997	83.7333333333333\\
0.420000000000002	61.3111111111111\\
0.430000000000007	46.3444444444444\\
0.450000000000003	20.4\\
0.459999999999994	9.83333333333333\\
0.469999999999999	1.12222222222222\\
0.480000000000004	-6.51111111111111\\
0.489999999999995	-6.34444444444445\\
0.5	0.0888888888888886\\
0.510000000000005	10.2222222222222\\
0.519999999999996	17.4333333333333\\
0.530000000000001	25.1111111111111\\
0.540000000000006	28.9333333333333\\
0.560000000000002	36.1333333333333\\
0.569999999999993	39.2555555555556\\
0.579999999999998	42.7555555555556\\
0.590000000000003	42.3\\
0.599999999999994	47.5444444444445\\
0.609999999999999	26.7\\
0.620000000000005	14.5333333333333\\
0.629999999999995	3.98888888888889\\
0.640000000000001	-3.53333333333333\\
0.650000000000006	-11.3888888888889\\
0.659999999999997	-3.27777777777777\\
0.670000000000002	-4.62222222222222\\
0.680000000000007	2.35555555555555\\
0.689999999999998	17.0666666666667\\
0.700000000000003	23.1888888888889\\
0.709999999999994	28.9666666666667\\
0.739999999999995	48.0111111111111\\
0.75	53.0555555555556\\
0.760000000000005	58.6888888888889\\
0.769999999999996	62.8222222222222\\
0.780000000000001	66.7222222222222\\
0.790000000000006	71.3444444444444\\
0.810000000000002	76.5333333333333\\
0.819999999999993	74.6555555555556\\
0.829999999999998	76.4444444444444\\
0.840000000000003	57.6555555555556\\
0.849999999999994	40.7444444444445\\
0.859999999999999	29.1555555555556\\
0.870000000000005	17\\
0.879999999999995	5.86666666666666\\
0.890000000000001	-3.22222222222223\\
};
\addlegendentry{$p_{1\textrm{b}} $}

\addplot [color=mycolor2, dashed, mark=o, mark options={solid, mycolor2}]
  table[row sep=crcr]{%
0.0100000000000051	50\\
0.019999999999996	53.0666666666667\\
0.0300000000000011	52.5777777777778\\
0.0400000000000063	48.5666666666667\\
0.0499999999999972	42.9111111111111\\
0.0600000000000023	40.9666666666667\\
0.0699999999999932	23.1111111111111\\
0.0799999999999983	26.0888888888889\\
0.0900000000000034	25.2111111111111\\
0.0999999999999943	9.78888888888889\\
0.109999999999999	22.6444444444444\\
0.120000000000005	22.2666666666667\\
0.129999999999995	40.3444444444444\\
0.140000000000001	50.2777777777778\\
0.150000000000006	41.2555555555556\\
0.159999999999997	63.2333333333333\\
0.170000000000002	85.4333333333333\\
0.180000000000007	85.6555555555556\\
0.189999999999998	86.8666666666667\\
0.200000000000003	83.2888888888889\\
0.209999999999994	71.0222222222222\\
0.219999999999999	46.5888888888889\\
0.230000000000004	36.9\\
0.239999999999995	35\\
0.25	28.2222222222222\\
0.260000000000005	40.9666666666667\\
0.269999999999996	40.4444444444444\\
0.280000000000001	56.7666666666667\\
0.290000000000006	79.0111111111111\\
0.299999999999997	95.2777777777778\\
0.310000000000002	100\\
0.319999999999993	100\\
0.329999999999998	100\\
0.340000000000003	81.5666666666667\\
0.349999999999994	92.7222222222222\\
0.359999999999999	100\\
0.370000000000005	100\\
0.379999999999995	76.8\\
0.390000000000001	51.9\\
0.400000000000006	29.2\\
0.409999999999997	13.3444444444444\\
0.420000000000002	0\\
0.430000000000007	0\\
0.439999999999998	0\\
0.450000000000003	0\\
0.459999999999994	10.3888888888889\\
0.469999999999999	15.4777777777778\\
0.480000000000004	36.5444444444445\\
0.489999999999995	34.5444444444445\\
0.5	30.1555555555556\\
0.510000000000005	40.0888888888889\\
0.519999999999996	52.6666666666667\\
0.530000000000001	51.8555555555556\\
0.540000000000006	53.2222222222222\\
0.549999999999997	48.2777777777778\\
0.560000000000002	28.4\\
0.569999999999993	24.5111111111111\\
0.579999999999998	18.6222222222222\\
0.590000000000003	3.76666666666667\\
0.599999999999994	0\\
0.609999999999999	0\\
0.620000000000005	0\\
0.629999999999995	8.95555555555556\\
0.640000000000001	14.8777777777778\\
0.650000000000006	27.8333333333333\\
0.659999999999997	47.1555555555556\\
0.670000000000002	71.0555555555556\\
0.680000000000007	72.2\\
0.689999999999998	87.1666666666667\\
0.700000000000003	98.8444444444444\\
0.709999999999994	91.6333333333333\\
0.719999999999999	100\\
0.730000000000004	100\\
0.739999999999995	100\\
0.75	100\\
0.760000000000005	100\\
0.769999999999996	96.3666666666667\\
0.780000000000001	76.8888888888889\\
0.790000000000006	71.4888888888889\\
0.799999999999997	57.9222222222222\\
0.810000000000002	58.2777777777778\\
0.819999999999993	42.5888888888889\\
0.829999999999998	15.8777777777778\\
0.840000000000003	5.33333333333333\\
0.849999999999994	0\\
0.859999999999999	0\\
0.870000000000005	0\\
0.879999999999995	25.7666666666667\\
0.890000000000001	15.0555555555556\\
};
\addlegendentry{$p^\textrm{des}_{1\textrm{b}}$}

\addplot[area legend, draw=none, fill=mycolor2, fill opacity=0.1, forget plot]
table[row sep=crcr] {%
x	y\\
0.6	-20\\
2	-20\\
2	100\\
0.6	100\\
}--cycle;

\addplot[area legend, draw=none, fill=mycolor3, fill opacity=0.1, forget plot]
table[row sep=crcr] {%
x	y\\
0	-20\\
0.6	-20\\
0.6	100\\
0	100\\
}--cycle;
\end{axis}
\end{tikzpicture}%

%% file: figures/ballspeed/speedhist.tex
%
%
\definecolor{mycolor1}{rgb}{0.00000,0.44700,0.74100}%
\definecolor{mycolor2}{rgb}{0.85000,0.32500,0.09800}%
\begin{tikzpicture}

\begin{axis}[%
width=0.951\figwidth,
height=\figheight,
at={(0\figwidth,0\figheight)},
scale only axis,
xmin=2.000,
xmax=17.000,
xlabel style={font=\color{white!15!black}},
xlabel={$|\mathbf{b}^\textrm{max}|~[\textrm{m}/\textrm{s}]$},
ymin=0.000,
ymax=0.300,
ylabel style={font=\color{white!15!black}},
ylabel={probability},
axis background/.style={fill=white},
legend style={legend cell align=left, align=left, fill=none, draw=none},
ylabel style={at={(-0.075,0.5)}},
xlabel near ticks,
legend style={at={(0.8,.95)},anchor=north west,legend cell align=left,align=left,fill=none,draw=none}
]
\addplot[ybar interval, fill=mycolor1, fill opacity=0.800, draw=black, area legend] table[row sep=crcr] {%
x	y\\
0.000	0.000\\
0.365	0.000\\
0.730	0.000\\
1.095	0.000\\
1.459	0.000\\
1.824	0.000\\
2.189	0.000\\
2.554	0.000\\
2.919	0.013\\
3.284	0.000\\
3.649	0.013\\
4.014	0.025\\
4.378	0.114\\
4.743	0.152\\
5.108	0.266\\
5.473	0.253\\
5.838	0.127\\
6.203	0.013\\
6.568	0.000\\
6.932	0.000\\
7.297	0.000\\
7.662	0.013\\
8.027	0.000\\
8.392	0.000\\
8.757	0.000\\
9.122	0.000\\
9.486	0.000\\
9.851	0.013\\
10.216	0.000\\
10.581	0.000\\
10.946	0.000\\
11.311	0.000\\
11.676	0.000\\
12.041	0.000\\
12.405	0.000\\
12.770	0.000\\
13.135	0.000\\
13.500	0.000\\
13.865	0.000\\
14.230	0.000\\
14.595	0.000\\
14.959	0.000\\
15.324	0.000\\
15.689	0.000\\
16.054	0.000\\
16.419	0.000\\
16.784	0.000\\
17.149	0.000\\
17.514	0.000\\
17.878	0.000\\
18.243	0.000\\
18.608	0.000\\
18.973	0.000\\
19.338	0.000\\
19.703	0.000\\
20.068	0.000\\
20.432	0.000\\
20.797	0.000\\
21.162	0.000\\
21.527	0.000\\
21.892	0.000\\
22.257	0.000\\
22.622	0.000\\
22.986	0.000\\
23.351	0.000\\
23.716	0.000\\
24.081	0.000\\
24.446	0.000\\
24.811	0.000\\
25.176	0.000\\
25.541	0.000\\
25.905	0.000\\
26.270	0.000\\
26.635	0.000\\
27.000	0.000\\
};
\addlegendentry{return}

\addplot[ybar interval, fill=mycolor2, fill opacity=0.800, draw=black, area legend] table[row sep=crcr] {%
x	y\\
0.000	0.000\\
0.365	0.000\\
0.730	0.000\\
1.095	0.000\\
1.459	0.000\\
1.824	0.000\\
2.189	0.000\\
2.554	0.000\\
2.919	0.000\\
3.284	0.000\\
3.649	0.000\\
4.014	0.000\\
4.378	0.000\\
4.743	0.024\\
5.108	0.000\\
5.473	0.000\\
5.838	0.000\\
6.203	0.000\\
6.568	0.000\\
6.932	0.000\\
7.297	0.000\\
7.662	0.000\\
8.027	0.000\\
8.392	0.000\\
8.757	0.000\\
9.122	0.000\\
9.486	0.000\\
9.851	0.000\\
10.216	0.000\\
10.581	0.024\\
10.946	0.098\\
11.311	0.195\\
11.676	0.195\\
12.041	0.171\\
12.405	0.122\\
12.770	0.073\\
13.135	0.024\\
13.500	0.024\\
13.865	0.024\\
14.230	0.000\\
14.595	0.000\\
14.959	0.024\\
15.324	0.000\\
15.689	0.000\\
16.054	0.000\\
16.419	0.000\\
16.784	0.000\\
17.149	0.000\\
17.514	0.000\\
17.878	0.000\\
18.243	0.000\\
18.608	0.000\\
18.973	0.000\\
19.338	0.000\\
19.703	0.000\\
20.068	0.000\\
20.432	0.000\\
20.797	0.000\\
21.162	0.000\\
21.527	0.000\\
21.892	0.000\\
22.257	0.000\\
22.622	0.000\\
22.986	0.000\\
23.351	0.000\\
23.716	0.000\\
24.081	0.000\\
24.446	0.000\\
24.811	0.000\\
25.176	0.000\\
25.541	0.000\\
25.905	0.000\\
26.270	0.000\\
26.635	0.000\\
27.000	0.000\\
};
\addlegendentry{smash}

\end{axis}
\end{tikzpicture}%

%% file: figures/return/prec.tex
%
%
\definecolor{mycolor1}{rgb}{0.92900,0.69400,0.12500}%
\definecolor{mycolor2}{rgb}{0.49400,0.18400,0.55600}%
\begin{tikzpicture}

\begin{axis}[%
width=0.988\figwidth,
height=\figheight,
at={(0\figwidth,0\figheight)},
scale only axis,
colormap={mymap}{[1pt] rgb(0pt)=(0.2422,0.1504,0.6603); rgb(1pt)=(0.2444,0.1534,0.6728); rgb(2pt)=(0.2464,0.1569,0.6847); rgb(3pt)=(0.2484,0.1607,0.6961); rgb(4pt)=(0.2503,0.1648,0.7071); rgb(5pt)=(0.2522,0.1689,0.7179); rgb(6pt)=(0.254,0.1732,0.7286); rgb(7pt)=(0.2558,0.1773,0.7393); rgb(8pt)=(0.2576,0.1814,0.7501); rgb(9pt)=(0.2594,0.1854,0.761); rgb(11pt)=(0.2628,0.1932,0.7828); rgb(12pt)=(0.2645,0.1972,0.7937); rgb(13pt)=(0.2661,0.2011,0.8043); rgb(14pt)=(0.2676,0.2052,0.8148); rgb(15pt)=(0.2691,0.2094,0.8249); rgb(16pt)=(0.2704,0.2138,0.8346); rgb(17pt)=(0.2717,0.2184,0.8439); rgb(18pt)=(0.2729,0.2231,0.8528); rgb(19pt)=(0.274,0.228,0.8612); rgb(20pt)=(0.2749,0.233,0.8692); rgb(21pt)=(0.2758,0.2382,0.8767); rgb(22pt)=(0.2766,0.2435,0.884); rgb(23pt)=(0.2774,0.2489,0.8908); rgb(24pt)=(0.2781,0.2543,0.8973); rgb(25pt)=(0.2788,0.2598,0.9035); rgb(26pt)=(0.2794,0.2653,0.9094); rgb(27pt)=(0.2798,0.2708,0.915); rgb(28pt)=(0.2802,0.2764,0.9204); rgb(29pt)=(0.2806,0.2819,0.9255); rgb(30pt)=(0.2809,0.2875,0.9305); rgb(31pt)=(0.2811,0.293,0.9352); rgb(32pt)=(0.2813,0.2985,0.9397); rgb(33pt)=(0.2814,0.304,0.9441); rgb(34pt)=(0.2814,0.3095,0.9483); rgb(35pt)=(0.2813,0.315,0.9524); rgb(36pt)=(0.2811,0.3204,0.9563); rgb(37pt)=(0.2809,0.3259,0.96); rgb(38pt)=(0.2807,0.3313,0.9636); rgb(39pt)=(0.2803,0.3367,0.967); rgb(40pt)=(0.2798,0.3421,0.9702); rgb(41pt)=(0.2791,0.3475,0.9733); rgb(42pt)=(0.2784,0.3529,0.9763); rgb(43pt)=(0.2776,0.3583,0.9791); rgb(44pt)=(0.2766,0.3638,0.9817); rgb(45pt)=(0.2754,0.3693,0.984); rgb(46pt)=(0.2741,0.3748,0.9862); rgb(47pt)=(0.2726,0.3804,0.9881); rgb(48pt)=(0.271,0.386,0.9898); rgb(49pt)=(0.2691,0.3916,0.9912); rgb(50pt)=(0.267,0.3973,0.9924); rgb(51pt)=(0.2647,0.403,0.9935); rgb(52pt)=(0.2621,0.4088,0.9946); rgb(53pt)=(0.2591,0.4145,0.9955); rgb(54pt)=(0.2556,0.4203,0.9965); rgb(55pt)=(0.2517,0.4261,0.9974); rgb(56pt)=(0.2473,0.4319,0.9983); rgb(57pt)=(0.2424,0.4378,0.9991); rgb(58pt)=(0.2369,0.4437,0.9996); rgb(59pt)=(0.2311,0.4497,0.9995); rgb(60pt)=(0.225,0.4559,0.9985); rgb(61pt)=(0.2189,0.462,0.9968); rgb(62pt)=(0.2128,0.4682,0.9948); rgb(63pt)=(0.2066,0.4743,0.9926); rgb(64pt)=(0.2006,0.4803,0.9906); rgb(65pt)=(0.195,0.4861,0.9887); rgb(66pt)=(0.1903,0.4919,0.9867); rgb(67pt)=(0.1869,0.4975,0.9844); rgb(68pt)=(0.1847,0.503,0.9819); rgb(69pt)=(0.1831,0.5084,0.9793); rgb(70pt)=(0.1818,0.5138,0.9766); rgb(71pt)=(0.1806,0.5191,0.9738); rgb(72pt)=(0.1795,0.5244,0.9709); rgb(73pt)=(0.1785,0.5296,0.9677); rgb(74pt)=(0.1778,0.5349,0.9641); rgb(75pt)=(0.1773,0.5401,0.9602); rgb(76pt)=(0.1768,0.5452,0.956); rgb(77pt)=(0.1764,0.5504,0.9516); rgb(78pt)=(0.1755,0.5554,0.9473); rgb(79pt)=(0.174,0.5605,0.9432); rgb(80pt)=(0.1716,0.5655,0.9393); rgb(81pt)=(0.1686,0.5705,0.9357); rgb(82pt)=(0.1649,0.5755,0.9323); rgb(83pt)=(0.161,0.5805,0.9289); rgb(84pt)=(0.1573,0.5854,0.9254); rgb(85pt)=(0.154,0.5902,0.9218); rgb(86pt)=(0.1513,0.595,0.9182); rgb(87pt)=(0.1492,0.5997,0.9147); rgb(88pt)=(0.1475,0.6043,0.9113); rgb(89pt)=(0.1461,0.6089,0.908); rgb(90pt)=(0.1446,0.6135,0.905); rgb(91pt)=(0.1429,0.618,0.9022); rgb(92pt)=(0.1408,0.6226,0.8998); rgb(93pt)=(0.1383,0.6272,0.8975); rgb(94pt)=(0.1354,0.6317,0.8953); rgb(95pt)=(0.1321,0.6363,0.8932); rgb(96pt)=(0.1288,0.6408,0.891); rgb(97pt)=(0.1253,0.6453,0.8887); rgb(98pt)=(0.1219,0.6497,0.8862); rgb(99pt)=(0.1185,0.6541,0.8834); rgb(100pt)=(0.1152,0.6584,0.8804); rgb(101pt)=(0.1119,0.6627,0.877); rgb(102pt)=(0.1085,0.6669,0.8734); rgb(103pt)=(0.1048,0.671,0.8695); rgb(104pt)=(0.1009,0.675,0.8653); rgb(105pt)=(0.0964,0.6789,0.8609); rgb(106pt)=(0.0914,0.6828,0.8562); rgb(107pt)=(0.0855,0.6865,0.8513); rgb(108pt)=(0.0789,0.6902,0.8462); rgb(109pt)=(0.0713,0.6938,0.8409); rgb(110pt)=(0.0628,0.6972,0.8355); rgb(111pt)=(0.0535,0.7006,0.8299); rgb(112pt)=(0.0433,0.7039,0.8242); rgb(113pt)=(0.0328,0.7071,0.8183); rgb(114pt)=(0.0234,0.7103,0.8124); rgb(115pt)=(0.0155,0.7133,0.8064); rgb(116pt)=(0.0091,0.7163,0.8003); rgb(117pt)=(0.0046,0.7192,0.7941); rgb(118pt)=(0.0019,0.722,0.7878); rgb(119pt)=(0.0009,0.7248,0.7815); rgb(120pt)=(0.0018,0.7275,0.7752); rgb(121pt)=(0.0046,0.7301,0.7688); rgb(122pt)=(0.0094,0.7327,0.7623); rgb(123pt)=(0.0162,0.7352,0.7558); rgb(124pt)=(0.0253,0.7376,0.7492); rgb(125pt)=(0.0369,0.74,0.7426); rgb(126pt)=(0.0504,0.7423,0.7359); rgb(127pt)=(0.0638,0.7446,0.7292); rgb(128pt)=(0.077,0.7468,0.7224); rgb(129pt)=(0.0899,0.7489,0.7156); rgb(130pt)=(0.1023,0.751,0.7088); rgb(131pt)=(0.1141,0.7531,0.7019); rgb(132pt)=(0.1252,0.7552,0.695); rgb(133pt)=(0.1354,0.7572,0.6881); rgb(134pt)=(0.1448,0.7593,0.6812); rgb(135pt)=(0.1532,0.7614,0.6741); rgb(136pt)=(0.1609,0.7635,0.6671); rgb(137pt)=(0.1678,0.7656,0.6599); rgb(138pt)=(0.1741,0.7678,0.6527); rgb(139pt)=(0.1799,0.7699,0.6454); rgb(140pt)=(0.1853,0.7721,0.6379); rgb(141pt)=(0.1905,0.7743,0.6303); rgb(142pt)=(0.1954,0.7765,0.6225); rgb(143pt)=(0.2003,0.7787,0.6146); rgb(144pt)=(0.2061,0.7808,0.6065); rgb(145pt)=(0.2118,0.7828,0.5983); rgb(146pt)=(0.2178,0.7849,0.5899); rgb(147pt)=(0.2244,0.7869,0.5813); rgb(148pt)=(0.2318,0.7887,0.5725); rgb(149pt)=(0.2401,0.7905,0.5636); rgb(150pt)=(0.2491,0.7922,0.5546); rgb(151pt)=(0.2589,0.7937,0.5454); rgb(152pt)=(0.2695,0.7951,0.536); rgb(153pt)=(0.2809,0.7964,0.5266); rgb(154pt)=(0.2929,0.7975,0.517); rgb(155pt)=(0.3052,0.7985,0.5074); rgb(156pt)=(0.3176,0.7994,0.4975); rgb(157pt)=(0.3301,0.8002,0.4876); rgb(158pt)=(0.3424,0.8009,0.4774); rgb(159pt)=(0.3548,0.8016,0.4669); rgb(160pt)=(0.3671,0.8021,0.4563); rgb(161pt)=(0.3795,0.8026,0.4454); rgb(162pt)=(0.3921,0.8029,0.4344); rgb(163pt)=(0.405,0.8031,0.4233); rgb(164pt)=(0.4184,0.803,0.4122); rgb(165pt)=(0.4322,0.8028,0.4013); rgb(166pt)=(0.4463,0.8024,0.3904); rgb(167pt)=(0.4608,0.8018,0.3797); rgb(168pt)=(0.4753,0.8011,0.3691); rgb(169pt)=(0.4899,0.8002,0.3586); rgb(170pt)=(0.5044,0.7993,0.348); rgb(171pt)=(0.5187,0.7982,0.3374); rgb(172pt)=(0.5329,0.797,0.3267); rgb(173pt)=(0.547,0.7957,0.3159); rgb(175pt)=(0.5748,0.7929,0.2941); rgb(176pt)=(0.5886,0.7913,0.2833); rgb(177pt)=(0.6024,0.7896,0.2726); rgb(178pt)=(0.6161,0.7878,0.2622); rgb(179pt)=(0.6297,0.7859,0.2521); rgb(180pt)=(0.6433,0.7839,0.2423); rgb(181pt)=(0.6567,0.7818,0.2329); rgb(182pt)=(0.6701,0.7796,0.2239); rgb(183pt)=(0.6833,0.7773,0.2155); rgb(184pt)=(0.6963,0.775,0.2075); rgb(185pt)=(0.7091,0.7727,0.1998); rgb(186pt)=(0.7218,0.7703,0.1924); rgb(187pt)=(0.7344,0.7679,0.1852); rgb(188pt)=(0.7468,0.7654,0.1782); rgb(189pt)=(0.759,0.7629,0.1717); rgb(190pt)=(0.771,0.7604,0.1658); rgb(191pt)=(0.7829,0.7579,0.1608); rgb(192pt)=(0.7945,0.7554,0.157); rgb(193pt)=(0.806,0.7529,0.1546); rgb(194pt)=(0.8172,0.7505,0.1535); rgb(195pt)=(0.8281,0.7481,0.1536); rgb(196pt)=(0.8389,0.7457,0.1546); rgb(197pt)=(0.8495,0.7435,0.1564); rgb(198pt)=(0.86,0.7413,0.1587); rgb(199pt)=(0.8703,0.7392,0.1615); rgb(200pt)=(0.8804,0.7372,0.165); rgb(201pt)=(0.8903,0.7353,0.1695); rgb(202pt)=(0.9,0.7336,0.1749); rgb(203pt)=(0.9093,0.7321,0.1815); rgb(204pt)=(0.9184,0.7308,0.189); rgb(205pt)=(0.9272,0.7298,0.1973); rgb(206pt)=(0.9357,0.729,0.2061); rgb(207pt)=(0.944,0.7285,0.2151); rgb(208pt)=(0.9523,0.7284,0.2237); rgb(209pt)=(0.9606,0.7285,0.2312); rgb(210pt)=(0.9689,0.7292,0.2373); rgb(211pt)=(0.977,0.7304,0.2418); rgb(212pt)=(0.9842,0.733,0.2446); rgb(213pt)=(0.99,0.7365,0.2429); rgb(214pt)=(0.9946,0.7407,0.2394); rgb(215pt)=(0.9966,0.7458,0.2351); rgb(216pt)=(0.9971,0.7513,0.2309); rgb(217pt)=(0.9972,0.7569,0.2267); rgb(218pt)=(0.9971,0.7626,0.2224); rgb(219pt)=(0.9969,0.7683,0.2181); rgb(220pt)=(0.9966,0.774,0.2138); rgb(221pt)=(0.9962,0.7798,0.2095); rgb(222pt)=(0.9957,0.7856,0.2053); rgb(223pt)=(0.9949,0.7915,0.2012); rgb(224pt)=(0.9938,0.7974,0.1974); rgb(225pt)=(0.9923,0.8034,0.1939); rgb(226pt)=(0.9906,0.8095,0.1906); rgb(227pt)=(0.9885,0.8156,0.1875); rgb(228pt)=(0.9861,0.8218,0.1846); rgb(229pt)=(0.9835,0.828,0.1817); rgb(230pt)=(0.9807,0.8342,0.1787); rgb(231pt)=(0.9778,0.8404,0.1757); rgb(232pt)=(0.9748,0.8467,0.1726); rgb(233pt)=(0.972,0.8529,0.1695); rgb(234pt)=(0.9694,0.8591,0.1665); rgb(235pt)=(0.9671,0.8654,0.1636); rgb(236pt)=(0.9651,0.8716,0.1608); rgb(237pt)=(0.9634,0.8778,0.1582); rgb(238pt)=(0.9619,0.884,0.1557); rgb(239pt)=(0.9608,0.8902,0.1532); rgb(240pt)=(0.9601,0.8963,0.1507); rgb(241pt)=(0.9596,0.9023,0.148); rgb(242pt)=(0.9595,0.9084,0.145); rgb(243pt)=(0.9597,0.9143,0.1418); rgb(244pt)=(0.9601,0.9203,0.1382); rgb(245pt)=(0.9608,0.9262,0.1344); rgb(246pt)=(0.9618,0.932,0.1304); rgb(247pt)=(0.9629,0.9379,0.1261); rgb(248pt)=(0.9642,0.9437,0.1216); rgb(249pt)=(0.9657,0.9494,0.1168); rgb(250pt)=(0.9674,0.9552,0.1116); rgb(251pt)=(0.9692,0.9609,0.1061); rgb(252pt)=(0.9711,0.9667,0.1001); rgb(253pt)=(0.973,0.9724,0.0938); rgb(254pt)=(0.9749,0.9782,0.0872); rgb(255pt)=(0.9769,0.9839,0.0805)},
xmin=-0.562,
xmax=2.062,
xlabel style={font=\color{white!15!black}},
xlabel={$x~[\textrm{m}]$},
ymin=0.000,
ymax=4.100,
ylabel style={font=\color{white!15!black}},
ylabel={$y~[\textrm{m}]$},
axis background/.style={fill=white},
axis x line*=bottom,
axis y line*=left,
legend style={legend cell align=left, align=left, fill=none, draw=none},
ylabel near ticks,
xlabel near ticks,
legend style={at={(0.65,.95)},anchor=north west,legend cell align=left,align=left,fill=none,draw=none}
]
\draw[line width=1.0pt, fill=black!20!teal, draw=teal] (axis cs:0.038,0.330) rectangle (axis cs:1.562,3.070);
\addplot [color=white!85!black, line width=2.0pt, forget plot]
  table[row sep=crcr]{%
0.038	0.330\\
0.038	3.070\\
1.562	3.070\\
1.562	0.330\\
0.038	0.330\\
};
\addplot [color=white!85!black, line width=2.0pt, forget plot]
  table[row sep=crcr]{%
0.800	0.330\\
0.800	3.070\\
};
\addplot [color=white!55!black, line width=3.0pt, forget plot]
  table[row sep=crcr]{%
0.038	1.700\\
1.562	1.700\\
};
\addplot[only marks, mark=o, mark options={}, mark size=7.9057pt, draw=green,line width=.5mm] table[row sep=crcr]{%
x	y\\
0.800	2.659\\
};
\addlegendentry{$\mathbf{b}^\textrm{des}$}

\addplot[only marks, mark=*, mark options={}, mark size=1.5000pt, draw=mycolor1, fill=mycolor1, opacity=0.5] table[row sep=crcr]{%
x	y\\
1.249	2.510\\
1.039	2.636\\
1.790	0.833\\
0.880	3.119\\
1.486	1.008\\
1.017	1.976\\
0.278	2.396\\
0.743	2.972\\
0.708	2.877\\
0.597	3.164\\
0.370	2.870\\
0.687	3.307\\
0.552	3.000\\
0.776	3.150\\
1.091	0.971\\
0.413	2.853\\
0.153	3.071\\
1.138	2.457\\
0.414	3.284\\
1.952	2.197\\
0.681	3.082\\
0.387	2.939\\
0.888	2.904\\
0.632	2.940\\
0.942	2.636\\
0.530	2.931\\
0.930	2.455\\
0.754	2.810\\
0.447	3.177\\
1.173	2.638\\
1.127	2.874\\
0.674	2.673\\
0.699	2.647\\
0.979	2.676\\
0.806	3.015\\
0.859	2.764\\
0.347	2.301\\
1.258	2.381\\
1.132	3.142\\
0.972	2.665\\
0.813	3.000\\
0.410	2.874\\
0.403	2.753\\
0.387	2.641\\
0.650	2.798\\
0.658	2.834\\
0.699	2.806\\
0.461	2.792\\
0.359	2.838\\
0.395	2.478\\
0.916	2.906\\
0.436	3.269\\
1.289	2.451\\
0.672	3.090\\
0.793	2.744\\
1.210	2.345\\
0.419	2.932\\
0.783	2.938\\
0.592	2.897\\
1.174	2.528\\
1.897	0.597\\
0.916	2.584\\
0.454	2.811\\
0.824	2.996\\
1.932	1.920\\
0.895	2.629\\
0.326	3.139\\
0.889	2.459\\
0.827	2.645\\
0.531	2.983\\
1.129	2.618\\
0.903	2.877\\
0.796	2.602\\
0.303	2.812\\
0.416	2.868\\
0.306	2.932\\
0.762	3.299\\
0.843	2.720\\
0.775	2.624\\
0.763	2.726\\
1.049	2.332\\
0.773	2.995\\
1.299	2.440\\
0.559	2.912\\
0.802	2.650\\
0.356	2.936\\
0.840	2.717\\
1.905	0.109\\
0.984	2.538\\
0.350	2.689\\
0.528	2.992\\
0.497	2.874\\
0.269	2.977\\
0.664	2.976\\
0.712	3.045\\
0.555	2.824\\
1.605	1.634\\
0.947	2.916\\
0.831	2.811\\
0.482	3.188\\
1.430	1.126\\
1.079	2.793\\
0.686	0.407\\
0.842	0.005\\
0.819	0.860\\
};
\addlegendentry{return}

\addplot[only marks, mark=*, mark options={}, mark size=1.5000pt, draw=mycolor2, fill=mycolor2,opacity=.5] table[row sep=crcr]{%
x	y\\
1.804	1.293\\
0.776	3.125\\
0.440	4.019\\
0.334	3.557\\
0.542	3.192\\
0.353	3.164\\
0.482	3.782\\
0.545	1.635\\
0.868	2.435\\
0.467	2.278\\
0.549	1.656\\
0.143	2.739\\
0.538	2.810\\
1.112	1.678\\
0.854	2.032\\
0.393	2.555\\
0.464	2.893\\
0.405	3.583\\
0.573	1.684\\
0.500	3.034\\
0.038	3.074\\
0.339	2.921\\
0.380	3.560\\
1.592	2.542\\
0.428	3.322\\
0.952	2.102\\
0.951	2.086\\
0.490	3.805\\
1.484	2.094\\
0.190	3.748\\
0.159	3.685\\
0.168	3.670\\
0.127	4.004\\
0.574	3.547\\
1.953	4.100\\
0.306	2.892\\
0.503	2.776\\
0.515	1.729\\
0.235	2.635\\
0.413	2.892\\
0.404	3.449\\
-0.062	3.005\\
-0.065	3.142\\
0.382	3.307\\
0.473	3.136\\
-0.060	3.105\\
0.625	1.343\\
0.315	3.071\\
0.735	2.359\\
0.656	2.441\\
0.038	1.072\\
0.452	2.770\\
0.306	3.249\\
0.883	0.235\\
-0.339	2.017\\
0.577	2.849\\
0.247	2.596\\
0.575	0.256\\
0.251	2.897\\
1.019	1.672\\
0.528	3.005\\
0.301	3.915\\
-0.447	3.612\\
0.097	3.793\\
0.865	3.527\\
-0.055	3.005\\
0.522	2.210\\
0.402	3.085\\
0.303	3.337\\
0.065	3.064\\
0.649	3.898\\
0.353	3.674\\
0.538	3.372\\
0.577	2.877\\
0.209	3.492\\
-0.421	2.485\\
0.354	3.131\\
0.503	3.051\\
0.192	3.279\\
0.429	3.260\\
0.215	2.783\\
0.330	1.783\\
0.702	2.882\\
0.457	2.956\\
0.750	2.334\\
0.275	2.739\\
0.548	1.784\\
0.538	3.399\\
0.880	2.667\\
0.393	0.574\\
-0.441	3.774\\
1.210	2.779\\
0.658	3.153\\
0.429	3.706\\
};
\addlegendentry{smash}

\addplot[contour prepared, contour prepared format=matlab, contour/labels=false, dashed, line width=2.0pt, contour/draw color=mycolor1] table[row sep=crcr] {%
0.497	41.000\\
0.946	2.000\\
1.000	1.980\\
1.057	2.000\\
1.100	2.025\\
1.136	2.100\\
1.155	2.200\\
1.155	2.300\\
1.143	2.400\\
1.123	2.500\\
1.100	2.577\\
1.093	2.600\\
1.056	2.700\\
1.011	2.800\\
1.000	2.822\\
0.956	2.900\\
0.900	2.988\\
0.891	3.000\\
0.807	3.100\\
0.800	3.108\\
0.700	3.186\\
0.651	3.200\\
0.600	3.216\\
0.558	3.200\\
0.500	3.161\\
0.472	3.100\\
0.454	3.000\\
0.453	2.900\\
0.464	2.800\\
0.484	2.700\\
0.500	2.646\\
0.514	2.600\\
0.551	2.500\\
0.595	2.400\\
0.600	2.390\\
0.650	2.300\\
0.700	2.219\\
0.715	2.200\\
0.797	2.100\\
0.800	2.096\\
0.900	2.015\\
0.946	2.000\\
0.669	23.000\\
0.837	2.300\\
0.900	2.259\\
0.969	2.300\\
1.000	2.388\\
1.001	2.400\\
1.000	2.413\\
0.988	2.500\\
0.959	2.600\\
0.921	2.700\\
0.900	2.744\\
0.860	2.800\\
0.800	2.882\\
0.767	2.900\\
0.700	2.938\\
0.644	2.900\\
0.610	2.800\\
0.621	2.700\\
0.648	2.600\\
0.684	2.500\\
0.700	2.467\\
0.746	2.400\\
0.800	2.323\\
0.837	2.300\\
};
\addplot[contour prepared, contour prepared format=matlab, contour/labels=false, dashed, line width=2.0pt, contour/draw color=mycolor2] table[row sep=crcr] {%
0.324	47.000\\
0.455	2.100\\
0.500	2.071\\
0.600	2.059\\
0.681	2.100\\
0.700	2.112\\
0.761	2.200\\
0.800	2.282\\
0.806	2.300\\
0.829	2.400\\
0.842	2.500\\
0.847	2.600\\
0.845	2.700\\
0.836	2.800\\
0.821	2.900\\
0.800	2.995\\
0.799	3.000\\
0.769	3.100\\
0.732	3.200\\
0.700	3.272\\
0.685	3.300\\
0.623	3.400\\
0.600	3.433\\
0.531	3.500\\
0.500	3.529\\
0.400	3.572\\
0.300	3.559\\
0.228	3.500\\
0.200	3.470\\
0.164	3.400\\
0.129	3.300\\
0.107	3.200\\
0.100	3.140\\
0.096	3.100\\
0.094	3.000\\
0.099	2.900\\
0.100	2.888\\
0.110	2.800\\
0.128	2.700\\
0.152	2.600\\
0.183	2.500\\
0.200	2.455\\
0.224	2.400\\
0.275	2.300\\
0.300	2.257\\
0.345	2.200\\
0.400	2.136\\
0.455	2.100\\
0.437	27.000\\
0.492	2.400\\
0.500	2.395\\
0.525	2.400\\
0.600	2.424\\
0.641	2.500\\
0.670	2.600\\
0.679	2.700\\
0.673	2.800\\
0.654	2.900\\
0.626	3.000\\
0.600	3.068\\
0.580	3.100\\
0.505	3.200\\
0.500	3.206\\
0.400	3.238\\
0.347	3.200\\
0.300	3.145\\
0.285	3.100\\
0.267	3.000\\
0.264	2.900\\
0.273	2.800\\
0.292	2.700\\
0.300	2.671\\
0.329	2.600\\
0.381	2.500\\
0.400	2.470\\
0.492	2.400\\
};
\end{axis}
\end{tikzpicture}%

%% file: 4_conclusion.tex
Accurately returning and smashing table tennis balls to a desired landing point with anthropomorphic robots requires the execution of precise \emph{and} fast motions.
Exploration at fast speeds is highly desirable for improving accuracy but might damage the robot, e.g., exceeding joint limits. 
The synergy between soft robots and Reinforcement Learning for dynamic tasks resolves this dilemma. 
The robustness of soft actuation allows RL to act like in simulation and, vice versa, RL helps to overcome the difficulties of the table tennis tasks and the control problems of soft robots.
A particularly interesting finding is that in our experiments, it was neither necessary to take safety into account in the algorithm nor was any model or demonstration needed.
On the contrary: We even encouraged returning the ball with high speed by an additional term in the reward function resulting in explosive hitting motions while learning from scratch for millions of time steps. 
To make training more practical, we introduce HYSR, a hybrid sim and real training procedure that allows the robot to learn from thousands of strokes without the need to touch any real balls during training. 
With these choices, this paper is the first 1) to achieve sufficiently accurate motions for the precision demanding task of table tennis with PAM-driven soft robots, 2) to enable fail-safe learning of the safety-critical task of smashing real balls directly on a real robot using model-free RL \emph{from scratch} and 3) to learn robot table tennis without using real balls during training. 

In future work, we aim at improving the sample-efficiency of our training. 
HYSR, for instance, can be extended with data-augmentation techniques or by prioritizing the replayed ball trajectory using curriculum learning.
Although not the focus of this paper, it is worth noting that our approach to learning table tennis is less sample-efficient compared to previous robot table tennis approaches.
More efficient training would enable us to improve table tennis performance. 
Here, the most crucial objective will be to perfect the precision of returned balls.
Subsequently, we aim at extending the task itself, such as serving balls, playing fore- and backhand strikes, or two robot play.
It is important to mention that the current version of the real PAM-driven system still suffers from severe non-linear friction and stiction effects due to the cable drive. 
Improving in this aspect will additionally lead to better performance.

We believe that this paper is a step towards highlighting the beneficial synergy of soft robots and learning approaches, which is not widely explored yet.
For this purpose, we open-source the data collected throughout the experiments.
These rich data sets are suitable for benchmarking dynamics models as they contain a variety of motions at different speeds. We present the data sets, the videos of the full training, and a summarizing video at \texttt{muscularTT.embodied.ml}.

%% file: 6_appendices.tex
\section{Visualization of a Return and Smash}
\begin{figure*}
    \centering
    \textbf{~~~~~~~~~~~~Return~~~~~~~~~~~~~~~~~~~~~~~~~~~~~~~~~~~~~~~~~~~~~~~~~~~~~~~~~~~Smash}\par\medskip
    \vspace{-.5cm}
    \subfloat{
        \centering\scriptsize%
        \hspace{-.2cm}
        \input{figures/appendix/q1}
        \label{sfig:aq1}
    }
    \subfloat{
        \centering\scriptsize%
        \hspace{.1cm}
        \input{figures/appendix/q1smash}
        \label{sfig:aq1samsh}
    }
    \vspace{-.5cm}
    \newline
    \subfloat{
        \centering\scriptsize%
        \hspace{-.15cm}
        \input{figures/appendix/q2}
        \label{sfig:aq2}
    }
    \subfloat{
        \centering\scriptsize%
        \hspace{-.1cm}
        \input{figures/appendix/q2smash}
        \label{sfig:aq2smash}
    }
    \newline
    \vspace{-1cm}
    \subfloat{
        \centering\scriptsize%
        \hspace{-.3cm}
        \input{figures/appendix/q3}
        \label{sfig:aq3}
    }
    \subfloat{
        \centering\scriptsize%
        \input{figures/appendix/q3smash}
        \label{sfig:aq3smash}
    }
    \vspace{-.4cm}
    \newline
    \subfloat{
        \centering\scriptsize%
        \hspace{.005cm}
        \input{figures/appendix/q4}
        \label{sfig:aq4}
    }
    \subfloat{
        \centering\scriptsize%
        \hspace{.1cm}
        \input{figures/appendix/q4smash}
        \label{sfig:aq4}
    }
    \newline
    \vspace{-.75cm}
    \subfloat{
        \centering\scriptsize%
        \input{figures/appendix/pago1}
    }
    \subfloat{
        \centering\scriptsize%
        \input{figures/appendix/pago1smash}
    }    
    \newline
    \vspace{-.75cm}
    \subfloat{
        \centering\scriptsize%
        \input{figures/appendix/pantago1}
        \label{sfig:antago1}
    }
    \subfloat{
        \centering\scriptsize%
        \input{figures/appendix/pantago1smash}
        \label{sfig:antago1}
    }
    \newline
    \vspace{-.75cm}
    \subfloat{
        \centering\scriptsize%
        \input{figures/appendix/pago2}
    }
    \subfloat{
        \centering\scriptsize%
        \input{figures/appendix/pago2smash}
    }
    \newline
    \vspace{-.75cm}
    \subfloat{
        \centering\scriptsize%
        \input{figures/appendix/pantago2}
        \label{sfig:antago2}
    }
    \subfloat{
        \centering\scriptsize%
        \input{figures/appendix/pantago2smash}
        \label{sfig:antago2}
    }
    \newline
    \vspace{-.75cm}
    \subfloat{
        \centering\scriptsize%
        \input{figures/appendix/pago3}
    }
    \subfloat{
        \centering\scriptsize%
        \input{figures/appendix/pago3smash}
    }
    \newline
    \vspace{-.75cm}
    \subfloat{
        \centering\scriptsize%
        \input{figures/appendix/pantago3}
        \label{sfig:antago3}
    }
    \subfloat{
        \centering\scriptsize%
        \input{figures/appendix/pantago3smash}
        \label{sfig:antago3smash}
    }
    \newline
    \vspace{-.75cm}
    \subfloat{
        \centering\scriptsize%
        \input{figures/appendix/pago4}
    }
    \subfloat{
        \centering\scriptsize%
        \input{figures/appendix/pago4smash}
    }
    \newline
    \vspace{-.75cm}
    \subfloat{
        \centering\scriptsize%
        \input{figures/appendix/pantago4}
        \label{sfig:antago4}
    }
    \subfloat{
        \centering\scriptsize%
        \input{figures/appendix/pantago4smash}
        \label{sfig:antagosmash4}
    }
    \caption{
        Joint angle and action trajectories of a single return~(left) and smash~(right) motion from \sect{ssec:return} and \sect{ssec:smash}.
        The pressures are normalized between $[0\%\ldots100\%]$ to indicate that pressure ranges which map to $[\SI{0}{\bar}\ldots\SI{3}{\bar}]$.
        The impact with the ball happens at $t=\sim\SI{0.9}{\second}$ for the return and the smash.
        Green and red background color indicate the prepare and hit phase from \fig{fig:video}.
    }
    \label{fig:detailmotion}
\end{figure*}
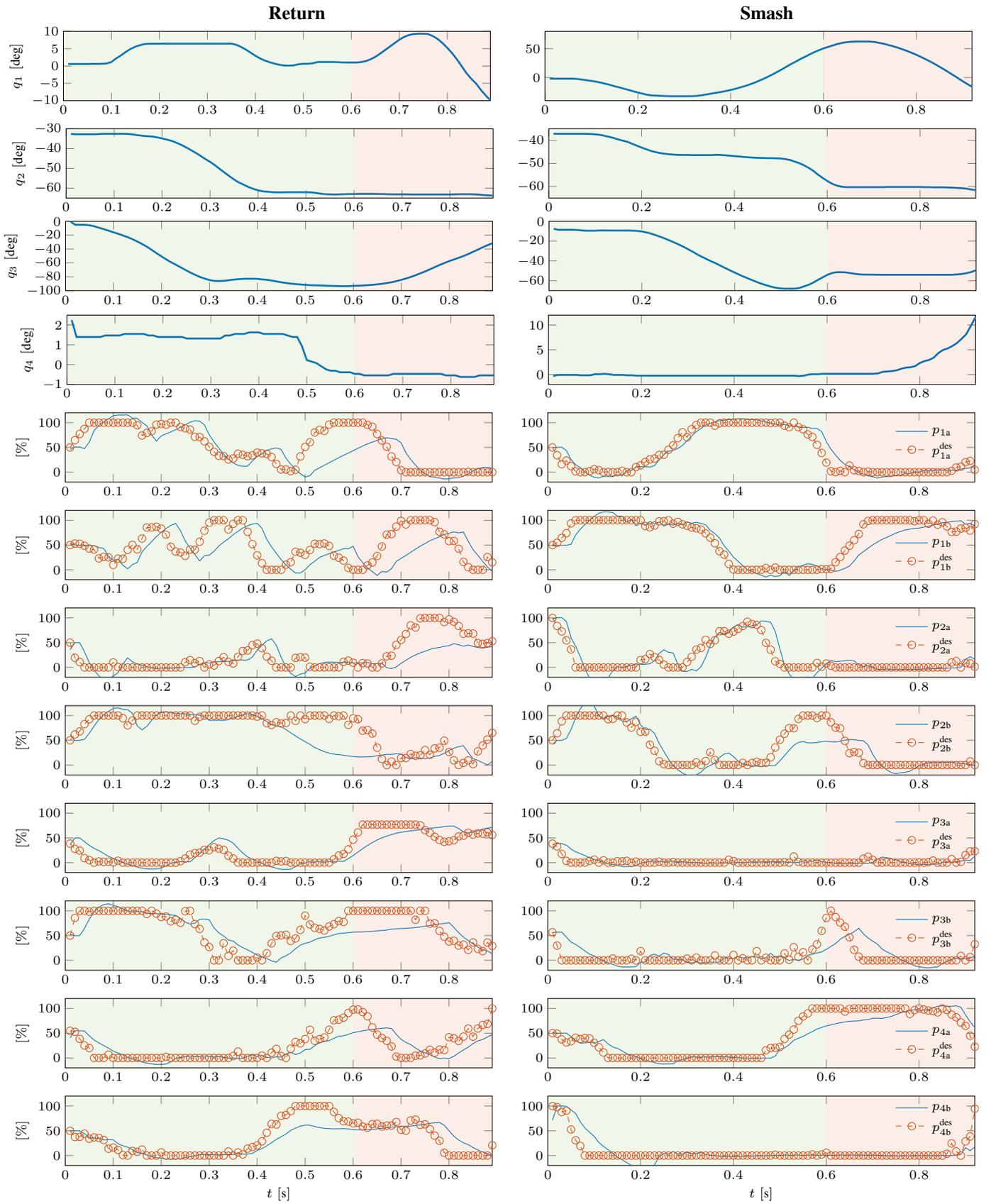

Using the training procedure from \sect{sec:methods}, the RL agent automatically learned two distinct phases of a strike motion~(prepare and hit). 
This two distinct phases are depicted in \fig{fig:video} and indicated by the green~(prepare) and red~(hit) background color in \fig{fig:stpolicy} and, in more detail, in \fig{fig:detailmotion}.
\fig{fig:detailmotion} shows a smash motion from \sect{ssec:smash} in addition to a return strike from \sect{ssec:return}. 
In contrast to the return motion, the agent learned to gain momentum in Dof 1 from the beginning of the episode and uses the other Dofs for finetuning within the hitting phase.
The agent also learned two different strategies for the return and smash motion: The return uses the third DoF in addition to the first DoF, whereas DoF 1 and 4 are used for the smash.
The low-level actions $\pdes{}$ switch multiple times within the allowed range without damaging the system. 
In this manner, the robustness of the system enables the RL agent to try and fail in a safety-critical task to find the optimal policy. 

\setlength{\figwidth }{0.95\columnwidth} 
\setlength{\figheight }{.1545\figwidth} 

%% file: figures/appendix/q1.tex
%
%
\definecolor{mycolor1}{rgb}{0.00000,0.44700,0.74100}%
\definecolor{mycolor2}{rgb}{0.85000,0.32500,0.09800}%
\definecolor{mycolor3}{rgb}{0.46600,0.67400,0.18800}%
\begin{tikzpicture}

\begin{axis}[%
width=0.951\figwidth,
height=\figheight,
at={(0\figwidth,0\figheight)},
scale only axis,
xmin=0,
xmax=0.89,
ymin=-10,
ymax=10,
ylabel style={font=\color{white!15!black}},
ylabel={$q_1~[\textrm{deg}]$},
axis background/.style={fill=white},
ylabel near ticks,
xlabel near ticks,
legend style={at={(1,1)},anchor=north west,legend cell align=left,align=left,fill=none,draw=none}
]
\addplot [color=mycolor1, line width=1.0pt, forget plot]
  table[row sep=crcr]{%
0.00999999999999979	0.57403175967297\\
0.0500000000000007	0.57403175967297\\
0.0600000000000005	0.610322079652295\\
0.0700000000000003	0.610322079652295\\
0.0800000000000001	0.64661239963162\\
0.0899999999999999	0.719193039590269\\
0.0999999999999996	1.15467687934217\\
0.109999999999999	2.2070961587426\\
0.119999999999999	3.0780638382464\\
0.130000000000001	4.05790247768818\\
0.140000000000001	4.92887015719198\\
0.15	5.58209591681983\\
0.16	6.01757975657173\\
0.17	6.34419263638566\\
0.18	6.41677327634431\\
0.199999999999999	6.41677327634431\\
0.210000000000001	6.45306359632363\\
0.35	6.45306359632363\\
0.359999999999999	6.12645071650971\\
0.369999999999999	5.50951527686118\\
0.380000000000001	4.74741855729535\\
0.4	2.96919287830843\\
0.41	2.17080583876327\\
0.42	1.55387039911475\\
0.43	1.08209623938352\\
0.44	0.682902719610945\\
0.449999999999999	0.35628983979702\\
0.460000000000001	0.138547919921068\\
0.470000000000001	0.102257599941744\\
0.48	0.174838239900394\\
0.49	0.50145111971432\\
0.5	0.64661239963162\\
0.51	0.64661239963162\\
0.52	0.791773679548919\\
0.529999999999999	1.15467687934217\\
0.56	1.15467687934217\\
0.57	1.11838655936285\\
0.58	1.0458059194042\\
0.59	1.00951559942487\\
0.609999999999999	1.00951559942487\\
0.619999999999999	1.11838655936285\\
0.630000000000001	1.40870911919745\\
0.640000000000001	1.916773598908\\
0.65	2.6425799984945\\
0.66	3.54983799797763\\
0.699999999999999	7.57806351568271\\
0.710000000000001	8.37645055522786\\
0.720000000000001	8.92080535491774\\
0.73	9.21112791475234\\
0.74	9.31999887469031\\
0.75	9.31999887469031\\
0.76	9.21112791475234\\
0.77	8.63048279508314\\
0.779999999999999	7.79580543555866\\
0.790000000000001	6.63451519622026\\
0.800000000000001	5.25548303700591\\
0.81	3.73128959787425\\
0.82	1.95306391888732\\
0.84	-2.03887127883844\\
0.85	-3.67193567790807\\
0.859999999999999	-5.05096783712242\\
0.869999999999999	-6.86548383608867\\
0.880000000000001	-8.46225791517898\\
0.890000000000001	-9.9501610343313\\
};

\addplot[area legend, draw=none, fill=mycolor2, fill opacity=0.1, forget plot]
table[row sep=crcr] {%
x	y\\
0.6	-10\\
2	-10\\
2	10\\
0.6	10\\
}--cycle;

\addplot[area legend, draw=none, fill=mycolor3, fill opacity=0.1, forget plot]
table[row sep=crcr] {%
x	y\\
0	-10\\
0.6	-10\\
0.6	10\\
0	10\\
}--cycle;
\end{axis}
\end{tikzpicture}%

%% file: figures/appendix/q1smash.tex
%
%
\definecolor{mycolor1}{rgb}{0.00000,0.44700,0.74100}%
\definecolor{mycolor2}{rgb}{0.85000,0.32500,0.09800}%
\definecolor{mycolor3}{rgb}{0.46600,0.67400,0.18800}%
\begin{tikzpicture}

\begin{axis}[%
width=0.951\figwidth,
height=\figheight,
at={(0\figwidth,0\figheight)},
scale only axis,
xmin=0,
xmax=0.92,
ymin=-40,
ymax=80,
axis background/.style={fill=white},
ylabel near ticks,
xlabel near ticks,
legend style={at={(1,1)},anchor=north west,legend cell align=left,align=left,fill=none,draw=none}
]
\addplot [color=mycolor1, line width=1.0pt, forget plot]
  table[row sep=crcr]{%
0.00999999999999801	-1.96629063887978\\
0.0600000000000023	-1.96629063887978\\
0.0700000000000003	-2.47435511859034\\
0.0799999999999983	-3.56306471797009\\
0.0900000000000034	-4.76064527728781\\
0.100000000000001	-5.81306455668825\\
0.109999999999999	-7.15580639592327\\
0.119999999999997	-8.82516111497223\\
0.130000000000003	-10.7122577538971\\
0.140000000000001	-12.7445156727393\\
0.149999999999999	-15.1759671113541\\
0.159999999999997	-17.063063750279\\
0.18	-21.4904827877567\\
0.200000000000003	-25.9179018252343\\
0.210000000000001	-27.9138694240972\\
0.219999999999999	-29.8009660630221\\
0.229999999999997	-30.9985466223399\\
0.240000000000002	-31.2888691821745\\
0.25	-31.688062701947\\
0.259999999999998	-32.014675581761\\
0.270000000000003	-32.2324175016369\\
0.280000000000001	-32.3049981415956\\
0.299999999999997	-32.3049981415956\\
0.310000000000002	-32.2324175016369\\
0.32	-32.1235465416989\\
0.329999999999998	-31.4340304620918\\
0.340000000000003	-30.3816111826913\\
0.359999999999999	-27.5146759043246\\
0.369999999999997	-26.2445147050483\\
0.380000000000003	-24.8291922258546\\
0.390000000000001	-23.196127826785\\
0.399999999999999	-21.4904827877567\\
0.409999999999997	-19.4945151888938\\
0.43	-14.9219348714988\\
0.439999999999998	-12.2727415130081\\
0.450000000000003	-9.33322559468278\\
0.460000000000001	-5.95822583660554\\
0.469999999999999	-2.43806479861102\\
0.479999999999997	1.33612847923879\\
0.490000000000002	5.32806367696455\\
0.509999999999998	14.0740307919819\\
0.530000000000001	23.3643527066891\\
0.539999999999999	27.5377395043115\\
0.549999999999997	32.0377391817478\\
0.560000000000002	35.9570937395149\\
0.57	40.0578998971787\\
0.579999999999998	43.9409641349664\\
0.590000000000003	47.2796735730644\\
0.600000000000001	50.473221731245\\
0.630000000000003	58.1304792468826\\
0.640000000000001	59.9087049258695\\
0.649999999999999	61.2514467651045\\
0.659999999999997	61.977253164691\\
0.670000000000002	62.3764466844636\\
0.689999999999998	62.3764466844636\\
0.700000000000003	62.3038660445049\\
0.710000000000001	61.5417693249391\\
0.719999999999999	60.0538662057868\\
0.729999999999997	58.3845114867378\\
0.75	54.1748343691361\\
0.759999999999998	51.6708022905627\\
0.770000000000003	48.9490282921133\\
0.780000000000001	45.9732220538086\\
0.789999999999999	42.8159642156074\\
0.799999999999997	39.4046741375508\\
0.810000000000002	35.7030614996596\\
0.829999999999998	27.6829007842288\\
0.840000000000003	23.3643527066891\\
0.859999999999999	13.8925791920853\\
0.880000000000003	3.9490315177502\\
0.899999999999999	-6.42999999633678\\
0.909999999999997	-11.365483513525\\
0.920000000000002	-15.9017735109406\\
};

\addplot[area legend, draw=none, fill=mycolor2, fill opacity=0.1, forget plot]
table[row sep=crcr] {%
x	y\\
0.6	-40\\
2	-40\\
2	80\\
0.6	80\\
}--cycle;

\addplot[area legend, draw=none, fill=mycolor3, fill opacity=0.1, forget plot]
table[row sep=crcr] {%
x	y\\
0	-40\\
0.6	-40\\
0.6	80\\
0	80\\
}--cycle;
\end{axis}
\end{tikzpicture}%

%% file: figures/appendix/q2.tex
%
%
\definecolor{mycolor1}{rgb}{0.00000,0.44700,0.74100}%
\definecolor{mycolor2}{rgb}{0.85000,0.32500,0.09800}%
\definecolor{mycolor3}{rgb}{0.46600,0.67400,0.18800}%
\begin{tikzpicture}

\begin{axis}[%
width=0.951\figwidth,
height=\figheight,
at={(0\figwidth,0\figheight)},
scale only axis,
xmin=0,
xmax=0.89,
ymin=-65,
ymax=-30,
ylabel style={font=\color{white!15!black}},
ylabel={$q_2~[\textrm{deg}]$},
axis background/.style={fill=white},
ylabel near ticks,
xlabel near ticks,
legend style={at={(1,1)},anchor=north west,legend cell align=left,align=left,fill=none,draw=none}
]
\addplot [color=mycolor1, line width=1.0pt, forget plot]
  table[row sep=crcr]{%
0.00999999999999801	-32.6003249814273\\
0.0200000000000031	-32.7454862813446\\
0.0600000000000023	-32.7454862813446\\
0.0700000000000003	-32.6003249814273\\
0.0799999999999983	-32.5277443314687\\
0.119999999999997	-32.5277443314687\\
0.130000000000003	-32.6003249814273\\
0.140000000000001	-32.8906475812619\\
0.149999999999999	-33.2535508310552\\
0.159999999999997	-33.5438734308898\\
0.170000000000002	-33.7616153807657\\
0.18	-33.906776680683\\
0.189999999999998	-34.3422605804349\\
0.200000000000003	-34.8503251301455\\
0.210000000000001	-35.4309703298147\\
0.219999999999999	-36.0841961794425\\
0.229999999999997	-37.027744628905\\
0.240000000000002	-38.1164543782847\\
0.25	-39.3503254275818\\
0.270000000000003	-42.1083901260105\\
0.280000000000001	-43.6325837751421\\
0.289999999999999	-45.2293580742324\\
0.299999999999997	-46.6809710734054\\
0.310000000000002	-48.3503260224544\\
0.32	-50.2374229213793\\
0.340000000000003	-53.8664554193118\\
0.350000000000001	-55.6083910183194\\
0.359999999999999	-57.0600040174924\\
0.380000000000003	-59.3825848161692\\
0.390000000000001	-60.3261332656317\\
0.399999999999999	-61.0519397652182\\
0.409999999999997	-61.5600043149287\\
0.43	-61.9954882146806\\
0.439999999999998	-62.0680688646393\\
0.450000000000003	-62.0680688646393\\
0.460000000000001	-61.9954882146806\\
0.5	-61.9954882146806\\
0.509999999999998	-62.0680688646393\\
0.520000000000003	-62.3583914644739\\
0.530000000000001	-62.7212947142671\\
0.539999999999999	-63.0116173141017\\
0.560000000000002	-63.156778614019\\
0.590000000000003	-62.9390366641431\\
0.600000000000001	-62.9390366641431\\
0.609999999999999	-62.8664560141844\\
0.640000000000001	-62.8664560141844\\
0.649999999999999	-62.9390366641431\\
0.659999999999997	-62.9390366641431\\
0.68	-63.0841979640604\\
0.710000000000001	-63.0841979640604\\
0.719999999999999	-63.156778614019\\
0.799999999999997	-63.156778614019\\
0.82	-63.0116173141017\\
0.840000000000003	-63.0116173141017\\
0.850000000000001	-63.0841979640604\\
0.859999999999999	-63.2293592639777\\
0.869999999999997	-63.4471012138536\\
0.890000000000001	-63.7374238136882\\
};

\addplot[area legend, draw=none, fill=mycolor2, fill opacity=0.1, forget plot]
table[row sep=crcr] {%
x	y\\
0.6	-65\\
2	-65\\
2	-30\\
0.6	-30\\
}--cycle;

\addplot[area legend, draw=none, fill=mycolor3, fill opacity=0.1, forget plot]
table[row sep=crcr] {%
x	y\\
0	-65\\
0.6	-65\\
0.6	-30\\
0	-30\\
}--cycle;
\end{axis}
\end{tikzpicture}%

%% file: figures/appendix/q2smash.tex
%
%
\definecolor{mycolor1}{rgb}{0.00000,0.44700,0.74100}%
\definecolor{mycolor2}{rgb}{0.85000,0.32500,0.09800}%
\definecolor{mycolor3}{rgb}{0.46600,0.67400,0.18800}%
\begin{tikzpicture}

\begin{axis}[%
width=0.951\figwidth,
height=\figheight,
at={(0\figwidth,0\figheight)},
scale only axis,
xmin=0,
xmax=0.92,
ymin=-65,
ymax=-35,
axis background/.style={fill=white},
ylabel near ticks,
xlabel near ticks,
legend style={at={(1,1)},anchor=north west,legend cell align=left,align=left,fill=none,draw=none}
]
\addplot [color=mycolor1, line width=1.0pt, forget plot]
  table[row sep=crcr]{%
0.00999999999999801	-37.1729059288223\\
0.0799999999999983	-37.1729059288223\\
0.0900000000000034	-37.2454865787809\\
0.100000000000001	-37.3906478786982\\
0.109999999999999	-37.6083898285742\\
0.130000000000003	-38.479357628078\\
0.140000000000001	-38.9874221777885\\
0.149999999999999	-39.6406480274164\\
0.170000000000002	-40.6567771268375\\
0.18	-41.5277449263413\\
0.200000000000003	-43.1245192254316\\
0.219999999999999	-44.5761322246046\\
0.229999999999997	-45.0841967743151\\
0.240000000000002	-45.519680674067\\
0.25	-45.8100032739016\\
0.259999999999998	-46.0277452237776\\
0.280000000000001	-46.3180678236122\\
0.299999999999997	-46.3180678236122\\
0.310000000000002	-46.3906484735708\\
0.350000000000001	-46.3906484735708\\
0.359999999999999	-46.2454871736535\\
0.369999999999997	-46.3180678236122\\
0.390000000000001	-46.6083904234468\\
0.399999999999999	-46.8261323733227\\
0.409999999999997	-46.97129367324\\
0.420000000000002	-47.189035623116\\
0.439999999999998	-47.4793582229506\\
0.469999999999999	-47.6971001728265\\
0.479999999999997	-47.6971001728265\\
0.5	-47.8422614727438\\
0.509999999999998	-48.1325840725784\\
0.520000000000003	-48.4954873223717\\
0.530000000000001	-49.2212938219582\\
0.539999999999999	-49.874519671586\\
0.549999999999997	-50.8180681210485\\
0.560000000000002	-51.9067778704283\\
0.57	-53.213229569684\\
0.579999999999998	-54.8100038687743\\
0.590000000000003	-56.2616168679473\\
0.600000000000001	-57.568068567203\\
0.609999999999999	-58.7293589665414\\
0.619999999999997	-59.5277461160865\\
0.630000000000003	-60.0358106657971\\
0.640000000000001	-60.253552615673\\
0.719999999999999	-60.253552615673\\
0.729999999999997	-60.1809719657144\\
0.799999999999997	-60.1809719657144\\
0.810000000000002	-60.253552615673\\
0.82	-60.253552615673\\
0.829999999999998	-60.3261332656317\\
0.850000000000001	-60.3261332656317\\
0.869999999999997	-60.471294565549\\
0.880000000000003	-60.6164558654663\\
0.890000000000001	-60.6890365154249\\
0.899999999999999	-60.9067784653009\\
0.909999999999997	-61.2696817150941\\
0.920000000000002	-61.5600043149287\\
};

\addplot[area legend, draw=none, fill=mycolor2, fill opacity=0.1, forget plot]
table[row sep=crcr] {%
x	y\\
0.6	-65\\
2	-65\\
2	-35\\
0.6	-35\\
}--cycle;

\addplot[area legend, draw=none, fill=mycolor3, fill opacity=0.1, forget plot]
table[row sep=crcr] {%
x	y\\
0	-65\\
0.6	-65\\
0.6	-35\\
0	-35\\
}--cycle;
\end{axis}
\end{tikzpicture}%

%% file: figures/appendix/q3.tex
%
%
\definecolor{mycolor1}{rgb}{0.00000,0.44700,0.74100}%
\definecolor{mycolor2}{rgb}{0.85000,0.32500,0.09800}%
\definecolor{mycolor3}{rgb}{0.46600,0.67400,0.18800}%
\begin{tikzpicture}

\begin{axis}[%
width=0.951\figwidth,
height=\figheight,
at={(0\figwidth,0\figheight)},
scale only axis,
xmin=0,
xmax=0.89,
ymin=-100,
ymax=0,
ylabel style={font=\color{white!15!black}},
ylabel={$q_3~[\textrm{deg}]$},
axis background/.style={fill=white},
ylabel near ticks,
xlabel near ticks,
legend style={at={(1,1)},anchor=north west,legend cell align=left,align=left,fill=none,draw=none}
]
\addplot [color=mycolor1, line width=1.0pt, forget plot]
  table[row sep=crcr]{%
0.0100000000000051	-0.288709899835524\\
0.019999999999996	-5.00645214714778\\
0.0400000000000063	-5.00645214714778\\
0.0499999999999972	-5.65967799677563\\
0.0600000000000023	-7.25645229586594\\
0.0799999999999983	-11.393549343509\\
0.109999999999999	-18.0709691397048\\
0.129999999999995	-23.006453336893\\
0.140000000000001	-25.9822599851977\\
0.150000000000006	-29.2483892333369\\
0.159999999999997	-32.9500023812281\\
0.170000000000002	-36.8693574789952\\
0.180000000000007	-41.3693577764315\\
0.189999999999998	-46.2322613236611\\
0.209999999999994	-55.0145199686577\\
0.219999999999999	-59.0790363663422\\
0.230000000000004	-62.635488214316\\
0.260000000000005	-73.8854889579068\\
0.269999999999996	-77.151618206046\\
0.280000000000001	-80.272586154268\\
0.290000000000006	-82.6677476029034\\
0.299999999999997	-84.7000058017457\\
0.310000000000002	-85.9338768510427\\
0.319999999999993	-86.1516188009187\\
0.329999999999998	-85.7887155511254\\
0.340000000000003	-84.9903284015802\\
0.349999999999994	-84.1193606020765\\
0.359999999999999	-83.4661347524486\\
0.370000000000005	-82.9580702027381\\
0.379999999999995	-82.8129089028207\\
0.390000000000001	-82.8129089028207\\
0.400000000000006	-83.0306508526967\\
0.409999999999997	-83.8290380022418\\
0.420000000000002	-84.4096832019111\\
0.430000000000007	-85.4258123013322\\
0.439999999999998	-86.7322640005878\\
0.450000000000003	-87.9661350498849\\
0.459999999999994	-89.0548447992647\\
0.469999999999999	-89.8532319488098\\
0.480000000000004	-90.5790384483963\\
0.489999999999995	-91.0871029981069\\
0.5	-91.6677481977761\\
0.510000000000005	-92.103232097528\\
0.519999999999996	-92.3209740474039\\
0.530000000000001	-92.4661353473212\\
0.549999999999997	-93.0467805469904\\
0.569999999999993	-93.4822644467423\\
0.579999999999998	-93.554845096701\\
0.590000000000003	-93.4822644467423\\
0.599999999999994	-93.1193611969491\\
0.620000000000005	-92.5387159972799\\
0.629999999999995	-92.103232097528\\
0.640000000000001	-91.5225868978588\\
0.650000000000006	-90.8693610482309\\
0.659999999999997	-90.0709738986858\\
0.670000000000002	-88.982264149306\\
0.680000000000007	-87.8209737499676\\
0.689999999999998	-86.2241994508773\\
0.700000000000003	-84.4096832019111\\
0.709999999999994	-82.3774250030689\\
0.719999999999999	-80.2000055043093\\
0.730000000000004	-77.6596827557566\\
0.75	-72.1435533588992\\
0.760000000000005	-69.0951660606359\\
0.769999999999996	-65.9016174624553\\
0.780000000000001	-62.853230164192\\
0.790000000000006	-60.0225848158046\\
0.799999999999997	-57.3371007673345\\
0.819999999999993	-52.474197220105\\
0.829999999999998	-49.9338744715522\\
0.840000000000003	-47.0306484732062\\
0.849999999999994	-43.764519225067\\
0.859999999999999	-40.643551276845\\
0.870000000000005	-37.3048413787471\\
0.879999999999995	-34.3290347304424\\
0.890000000000001	-31.5709700320137\\
};

\addplot[area legend, draw=none, fill=mycolor2, fill opacity=0.1, forget plot]
table[row sep=crcr] {%
x	y\\
0.6	-100\\
2	-100\\
2	0\\
0.6	0\\
}--cycle;

\addplot[area legend, draw=none, fill=mycolor3, fill opacity=0.1, forget plot]
table[row sep=crcr] {%
x	y\\
0	-100\\
0.6	-100\\
0.6	0\\
0	0\\
}--cycle;
\end{axis}
\end{tikzpicture}%

%% file: figures/appendix/q3smash.tex
%
%
\definecolor{mycolor1}{rgb}{0.00000,0.44700,0.74100}%
\definecolor{mycolor2}{rgb}{0.85000,0.32500,0.09800}%
\definecolor{mycolor3}{rgb}{0.46600,0.67400,0.18800}%
\begin{tikzpicture}

\begin{axis}[%
width=0.951\figwidth,
height=\figheight,
at={(0\figwidth,0\figheight)},
scale only axis,
xmin=0,
xmax=0.92,
ymin=-70,
ymax=0,
axis background/.style={fill=white},
ylabel near ticks,
xlabel near ticks,
legend style={at={(1,1)},anchor=north west,legend cell align=left,align=left,fill=none,draw=none}
]
\addplot [color=mycolor1, line width=1.0pt, forget plot]
  table[row sep=crcr]{%
0.0100000000000051	-7.54677489570054\\
0.019999999999996	-8.56290399512164\\
0.0499999999999972	-8.56290399512164\\
0.0600000000000023	-8.70806529503894\\
0.0699999999999932	-9.14354919479084\\
0.0799999999999983	-9.50645244458408\\
0.0999999999999943	-9.50645244458408\\
0.109999999999999	-9.28871049470814\\
0.120000000000005	-9.28871049470814\\
0.129999999999995	-9.36129114466679\\
0.170000000000002	-9.36129114466679\\
0.180000000000007	-9.43387179462545\\
0.189999999999998	-9.72419439446004\\
0.200000000000003	-10.2322589441706\\
0.209999999999994	-11.1758073936331\\
0.219999999999999	-12.4096784429301\\
0.230000000000004	-14.0064527420204\\
0.239999999999995	-15.6758076910694\\
0.25	-17.417743290077\\
0.269999999999996	-21.4096790378027\\
0.280000000000001	-23.5145178866036\\
0.290000000000006	-25.7645180353217\\
0.310000000000002	-30.7725828824686\\
0.319999999999993	-33.6032282308559\\
0.329999999999998	-35.9983896794914\\
0.349999999999994	-40.4983899769277\\
0.359999999999999	-42.5306481757699\\
0.379999999999995	-47.2483904230822\\
0.400000000000006	-51.7483907205185\\
0.420000000000002	-55.9580684181202\\
0.439999999999998	-59.8048428659286\\
0.450000000000003	-61.6919397648536\\
0.459999999999994	-63.3612947139025\\
0.469999999999999	-64.8129077130755\\
0.480000000000004	-66.1919400622899\\
0.489999999999995	-67.208069161711\\
0.5	-67.7887143613802\\
0.510000000000005	-68.0064563112561\\
0.519999999999996	-68.0064563112561\\
0.530000000000001	-67.9338756612975\\
0.540000000000006	-67.2806498116696\\
0.549999999999997	-65.538714212662\\
0.560000000000002	-64.0145205635304\\
0.569999999999993	-61.6919397648536\\
0.579999999999998	-59.0064557163835\\
0.599999999999994	-53.9983908692366\\
0.609999999999999	-52.256455270229\\
0.620000000000005	-51.3854874707252\\
0.629999999999995	-51.3854874707252\\
0.640000000000001	-51.6032294206012\\
0.659999999999997	-53.0548424197742\\
0.670000000000002	-53.6354876194434\\
0.680000000000007	-53.8532295693193\\
0.700000000000003	-53.8532295693193\\
0.719999999999999	-53.9983908692366\\
0.859999999999999	-53.9983908692366\\
0.870000000000005	-53.925810219278\\
0.879999999999995	-53.5629069694847\\
0.890000000000001	-53.1274230697328\\
0.900000000000006	-52.3290359201877\\
0.909999999999997	-51.1677455208493\\
0.920000000000002	-49.4258099218417\\
};

\addplot[area legend, draw=none, fill=mycolor2, fill opacity=0.1, forget plot]
table[row sep=crcr] {%
x	y\\
0.6	-70\\
2	-70\\
2	0\\
0.6	0\\
}--cycle;

\addplot[area legend, draw=none, fill=mycolor3, fill opacity=0.1, forget plot]
table[row sep=crcr] {%
x	y\\
0	-70\\
0.6	-70\\
0.6	0\\
0	0\\
}--cycle;
\end{axis}
\end{tikzpicture}%

%% file: figures/appendix/q4.tex
%
%
\definecolor{mycolor1}{rgb}{0.00000,0.44700,0.74100}%
\definecolor{mycolor2}{rgb}{0.85000,0.32500,0.09800}%
\definecolor{mycolor3}{rgb}{0.46600,0.67400,0.18800}%
\begin{tikzpicture}

\begin{axis}[%
width=0.951\figwidth,
height=\figheight,
at={(0\figwidth,0\figheight)},
scale only axis,
xmin=0,
xmax=0.89,
ymin=-1,
ymax=2.5,
ylabel style={font=\color{white!15!black}},
ylabel={$q_4~[\textrm{deg}]$},
axis background/.style={fill=white},
ylabel near ticks,
xlabel near ticks,
legend style={at={(1,1)},anchor=north west,legend cell align=left,align=left,fill=none,draw=none}
]
\addplot [color=mycolor1, line width=1.0pt, forget plot]
  table[row sep=crcr]{%
0.00999999999999979	2.25000008871816\\
0.02	1.39655177920437\\
0.0699999999999998	1.39655177920437\\
0.0800000000000001	1.47413798916017\\
0.11	1.47413798916017\\
0.12	1.55172419911597\\
0.16	1.55172419911597\\
0.17	1.47413798916017\\
0.18	1.47413798916017\\
0.19	1.39655177920437\\
0.24	1.39655177920437\\
0.25	1.31896556924857\\
0.32	1.31896556924857\\
0.33	1.47413798916017\\
0.34	1.47413798916017\\
0.35	1.55172419911597\\
0.37	1.55172419911597\\
0.38	1.62931040907177\\
0.4	1.62931040907177\\
0.41	1.55172419911597\\
0.45	1.55172419911597\\
0.47	1.39655177920437\\
0.48	1.39655177920437\\
0.49	0.931034519469581\\
0.5	0.232758629867395\\
0.52	0.0775862099557987\\
0.54	-0.232758629867395\\
0.55	-0.310344839823194\\
0.56	-0.310344839823194\\
0.57	-0.387931049778992\\
0.59	-0.387931049778992\\
0.6	-0.465517259734791\\
0.61	-0.465517259734791\\
0.62	-0.543103469690589\\
0.67	-0.543103469690589\\
0.68	-0.465517259734791\\
0.78	-0.465517259734791\\
0.79	-0.543103469690589\\
0.81	-0.543103469690589\\
0.82	-0.620689679646388\\
0.85	-0.620689679646388\\
0.86	-0.543103469690589\\
0.89	-0.543103469690589\\
};

\addplot[area legend, draw=none, fill=mycolor2, fill opacity=0.1, forget plot]
table[row sep=crcr] {%
x	y\\
0.6	-1\\
2	-1\\
2	2.5\\
0.6	2.5\\
}--cycle;

\addplot[area legend, draw=none, fill=mycolor3, fill opacity=0.1, forget plot]
table[row sep=crcr] {%
x	y\\
0	-1\\
0.6	-1\\
0.6	2.5\\
0	2.5\\
}--cycle;
\end{axis}
\end{tikzpicture}%

%% file: figures/appendix/q4smash.tex
%
%
\definecolor{mycolor1}{rgb}{0.00000,0.44700,0.74100}%
\definecolor{mycolor2}{rgb}{0.85000,0.32500,0.09800}%
\definecolor{mycolor3}{rgb}{0.46600,0.67400,0.18800}%
\begin{tikzpicture}

\begin{axis}[%
width=0.951\figwidth,
height=\figheight,
at={(0\figwidth,0\figheight)},
scale only axis,
xmin=0,
xmax=0.92,
ymin=-2,
ymax=12,
axis background/.style={fill=white},
ylabel near ticks,
xlabel near ticks,
legend style={at={(1,1)},anchor=north west,legend cell align=left,align=left,fill=none,draw=none}
]
\addplot [color=mycolor1, line width=1.0pt, forget plot]
  table[row sep=crcr]{%
0.00999999999999979	-0.310344839823195\\
0.0199999999999996	-0.0775862099557987\\
0.0600000000000005	-0.0775862099557987\\
0.0700000000000003	-0.155172419911597\\
0.0899999999999999	-0.155172419911597\\
0.0999999999999996	0.0775862099557987\\
0.109999999999999	0.0775862099557987\\
0.119999999999999	0.155172419911597\\
0.130000000000001	0.0775862099557987\\
0.140000000000001	-0.0775862099557987\\
0.16	-0.0775862099557987\\
0.17	-0.155172419911597\\
0.18	-0.155172419911597\\
0.19	-0.232758629867396\\
0.529999999999999	-0.232758629867396\\
0.540000000000001	-0.310344839823195\\
0.550000000000001	-0.232758629867396\\
0.56	0\\
0.57	0.0775862099557987\\
0.58	0.0775862099557987\\
0.59	0.155172419911597\\
0.699999999999999	0.155172419911597\\
0.710000000000001	0.232758629867396\\
0.720000000000001	0.543103469690589\\
0.74	0.698275889602186\\
0.75	0.853448309513784\\
0.76	1.16379314933698\\
0.77	1.39655177920437\\
0.779999999999999	1.47413798916017\\
0.790000000000001	1.70689661902757\\
0.800000000000001	2.32758629867395\\
0.81	2.71551734845295\\
0.82	2.94827597832034\\
0.83	3.41379323805513\\
0.84	4.34482775752471\\
0.85	4.9655174371711\\
0.859999999999999	5.2758622769943\\
0.869999999999999	5.74137953672909\\
0.880000000000001	6.36206921637547\\
0.890000000000001	7.21551752588926\\
0.9	8.14655204535884\\
0.92	11.560345283414\\
};

\addplot[area legend, draw=none, fill=mycolor2, fill opacity=0.1, forget plot]
table[row sep=crcr] {%
x	y\\
0.6	-2\\
2	-2\\
2	12\\
0.6	12\\
}--cycle;

\addplot[area legend, draw=none, fill=mycolor3, fill opacity=0.1, forget plot]
table[row sep=crcr] {%
x	y\\
0	-2\\
0.6	-2\\
0.6	12\\
0	12\\
}--cycle;
\end{axis}
\end{tikzpicture}%

%% file: figures/appendix/pago1.tex
%
%
\definecolor{mycolor1}{rgb}{0.00000,0.44700,0.74100}%
\definecolor{mycolor2}{rgb}{0.85000,0.32500,0.09800}%
\definecolor{mycolor3}{rgb}{0.46600,0.67400,0.18800}%
\begin{tikzpicture}

\begin{axis}[%
width=0.951\figwidth,
height=\figheight,
at={(0\figwidth,0\figheight)},
scale only axis,
xmin=0,
xmax=0.89,
ymin=-20,
ymax=120,
ylabel style={font=\color{white!15!black}},
ylabel={$[\%]$},
axis background/.style={fill=white},
ylabel near ticks,
xlabel near ticks,
legend style={at={(.8,.9)},anchor=north west,legend cell align=left,align=left,fill=none,draw=none}
]
\addplot [color=mycolor1, forget plot]
  table[row sep=crcr]{%
0.0100000000000051	45.6555555555556\\
0.019999999999996	50.8111111111111\\
0.0300000000000011	50.8111111111111\\
0.0400000000000063	47.6\\
0.0499999999999972	60.8444444444444\\
0.0600000000000023	84.0111111111111\\
0.0699999999999932	94.6333333333333\\
0.0799999999999983	101.666666666667\\
0.0900000000000034	107.744444444444\\
0.0999999999999943	114.088888888889\\
0.109999999999999	115.411111111111\\
0.120000000000005	115.466666666667\\
0.129999999999995	115.688888888889\\
0.140000000000001	109.022222222222\\
0.150000000000006	108.677777777778\\
0.159999999999997	100.288888888889\\
0.170000000000002	93.7666666666667\\
0.180000000000007	75.0777777777778\\
0.189999999999998	62.7777777777778\\
0.200000000000003	73.3555555555556\\
0.209999999999994	77.5111111111111\\
0.219999999999999	81.3111111111111\\
0.230000000000004	85.7333333333333\\
0.239999999999995	90.4333333333333\\
0.25	95.5333333333333\\
0.260000000000005	100.911111111111\\
0.269999999999996	104.155555555556\\
0.280000000000001	103.122222222222\\
0.290000000000006	95.5444444444445\\
0.299999999999997	75.5555555555556\\
0.310000000000002	61.2444444444445\\
0.329999999999998	36.2222222222222\\
0.340000000000003	26.1444444444444\\
0.349999999999994	18.7222222222222\\
0.359999999999999	13.7222222222222\\
0.370000000000005	11.7666666666667\\
0.379999999999995	12.1111111111111\\
0.390000000000001	22.5\\
0.400000000000006	31.3777777777778\\
0.409999999999997	37.0555555555556\\
0.420000000000002	44.0888888888889\\
0.430000000000007	47.0333333333333\\
0.439999999999998	48.9333333333333\\
0.450000000000003	45.8\\
0.459999999999994	28.2888888888889\\
0.469999999999999	16.1777777777778\\
0.480000000000004	5.58888888888889\\
0.489999999999995	-3.07777777777778\\
0.5	-9.61111111111111\\
0.510000000000005	-7.2\\
0.519999999999996	3.86666666666666\\
0.569999999999993	29.2111111111111\\
0.579999999999998	34.0888888888889\\
0.590000000000003	38.5\\
0.609999999999999	48.2777777777778\\
0.620000000000005	52.8555555555556\\
0.629999999999995	57.1666666666667\\
0.650000000000006	64.3444444444444\\
0.659999999999997	67.5888888888889\\
0.670000000000002	69.4777777777778\\
0.680000000000007	68.0555555555556\\
0.689999999999998	66.4222222222222\\
0.700000000000003	59.4222222222222\\
0.709999999999994	39.1666666666667\\
0.719999999999999	25.8\\
0.730000000000004	14.3111111111111\\
0.739999999999995	4.06666666666666\\
0.75	-3.77777777777777\\
0.760000000000005	-8.97777777777777\\
0.769999999999996	-10.3\\
0.780000000000001	-11.8555555555556\\
0.790000000000006	-13.9444444444444\\
0.799999999999997	-13.2777777777778\\
0.810000000000002	-11.3555555555556\\
0.819999999999993	-10.9888888888889\\
0.829999999999998	-8.11111111111111\\
0.840000000000003	-3.74444444444444\\
0.849999999999994	2.64444444444445\\
0.859999999999999	7.2\\
0.870000000000005	9.37777777777778\\
0.879999999999995	9.68888888888888\\
0.890000000000001	8.17777777777778\\
};
\addplot [color=mycolor2, dashed, mark=o, mark options={solid, mycolor2}, forget plot]
  table[row sep=crcr]{%
0.0100000000000051	50\\
0.019999999999996	64.6222222222222\\
0.0300000000000011	77.1444444444444\\
0.0400000000000063	87.5222222222222\\
0.0499999999999972	100\\
0.0600000000000023	100\\
0.0699999999999932	100\\
0.0799999999999983	100\\
0.0900000000000034	100\\
0.0999999999999943	100\\
0.109999999999999	100\\
0.120000000000005	100\\
0.129999999999995	100\\
0.140000000000001	100\\
0.150000000000006	94.8222222222222\\
0.159999999999997	74.0888888888889\\
0.170000000000002	79.2444444444444\\
0.180000000000007	82.7333333333333\\
0.189999999999998	96.1111111111111\\
0.200000000000003	97.3444444444444\\
0.209999999999994	95.8111111111111\\
0.219999999999999	100\\
0.230000000000004	100\\
0.239999999999995	91.2666666666667\\
0.25	88.6\\
0.260000000000005	86.2222222222222\\
0.269999999999996	71.4\\
0.280000000000001	70.9888888888889\\
0.290000000000006	62.2222222222222\\
0.299999999999997	51.2555555555556\\
0.310000000000002	41.1111111111111\\
0.319999999999993	33.0333333333333\\
0.329999999999998	32.3777777777778\\
0.340000000000003	25.7222222222222\\
0.349999999999994	27.4111111111111\\
0.359999999999999	37.2333333333333\\
0.370000000000005	35.0222222222222\\
0.379999999999995	32.2333333333333\\
0.390000000000001	34.0777777777778\\
0.400000000000006	39.6666666666667\\
0.409999999999997	38.0222222222222\\
0.420000000000002	36.8777777777778\\
0.430000000000007	30.5111111111111\\
0.439999999999998	16.6333333333333\\
0.450000000000003	5.35555555555555\\
0.459999999999994	5.36666666666666\\
0.469999999999999	2.05555555555556\\
0.480000000000004	4.83333333333333\\
0.489999999999995	31.7222222222222\\
0.5	51\\
0.510000000000005	60.4777777777778\\
0.519999999999996	80.2666666666667\\
0.530000000000001	85.4666666666667\\
0.540000000000006	83.4111111111111\\
0.549999999999997	96.0555555555556\\
0.560000000000002	100\\
0.569999999999993	100\\
0.579999999999998	100\\
0.590000000000003	100\\
0.599999999999994	100\\
0.609999999999999	100\\
0.620000000000005	100\\
0.629999999999995	95.6\\
0.640000000000001	91.8\\
0.650000000000006	77.1\\
0.659999999999997	64.1111111111111\\
0.670000000000002	51.4222222222222\\
0.680000000000007	31.6666666666667\\
0.689999999999998	20.5444444444445\\
0.700000000000003	4.02222222222223\\
0.709999999999994	0\\
0.719999999999999	0\\
0.730000000000004	0\\
0.739999999999995	0\\
0.75	0\\
0.760000000000005	0\\
0.769999999999996	0\\
0.780000000000001	0\\
0.790000000000006	0\\
0.799999999999997	0\\
0.810000000000002	0\\
0.819999999999993	0\\
0.829999999999998	0\\
0.840000000000003	0.688888888888883\\
0.849999999999994	0\\
0.859999999999999	0\\
0.870000000000005	0\\
0.879999999999995	0\\
0.890000000000001	0\\
};

\addplot[area legend, draw=none, fill=mycolor2, fill opacity=0.1, forget plot]
table[row sep=crcr] {%
x	y\\
0.6	-20\\
2	-20\\
2	120\\
0.6	120\\
}--cycle;

\addplot[area legend, draw=none, fill=mycolor3, fill opacity=0.1, forget plot]
table[row sep=crcr] {%
x	y\\
0	-20\\
0.6	-20\\
0.6	120\\
0	120\\
}--cycle;
\end{axis}
\end{tikzpicture}%

%% file: figures/appendix/pago1smash.tex
%
%
\definecolor{mycolor1}{rgb}{0.00000,0.44700,0.74100}%
\definecolor{mycolor2}{rgb}{0.85000,0.32500,0.09800}%
\definecolor{mycolor3}{rgb}{0.46600,0.67400,0.18800}%
\begin{tikzpicture}

\begin{axis}[%
width=0.951\figwidth,
height=\figheight,
at={(0\figwidth,0\figheight)},
scale only axis,
xmin=0,
xmax=0.92,
ymin=-20,
ymax=120,
axis background/.style={fill=white},
legend style={legend cell align=left, align=left, fill=none, draw=none},
ylabel near ticks,
xlabel near ticks,
legend style={at={(.8,.9)},anchor=north west,legend cell align=left,align=left,fill=none,draw=none}
]
\addplot [color=mycolor1]
  table[row sep=crcr]{%
0.0100000000000051	49.0888888888889\\
0.019999999999996	50.0888888888889\\
0.0300000000000011	50.3444444444444\\
0.0400000000000063	43.7111111111111\\
0.0499999999999972	25.5777777777778\\
0.0600000000000023	14.9222222222222\\
0.0699999999999932	9.2\\
0.0799999999999983	0.477777777777774\\
0.0900000000000034	-5.94444444444444\\
0.0999999999999943	-10.4\\
0.109999999999999	-11.3666666666667\\
0.120000000000005	-8.5\\
0.129999999999995	-8.66666666666667\\
0.140000000000001	-9.04444444444445\\
0.150000000000006	-5.01111111111111\\
0.159999999999997	-7.57777777777778\\
0.170000000000002	2.88888888888889\\
0.180000000000007	3.13333333333334\\
0.189999999999998	11.0777777777778\\
0.200000000000003	10.8444444444444\\
0.209999999999994	10.9\\
0.219999999999999	10.0444444444445\\
0.230000000000004	13.4777777777778\\
0.239999999999995	24.9444444444444\\
0.25	35.2\\
0.260000000000005	42.3666666666667\\
0.269999999999996	43.8666666666667\\
0.280000000000001	47.8222222222222\\
0.290000000000006	53.9777777777778\\
0.299999999999997	63.7444444444445\\
0.319999999999993	81.9888888888889\\
0.329999999999998	88.2333333333333\\
0.340000000000003	95.4444444444444\\
0.349999999999994	98.8555555555556\\
0.359999999999999	101.844444444444\\
0.370000000000005	104.044444444444\\
0.390000000000001	106.444444444444\\
0.400000000000006	106.933333333333\\
0.409999999999997	108.044444444444\\
0.420000000000002	108.1\\
0.430000000000007	106.488888888889\\
0.439999999999998	106.411111111111\\
0.450000000000003	105.8\\
0.469999999999999	101.344444444444\\
0.480000000000004	99.0111111111111\\
0.489999999999995	97.1666666666667\\
0.5	95.4\\
0.510000000000005	96.5777777777778\\
0.519999999999996	93.9222222222222\\
0.530000000000001	92.1666666666667\\
0.540000000000006	91.4444444444444\\
0.549999999999997	90.3555555555556\\
0.560000000000002	90.7111111111111\\
0.569999999999993	88.3555555555556\\
0.579999999999998	88.4444444444444\\
0.590000000000003	81.8666666666667\\
0.599999999999994	61.8\\
0.609999999999999	46.3555555555556\\
0.620000000000005	34.5666666666667\\
0.629999999999995	24.0222222222222\\
0.640000000000001	15.6888888888889\\
0.650000000000006	11.1888888888889\\
0.659999999999997	5.03333333333333\\
0.670000000000002	-0.766666666666666\\
0.680000000000007	-4.26666666666667\\
0.689999999999998	-7.45555555555556\\
0.700000000000003	-9.38888888888889\\
0.709999999999994	-11.6444444444444\\
0.719999999999999	-10.0111111111111\\
0.730000000000004	-8.81111111111112\\
0.739999999999995	-8.53333333333333\\
0.75	-6.32222222222222\\
0.760000000000005	-6.52222222222223\\
0.769999999999996	-2.46666666666667\\
0.780000000000001	-2.87777777777778\\
0.790000000000006	0.74444444444444\\
0.799999999999997	-0.36666666666666\\
0.810000000000002	3.71111111111111\\
0.819999999999993	2.94444444444444\\
0.829999999999998	3.44444444444444\\
0.840000000000003	4.32222222222222\\
0.849999999999994	4.2\\
0.859999999999999	3.88888888888889\\
0.870000000000005	4.14444444444445\\
0.879999999999995	3.76666666666667\\
0.890000000000001	6.22222222222223\\
0.909999999999997	8.92222222222222\\
0.920000000000002	14.6111111111111\\
};
\addlegendentry{$p_{1\textrm{a}}$}

\addplot [color=mycolor2, dashed, mark=o, mark options={solid, mycolor2}]
  table[row sep=crcr]{%
0.0100000000000051	50\\
0.019999999999996	34.2777777777778\\
0.0300000000000011	16.1222222222222\\
0.0400000000000063	23.3666666666667\\
0.0499999999999972	12.1888888888889\\
0.0600000000000023	1.58888888888889\\
0.0699999999999932	0\\
0.0799999999999983	0\\
0.0900000000000034	2.05555555555556\\
0.0999999999999943	5.90000000000001\\
0.109999999999999	4.12222222222222\\
0.120000000000005	0\\
0.129999999999995	0.577777777777783\\
0.140000000000001	0\\
0.150000000000006	0\\
0.159999999999997	2.68888888888888\\
0.170000000000002	0\\
0.180000000000007	0\\
0.189999999999998	10.4333333333333\\
0.200000000000003	13.9111111111111\\
0.209999999999994	20.6\\
0.219999999999999	25.8444444444444\\
0.230000000000004	27.4\\
0.239999999999995	34.8333333333333\\
0.25	35.2555555555556\\
0.260000000000005	45.4111111111111\\
0.269999999999996	54.4222222222222\\
0.280000000000001	68.7555555555556\\
0.290000000000006	80.5333333333333\\
0.299999999999997	70.4\\
0.310000000000002	86.8444444444444\\
0.319999999999993	95.3888888888889\\
0.329999999999998	100\\
0.340000000000003	100\\
0.349999999999994	100\\
0.359999999999999	93.8666666666667\\
0.370000000000005	98.7777777777778\\
0.379999999999995	100\\
0.390000000000001	100\\
0.400000000000006	100\\
0.409999999999997	100\\
0.420000000000002	100\\
0.430000000000007	100\\
0.439999999999998	100\\
0.450000000000003	100\\
0.459999999999994	100\\
0.469999999999999	100\\
0.480000000000004	99.2\\
0.489999999999995	100\\
0.5	100\\
0.510000000000005	93.8222222222222\\
0.519999999999996	90.6444444444444\\
0.530000000000001	100\\
0.540000000000006	90.8666666666667\\
0.549999999999997	86.5555555555556\\
0.560000000000002	76.3111111111111\\
0.569999999999993	71.1111111111111\\
0.579999999999998	55.2666666666667\\
0.590000000000003	38.7888888888889\\
0.599999999999994	19.3111111111111\\
0.609999999999999	1.90000000000001\\
0.620000000000005	0\\
0.629999999999995	2.63333333333334\\
0.640000000000001	0\\
0.650000000000006	13.3111111111111\\
0.659999999999997	3.83333333333333\\
0.670000000000002	2.26666666666667\\
0.680000000000007	0\\
0.689999999999998	1.16666666666667\\
0.700000000000003	0\\
0.709999999999994	0\\
0.719999999999999	0\\
0.730000000000004	0\\
0.739999999999995	0\\
0.75	0\\
0.760000000000005	0\\
0.769999999999996	0\\
0.780000000000001	0\\
0.790000000000006	0\\
0.799999999999997	0\\
0.810000000000002	0\\
0.819999999999993	0\\
0.829999999999998	0\\
0.840000000000003	0\\
0.849999999999994	0\\
0.859999999999999	0\\
0.870000000000005	2.64444444444445\\
0.879999999999995	9.51111111111111\\
0.890000000000001	11.0888888888889\\
0.900000000000006	20.2333333333333\\
0.909999999999997	22.5555555555556\\
0.920000000000002	4.47777777777777\\
};
\addlegendentry{$p^\textrm{des}_{1\textrm{a}}$}

\addplot[area legend, draw=none, fill=mycolor2, fill opacity=0.1, forget plot]
table[row sep=crcr] {%
x	y\\
0.6	-20\\
2	-20\\
2	120\\
0.6	120\\
}--cycle;

\addplot[area legend, draw=none, fill=mycolor3, fill opacity=0.1, forget plot]
table[row sep=crcr] {%
x	y\\
0	-20\\
0.6	-20\\
0.6	120\\
0	120\\
}--cycle;
\end{axis}
\end{tikzpicture}%

%% file: figures/appendix/pantago1.tex
%
%
\definecolor{mycolor1}{rgb}{0.00000,0.44700,0.74100}%
\definecolor{mycolor2}{rgb}{0.85000,0.32500,0.09800}%
\definecolor{mycolor3}{rgb}{0.46600,0.67400,0.18800}%
\begin{tikzpicture}

\begin{axis}[%
width=0.951\figwidth,
height=\figheight,
at={(0\figwidth,0\figheight)},
scale only axis,
xmin=0,
xmax=0.89,
ymin=-20,
ymax=120,
ylabel style={font=\color{white!15!black}},
ylabel={$[\%]$},
axis background/.style={fill=white},
ylabel near ticks,
xlabel near ticks,
legend style={at={(.8,.9)},anchor=north west,legend cell align=left,align=left,fill=none,draw=none}
]
\addplot [color=mycolor1, forget plot]
  table[row sep=crcr]{%
0.0100000000000051	53.0555555555556\\
0.019999999999996	50.7333333333333\\
0.0300000000000011	49.4888888888889\\
0.0400000000000063	49.8222222222222\\
0.0600000000000023	49\\
0.0699999999999932	45.0333333333333\\
0.0799999999999983	43.1444444444444\\
0.0900000000000034	27.7\\
0.0999999999999943	19.0222222222222\\
0.109999999999999	21.3666666666667\\
0.120000000000005	9.21111111111111\\
0.129999999999995	2.07777777777778\\
0.140000000000001	10.4111111111111\\
0.150000000000006	14.9\\
0.159999999999997	36.3111111111111\\
0.170000000000002	47.1888888888889\\
0.180000000000007	56.7555555555556\\
0.189999999999998	67.1666666666667\\
0.200000000000003	75.6333333333333\\
0.209999999999994	83.4555555555555\\
0.219999999999999	89.6444444444444\\
0.230000000000004	93.5555555555556\\
0.239999999999995	79.3777777777778\\
0.25	58.0888888888889\\
0.269999999999996	30.9555555555556\\
0.280000000000001	17.8222222222222\\
0.290000000000006	28.1444444444444\\
0.299999999999997	29.4666666666667\\
0.310000000000002	37.5444444444445\\
0.319999999999993	48.0111111111111\\
0.329999999999998	55.8888888888889\\
0.340000000000003	62.6888888888889\\
0.349999999999994	71.0333333333333\\
0.370000000000005	84.3777777777778\\
0.379999999999995	88.7333333333333\\
0.390000000000001	92.3666666666667\\
0.400000000000006	93.8333333333333\\
0.409999999999997	83.7333333333333\\
0.420000000000002	61.3111111111111\\
0.430000000000007	46.3444444444444\\
0.450000000000003	20.4\\
0.459999999999994	9.83333333333333\\
0.469999999999999	1.12222222222222\\
0.480000000000004	-6.51111111111111\\
0.489999999999995	-6.34444444444445\\
0.5	0.0888888888888886\\
0.510000000000005	10.2222222222222\\
0.519999999999996	17.4333333333333\\
0.530000000000001	25.1111111111111\\
0.540000000000006	28.9333333333333\\
0.560000000000002	36.1333333333333\\
0.569999999999993	39.2555555555556\\
0.579999999999998	42.7555555555556\\
0.590000000000003	42.3\\
0.599999999999994	47.5444444444445\\
0.609999999999999	26.7\\
0.620000000000005	14.5333333333333\\
0.629999999999995	3.98888888888889\\
0.640000000000001	-3.53333333333333\\
0.650000000000006	-11.3888888888889\\
0.659999999999997	-3.27777777777777\\
0.670000000000002	-4.62222222222222\\
0.680000000000007	2.35555555555555\\
0.689999999999998	17.0666666666667\\
0.700000000000003	23.1888888888889\\
0.709999999999994	28.9666666666667\\
0.739999999999995	48.0111111111111\\
0.75	53.0555555555556\\
0.760000000000005	58.6888888888889\\
0.769999999999996	62.8222222222222\\
0.780000000000001	66.7222222222222\\
0.790000000000006	71.3444444444444\\
0.810000000000002	76.5333333333333\\
0.819999999999993	74.6555555555556\\
0.829999999999998	76.4444444444444\\
0.840000000000003	57.6555555555556\\
0.849999999999994	40.7444444444445\\
0.859999999999999	29.1555555555556\\
0.870000000000005	17\\
0.879999999999995	5.86666666666666\\
0.890000000000001	-3.22222222222223\\
};
\addplot [color=mycolor2, dashed, mark=o, mark options={solid, mycolor2}, forget plot]
  table[row sep=crcr]{%
0.0100000000000051	50\\
0.019999999999996	53.0666666666667\\
0.0300000000000011	52.5777777777778\\
0.0400000000000063	48.5666666666667\\
0.0499999999999972	42.9111111111111\\
0.0600000000000023	40.9666666666667\\
0.0699999999999932	23.1111111111111\\
0.0799999999999983	26.0888888888889\\
0.0900000000000034	25.2111111111111\\
0.0999999999999943	9.78888888888889\\
0.109999999999999	22.6444444444444\\
0.120000000000005	22.2666666666667\\
0.129999999999995	40.3444444444444\\
0.140000000000001	50.2777777777778\\
0.150000000000006	41.2555555555556\\
0.159999999999997	63.2333333333333\\
0.170000000000002	85.4333333333333\\
0.180000000000007	85.6555555555556\\
0.189999999999998	86.8666666666667\\
0.200000000000003	83.2888888888889\\
0.209999999999994	71.0222222222222\\
0.219999999999999	46.5888888888889\\
0.230000000000004	36.9\\
0.239999999999995	35\\
0.25	28.2222222222222\\
0.260000000000005	40.9666666666667\\
0.269999999999996	40.4444444444444\\
0.280000000000001	56.7666666666667\\
0.290000000000006	79.0111111111111\\
0.299999999999997	95.2777777777778\\
0.310000000000002	100\\
0.319999999999993	100\\
0.329999999999998	100\\
0.340000000000003	81.5666666666667\\
0.349999999999994	92.7222222222222\\
0.359999999999999	100\\
0.370000000000005	100\\
0.379999999999995	76.8\\
0.390000000000001	51.9\\
0.400000000000006	29.2\\
0.409999999999997	13.3444444444444\\
0.420000000000002	0\\
0.430000000000007	0\\
0.439999999999998	0\\
0.450000000000003	0\\
0.459999999999994	10.3888888888889\\
0.469999999999999	15.4777777777778\\
0.480000000000004	36.5444444444445\\
0.489999999999995	34.5444444444445\\
0.5	30.1555555555556\\
0.510000000000005	40.0888888888889\\
0.519999999999996	52.6666666666667\\
0.530000000000001	51.8555555555556\\
0.540000000000006	53.2222222222222\\
0.549999999999997	48.2777777777778\\
0.560000000000002	28.4\\
0.569999999999993	24.5111111111111\\
0.579999999999998	18.6222222222222\\
0.590000000000003	3.76666666666667\\
0.599999999999994	0\\
0.609999999999999	0\\
0.620000000000005	0\\
0.629999999999995	8.95555555555556\\
0.640000000000001	14.8777777777778\\
0.650000000000006	27.8333333333333\\
0.659999999999997	47.1555555555556\\
0.670000000000002	71.0555555555556\\
0.680000000000007	72.2\\
0.689999999999998	87.1666666666667\\
0.700000000000003	98.8444444444444\\
0.709999999999994	91.6333333333333\\
0.719999999999999	100\\
0.730000000000004	100\\
0.739999999999995	100\\
0.75	100\\
0.760000000000005	100\\
0.769999999999996	96.3666666666667\\
0.780000000000001	76.8888888888889\\
0.790000000000006	71.4888888888889\\
0.799999999999997	57.9222222222222\\
0.810000000000002	58.2777777777778\\
0.819999999999993	42.5888888888889\\
0.829999999999998	15.8777777777778\\
0.840000000000003	5.33333333333333\\
0.849999999999994	0\\
0.859999999999999	0\\
0.870000000000005	0\\
0.879999999999995	25.7666666666667\\
0.890000000000001	15.0555555555556\\
};

\addplot[area legend, draw=none, fill=mycolor2, fill opacity=0.1, forget plot]
table[row sep=crcr] {%
x	y\\
0.6	-20\\
2	-20\\
2	120\\
0.6	120\\
}--cycle;

\addplot[area legend, draw=none, fill=mycolor3, fill opacity=0.1, forget plot]
table[row sep=crcr] {%
x	y\\
0	-20\\
0.6	-20\\
0.6	120\\
0	120\\
}--cycle;
\end{axis}
\end{tikzpicture}%

%% file: figures/appendix/pantago1smash.tex
%
%
\definecolor{mycolor1}{rgb}{0.00000,0.44700,0.74100}%
\definecolor{mycolor2}{rgb}{0.85000,0.32500,0.09800}%
\definecolor{mycolor3}{rgb}{0.46600,0.67400,0.18800}%
\begin{tikzpicture}

\begin{axis}[%
width=0.951\figwidth,
height=\figheight,
at={(0\figwidth,0\figheight)},
scale only axis,
xmin=0,
xmax=0.92,
ymin=-20,
ymax=120,
axis background/.style={fill=white},
legend style={legend cell align=left, align=left, fill=none, draw=none},
ylabel near ticks,
xlabel near ticks,
legend style={at={(.8,.7)},anchor=north west,legend cell align=left,align=left,fill=none,draw=none}
]
\addplot [color=mycolor1]
  table[row sep=crcr]{%
0.0100000000000051	48.2777777777778\\
0.019999999999996	49.5\\
0.0300000000000011	49.2888888888889\\
0.0400000000000063	49.7111111111111\\
0.0499999999999972	51.3444444444444\\
0.0600000000000023	70.4111111111111\\
0.0699999999999932	82.0333333333333\\
0.0799999999999983	91.4222222222222\\
0.0900000000000034	99.9333333333333\\
0.0999999999999943	107.744444444444\\
0.109999999999999	114.333333333333\\
0.120000000000005	116.8\\
0.129999999999995	116.7\\
0.140000000000001	115.588888888889\\
0.150000000000006	108.311111111111\\
0.159999999999997	108.366666666667\\
0.170000000000002	95.9\\
0.180000000000007	101.966666666667\\
0.189999999999998	89.1444444444444\\
0.200000000000003	94.2222222222222\\
0.209999999999994	90.2111111111111\\
0.219999999999999	97.6222222222222\\
0.230000000000004	92.9666666666667\\
0.239999999999995	96.4555555555555\\
0.25	88.5888888888889\\
0.260000000000005	91.9777777777778\\
0.269999999999996	90.1777777777778\\
0.280000000000001	93.9888888888889\\
0.290000000000006	90.8\\
0.299999999999997	96.5222222222222\\
0.310000000000002	94.0777777777778\\
0.319999999999993	95.9888888888889\\
0.329999999999998	93.7\\
0.340000000000003	91.0333333333333\\
0.349999999999994	79.8\\
0.359999999999999	73.1777777777778\\
0.370000000000005	63.9111111111111\\
0.379999999999995	48.2555555555556\\
0.390000000000001	37.8\\
0.400000000000006	25.6888888888889\\
0.420000000000002	8.56666666666666\\
0.430000000000007	0.25555555555556\\
0.439999999999998	-3.42222222222222\\
0.459999999999994	-11.0444444444445\\
0.469999999999999	-14.5222222222222\\
0.480000000000004	-10.8111111111111\\
0.489999999999995	-13.6444444444444\\
0.5	-11.8333333333333\\
0.510000000000005	-4.01111111111111\\
0.519999999999996	-11.6888888888889\\
0.530000000000001	-2.59999999999999\\
0.540000000000006	5.85555555555555\\
0.549999999999997	9.23333333333333\\
0.560000000000002	12.3555555555556\\
0.569999999999993	10.8333333333333\\
0.579999999999998	8.05555555555556\\
0.590000000000003	2.02222222222223\\
0.599999999999994	-0.522222222222226\\
0.609999999999999	-0.555555555555557\\
0.620000000000005	-3.06666666666666\\
0.629999999999995	-0.822222222222223\\
0.640000000000001	2.98888888888889\\
0.650000000000006	7.48888888888889\\
0.659999999999997	17.8\\
0.670000000000002	29.9222222222222\\
0.700000000000003	50.4111111111111\\
0.719999999999999	61.3777777777778\\
0.730000000000004	65.3444444444444\\
0.739999999999995	69.6222222222222\\
0.75	74.2777777777778\\
0.760000000000005	77.2111111111111\\
0.769999999999996	80.5111111111111\\
0.790000000000006	83.7444444444444\\
0.799999999999997	84.2222222222222\\
0.810000000000002	86.3222222222222\\
0.819999999999993	87.5666666666667\\
0.829999999999998	90.2555555555556\\
0.840000000000003	92.4777777777778\\
0.849999999999994	94.9333333333333\\
0.870000000000005	97.1777777777778\\
0.879999999999995	98.5222222222222\\
0.890000000000001	98.9333333333333\\
0.900000000000006	92.3111111111111\\
0.909999999999997	93.9222222222222\\
0.920000000000002	88.5555555555556\\
};
\addlegendentry{$p_{1\textrm{b}}$}

\addplot [color=mycolor2, dashed, mark=o, mark options={solid, mycolor2}]
  table[row sep=crcr]{%
0.0100000000000051	50\\
0.019999999999996	54.6\\
0.0300000000000011	64.4777777777778\\
0.0400000000000063	73.9222222222222\\
0.0499999999999972	93.6444444444444\\
0.0600000000000023	100\\
0.0699999999999932	100\\
0.0799999999999983	100\\
0.0900000000000034	100\\
0.0999999999999943	100\\
0.109999999999999	100\\
0.120000000000005	100\\
0.129999999999995	100\\
0.140000000000001	100\\
0.150000000000006	100\\
0.159999999999997	100\\
0.170000000000002	100\\
0.180000000000007	100\\
0.189999999999998	100\\
0.200000000000003	100\\
0.209999999999994	88.4444444444444\\
0.219999999999999	85.9444444444444\\
0.230000000000004	89.4222222222222\\
0.239999999999995	97.8222222222222\\
0.25	98.1666666666667\\
0.260000000000005	100\\
0.269999999999996	93.2111111111111\\
0.280000000000001	96.8888888888889\\
0.290000000000006	91.8888888888889\\
0.299999999999997	88.9111111111111\\
0.310000000000002	85.3666666666667\\
0.319999999999993	83.8555555555556\\
0.329999999999998	81.3555555555556\\
0.340000000000003	73.5666666666667\\
0.349999999999994	64.9222222222222\\
0.359999999999999	53.5555555555556\\
0.370000000000005	37.5777777777778\\
0.379999999999995	30.4444444444444\\
0.390000000000001	11.7777777777778\\
0.400000000000006	0.488888888888894\\
0.409999999999997	0\\
0.420000000000002	0\\
0.430000000000007	0\\
0.439999999999998	0\\
0.450000000000003	0\\
0.459999999999994	0\\
0.469999999999999	3.51111111111111\\
0.480000000000004	0\\
0.489999999999995	4.64444444444445\\
0.5	0\\
0.510000000000005	0\\
0.519999999999996	0\\
0.530000000000001	0\\
0.540000000000006	0\\
0.549999999999997	0\\
0.560000000000002	0\\
0.569999999999993	0\\
0.579999999999998	0\\
0.590000000000003	1.52222222222223\\
0.599999999999994	1.16666666666667\\
0.609999999999999	1.98888888888889\\
0.620000000000005	15.4777777777778\\
0.629999999999995	25.7888888888889\\
0.640000000000001	38.8333333333333\\
0.650000000000006	48.8444444444444\\
0.659999999999997	59.6777777777778\\
0.670000000000002	72.3444444444444\\
0.680000000000007	91.8333333333333\\
0.689999999999998	98.7\\
0.700000000000003	100\\
0.709999999999994	100\\
0.719999999999999	100\\
0.730000000000004	100\\
0.739999999999995	100\\
0.75	100\\
0.760000000000005	100\\
0.769999999999996	100\\
0.780000000000001	100\\
0.790000000000006	98.1\\
0.799999999999997	100\\
0.810000000000002	100\\
0.819999999999993	93\\
0.829999999999998	99.3888888888889\\
0.840000000000003	91.1333333333333\\
0.849999999999994	95.9666666666667\\
0.859999999999999	87.6\\
0.870000000000005	76.5555555555556\\
0.879999999999995	77.2444444444444\\
0.890000000000001	83.3333333333333\\
0.900000000000006	84.8666666666667\\
0.909999999999997	79.2444444444444\\
0.920000000000002	92.2\\
};
\addlegendentry{$p^\textrm{des}_{1\textrm{b}}$}

\addplot[area legend, draw=none, fill=mycolor2, fill opacity=0.1, forget plot]
table[row sep=crcr] {%
x	y\\
0.6	-20\\
2	-20\\
2	120\\
0.6	120\\
}--cycle;

\addplot[area legend, draw=none, fill=mycolor3, fill opacity=0.1, forget plot]
table[row sep=crcr] {%
x	y\\
0	-20\\
0.6	-20\\
0.6	120\\
0	120\\
}--cycle;
\end{axis}
\end{tikzpicture}%

%% file: figures/appendix/pago2.tex
%
%
\definecolor{mycolor1}{rgb}{0.00000,0.44700,0.74100}%
\definecolor{mycolor2}{rgb}{0.85000,0.32500,0.09800}%
\definecolor{mycolor3}{rgb}{0.46600,0.67400,0.18800}%
\begin{tikzpicture}

\begin{axis}[%
width=0.951\figwidth,
height=\figheight,
at={(0\figwidth,0\figheight)},
scale only axis,
xmin=0,
xmax=0.89,
ymin=-20,
ymax=120,
ylabel style={font=\color{white!15!black}},
ylabel={$[\%]$},
axis background/.style={fill=white},
ylabel near ticks,
xlabel near ticks,
legend style={at={(.8,.9)},anchor=north west,legend cell align=left,align=left,fill=none,draw=none}
]
\addplot [color=mycolor1, forget plot]
  table[row sep=crcr]{%
0.00999999999999801	35.9833333333333\\
0.0200000000000031	50\\
0.0300000000000011	50.15\\
0.0399999999999991	31.45\\
0.0499999999999972	9.71666666666667\\
0.0600000000000023	-4.78333333333333\\
0.0700000000000003	-17.2333333333333\\
0.0799999999999983	-22.8833333333333\\
0.0900000000000034	-22.7166666666667\\
0.100000000000001	-17.3833333333333\\
0.109999999999999	-10.7\\
0.119999999999997	6.93333333333334\\
0.130000000000003	19.5666666666667\\
0.140000000000001	11.2666666666667\\
0.149999999999999	6.96666666666667\\
0.159999999999997	4.48333333333333\\
0.170000000000002	1.81666666666667\\
0.18	-0.733333333333334\\
0.189999999999998	-2.36666666666667\\
0.200000000000003	-1.1\\
0.210000000000001	-1.73333333333333\\
0.219999999999999	-1.33333333333334\\
0.229999999999997	-1\\
0.240000000000002	-0.75\\
0.25	-0.0499999999999972\\
0.259999999999998	-0.0499999999999972\\
0.270000000000003	0.633333333333333\\
0.280000000000001	5.88333333333333\\
0.289999999999999	9.98333333333333\\
0.299999999999997	8.88333333333333\\
0.310000000000002	11.3\\
0.32	11.5\\
0.329999999999998	11.9\\
0.340000000000003	12.45\\
0.350000000000001	13.6166666666667\\
0.359999999999999	15.05\\
0.369999999999997	14.9333333333333\\
0.380000000000003	15.1\\
0.390000000000001	20.8666666666667\\
0.399999999999999	34.9666666666667\\
0.409999999999997	44.45\\
0.420000000000002	53.5666666666667\\
0.43	57.9\\
0.439999999999998	44.6\\
0.450000000000003	20.5833333333333\\
0.460000000000001	0.233333333333334\\
0.469999999999999	-11.8333333333333\\
0.479999999999997	-18.3166666666667\\
0.490000000000002	-13.4833333333333\\
0.5	-11.2166666666667\\
0.509999999999998	-1.63333333333333\\
0.520000000000003	3.43333333333333\\
0.530000000000001	7.68333333333334\\
0.539999999999999	10.1166666666667\\
0.549999999999997	10.85\\
0.560000000000002	10.5666666666667\\
0.57	10.1666666666667\\
0.579999999999998	10.3666666666667\\
0.590000000000003	10.35\\
0.600000000000001	8.85\\
0.609999999999999	9.2\\
0.619999999999997	9.3\\
0.630000000000003	9.88333333333333\\
0.640000000000001	7.56666666666667\\
0.649999999999999	7.63333333333333\\
0.659999999999997	8.5\\
0.670000000000002	-0.533333333333331\\
0.68	-2.25\\
0.689999999999998	1.03333333333333\\
0.700000000000003	8.3\\
0.719999999999999	20.4333333333333\\
0.729999999999997	25.2333333333333\\
0.740000000000002	29.5333333333333\\
0.75	32.8166666666667\\
0.759999999999998	36.3166666666667\\
0.770000000000003	38.9333333333333\\
0.780000000000001	40.3833333333333\\
0.789999999999999	42.5666666666667\\
0.799999999999997	44.0333333333333\\
0.810000000000002	48\\
0.82	45.7333333333333\\
0.829999999999998	42.8666666666667\\
0.840000000000003	45.35\\
0.850000000000001	42.35\\
0.859999999999999	40.2\\
0.869999999999997	39.0666666666667\\
0.880000000000003	40.8833333333333\\
0.890000000000001	41.8833333333333\\
};
\addplot [color=mycolor2, dashed, mark=o, mark options={solid, mycolor2}, forget plot]
  table[row sep=crcr]{%
0.0100000000000051	50\\
0.019999999999996	19.55\\
0.0300000000000011	10.8833333333333\\
0.0400000000000063	0\\
0.0499999999999972	0\\
0.0600000000000023	0\\
0.0699999999999932	0\\
0.0799999999999983	0\\
0.0900000000000034	0\\
0.0999999999999943	10.1\\
0.109999999999999	0.716666666666669\\
0.120000000000005	0\\
0.129999999999995	0\\
0.140000000000001	0\\
0.150000000000006	0\\
0.159999999999997	0\\
0.170000000000002	0\\
0.180000000000007	0\\
0.189999999999998	0\\
0.200000000000003	0\\
0.209999999999994	0\\
0.219999999999999	0\\
0.230000000000004	0\\
0.239999999999995	0\\
0.25	10.5\\
0.260000000000005	14.8\\
0.269999999999996	8.83333333333333\\
0.280000000000001	12.85\\
0.290000000000006	0\\
0.299999999999997	6.08333333333333\\
0.310000000000002	3.95\\
0.319999999999993	12.9333333333333\\
0.329999999999998	20.7\\
0.340000000000003	11.15\\
0.349999999999994	7.98333333333333\\
0.359999999999999	21.5166666666667\\
0.370000000000005	33.3166666666667\\
0.379999999999995	34.3\\
0.390000000000001	45.0666666666667\\
0.400000000000006	48.1166666666667\\
0.409999999999997	38.7833333333333\\
0.420000000000002	21.7833333333333\\
0.430000000000007	13.4166666666667\\
0.439999999999998	0\\
0.450000000000003	0\\
0.459999999999994	0\\
0.469999999999999	0\\
0.480000000000004	19.15\\
0.489999999999995	20.6333333333333\\
0.5	19.75\\
0.510000000000005	1.36666666666666\\
0.519999999999996	0\\
0.530000000000001	0.166666666666671\\
0.540000000000006	0\\
0.549999999999997	0.38333333333334\\
0.560000000000002	0\\
0.569999999999993	0\\
0.579999999999998	11.2833333333333\\
0.590000000000003	13.8666666666667\\
0.599999999999994	9.5\\
0.609999999999999	0\\
0.620000000000005	8.65000000000001\\
0.629999999999995	7.88333333333334\\
0.640000000000001	0\\
0.650000000000006	0\\
0.659999999999997	8.13333333333334\\
0.670000000000002	22.5333333333333\\
0.680000000000007	23.8166666666667\\
0.689999999999998	45.5333333333333\\
0.700000000000003	55.5833333333333\\
0.709999999999994	67.6166666666667\\
0.719999999999999	84.9333333333333\\
0.730000000000004	80.75\\
0.739999999999995	99.1666666666667\\
0.75	100\\
0.760000000000005	100\\
0.769999999999996	100\\
0.780000000000001	100\\
0.790000000000006	91.7\\
0.799999999999997	97\\
0.810000000000002	92.15\\
0.819999999999993	83.75\\
0.829999999999998	67.9666666666667\\
0.840000000000003	69.0666666666667\\
0.849999999999994	68.3833333333333\\
0.859999999999999	49.15\\
0.870000000000005	47.7166666666667\\
0.879999999999995	49.25\\
0.890000000000001	52.9166666666667\\
};

\addplot[area legend, draw=none, fill=mycolor2, fill opacity=0.1, forget plot]
table[row sep=crcr] {%
x	y\\
0.6	-20\\
2	-20\\
2	120\\
0.6	120\\
}--cycle;

\addplot[area legend, draw=none, fill=mycolor3, fill opacity=0.1, forget plot]
table[row sep=crcr] {%
x	y\\
0	-20\\
0.6	-20\\
0.6	120\\
0	120\\
}--cycle;
\end{axis}
\end{tikzpicture}%

%% file: figures/appendix/pago2smash.tex
%
%
\definecolor{mycolor1}{rgb}{0.00000,0.44700,0.74100}%
\definecolor{mycolor2}{rgb}{0.85000,0.32500,0.09800}%
\definecolor{mycolor3}{rgb}{0.46600,0.67400,0.18800}%
\begin{tikzpicture}

\begin{axis}[%
width=0.951\figwidth,
height=\figheight,
at={(0\figwidth,0\figheight)},
scale only axis,
unbounded coords=jump,
xmin=0,
xmax=0.92,
ymin=-20,
ymax=120,
axis background/.style={fill=white},
legend style={legend cell align=left, align=left, fill=none, draw=none},
ylabel near ticks,
xlabel near ticks,
legend style={at={(.8,.9)},anchor=north west,legend cell align=left,align=left,fill=none,draw=none}
]
\addplot [color=mycolor1]
  table[row sep=crcr]{%
0.0100000000000051	97.5142857142857\\
0.019999999999996	100.314285714286\\
0.0300000000000011	100.685714285714\\
0.0400000000000063	92.8571428571429\\
0.0499999999999972	60.0857142857143\\
0.0600000000000023	42.6\\
0.0699999999999932	21.7142857142857\\
0.0799999999999983	-2.57142857142857\\
0.0900000000000034	-22.0857142857143\\
nan	nan\\
0.109999999999999	-26.2\\
0.120000000000005	-19.4857142857143\\
0.129999999999995	-20.9142857142857\\
0.140000000000001	-6.37142857142857\\
0.150000000000006	1.82857142857142\\
0.159999999999997	-0.114285714285714\\
0.170000000000002	2.22857142857143\\
0.180000000000007	4.42857142857143\\
0.189999999999998	5.74285714285715\\
0.200000000000003	6.05714285714286\\
0.209999999999994	6.57142857142857\\
0.219999999999999	7.40000000000001\\
0.230000000000004	13.8\\
0.239999999999995	28.3428571428571\\
0.25	35.5428571428571\\
0.260000000000005	30.5714285714286\\
0.269999999999996	29.3714285714286\\
0.280000000000001	6.57142857142857\\
0.290000000000006	-7.65714285714286\\
0.299999999999997	-7.97142857142858\\
0.310000000000002	-3.8\\
0.319999999999993	-6.94285714285715\\
0.329999999999998	1.37142857142857\\
0.340000000000003	20.2857142857143\\
0.349999999999994	44.2\\
0.359999999999999	61.6571428571429\\
0.370000000000005	70.4285714285714\\
0.379999999999995	67.5714285714286\\
0.390000000000001	70.8\\
0.400000000000006	81.6285714285714\\
0.409999999999997	86.7714285714286\\
0.420000000000002	81.5428571428571\\
0.430000000000007	81.7428571428571\\
0.439999999999998	84.2\\
0.450000000000003	85.5142857142857\\
0.459999999999994	94.0285714285714\\
0.469999999999999	93.0285714285714\\
0.480000000000004	91.3428571428571\\
0.489999999999995	72.6571428571429\\
0.5	37.5714285714286\\
0.510000000000005	4.37142857142857\\
0.519999999999996	-13.3714285714286\\
0.530000000000001	-23.6571428571429\\
nan	nan\\
0.549999999999997	-21.1142857142857\\
0.560000000000002	-18.8857142857143\\
0.569999999999993	-21.4571428571429\\
0.579999999999998	-10.5142857142857\\
0.599999999999994	7.48571428571428\\
0.609999999999999	14.2285714285714\\
0.620000000000005	11.6\\
0.629999999999995	12.6\\
0.640000000000001	13.4571428571429\\
0.650000000000006	13.8285714285714\\
0.659999999999997	13.0571428571429\\
0.670000000000002	5.37142857142857\\
0.680000000000007	1.71428571428571\\
0.689999999999998	3.45714285714286\\
0.700000000000003	1.02857142857142\\
0.709999999999994	-4\\
0.719999999999999	-3.85714285714286\\
0.730000000000004	-3.08571428571429\\
0.739999999999995	-4.05714285714286\\
0.75	-4.02857142857142\\
0.760000000000005	-4.11428571428571\\
0.769999999999996	-3.40000000000001\\
0.780000000000001	-4.08571428571429\\
0.790000000000006	-3.59999999999999\\
0.799999999999997	-3.62857142857143\\
0.810000000000002	-2.97142857142858\\
0.819999999999993	-3\\
0.829999999999998	-2.54285714285714\\
0.840000000000003	-2.22857142857143\\
0.849999999999994	-1.8\\
0.859999999999999	-1.48571428571428\\
0.870000000000005	-1\\
0.879999999999995	-0.885714285714286\\
0.890000000000001	0.285714285714292\\
0.900000000000006	16.5428571428571\\
0.909999999999997	21.5142857142857\\
0.920000000000002	17.1714285714286\\
};
\addlegendentry{$p_{2\textrm{a}}$}

\addplot [color=mycolor2, dashed, mark=o, mark options={solid, mycolor2}]
  table[row sep=crcr]{%
0.0100000000000051	100\\
0.019999999999996	83.6571428571429\\
0.0300000000000011	72.7142857142857\\
0.0400000000000063	54.6\\
0.0499999999999972	36.4857142857143\\
0.0600000000000023	0\\
0.0699999999999932	0\\
0.0799999999999983	0\\
0.0900000000000034	0\\
0.0999999999999943	0\\
0.109999999999999	0\\
0.120000000000005	0\\
0.129999999999995	0\\
0.140000000000001	0.228571428571428\\
0.150000000000006	0\\
0.159999999999997	0\\
0.170000000000002	0\\
0.180000000000007	0\\
0.189999999999998	0\\
0.200000000000003	14.8857142857143\\
0.209999999999994	16.7714285714286\\
0.219999999999999	26.9428571428572\\
0.230000000000004	20.8285714285714\\
0.239999999999995	15.1142857142857\\
0.25	5.68571428571428\\
0.260000000000005	0\\
0.269999999999996	0\\
0.280000000000001	0\\
0.290000000000006	0\\
0.299999999999997	10.6285714285714\\
0.310000000000002	21.5428571428571\\
0.319999999999993	36.1142857142857\\
0.329999999999998	39.4285714285714\\
0.340000000000003	55.8571428571429\\
0.349999999999994	47.4285714285714\\
0.359999999999999	54.8\\
0.370000000000005	73\\
0.379999999999995	73.1714285714286\\
0.390000000000001	68\\
0.400000000000006	72.3428571428571\\
0.409999999999997	77.8\\
0.420000000000002	83.5714285714286\\
0.430000000000007	92.3142857142857\\
0.439999999999998	87.6285714285714\\
0.450000000000003	85.8857142857143\\
0.459999999999994	75.5714285714286\\
0.469999999999999	49.6571428571429\\
0.480000000000004	35.7142857142857\\
0.489999999999995	12.3428571428571\\
0.5	7.2\\
0.510000000000005	0\\
0.519999999999996	0\\
0.530000000000001	0\\
0.540000000000006	0\\
0.549999999999997	0\\
0.560000000000002	0\\
0.569999999999993	0\\
0.579999999999998	9.02857142857142\\
0.590000000000003	1.48571428571428\\
0.599999999999994	7.8\\
0.609999999999999	3.40000000000001\\
0.620000000000005	0\\
0.629999999999995	0\\
0.640000000000001	0\\
0.650000000000006	0\\
0.659999999999997	0\\
0.670000000000002	0\\
0.680000000000007	0\\
0.689999999999998	0\\
0.700000000000003	0\\
0.709999999999994	0\\
0.719999999999999	0\\
0.730000000000004	0\\
0.739999999999995	0\\
0.75	0\\
0.760000000000005	0\\
0.769999999999996	0\\
0.780000000000001	0\\
0.790000000000006	0\\
0.799999999999997	0\\
0.810000000000002	0\\
0.819999999999993	0\\
0.829999999999998	0\\
0.840000000000003	0\\
0.849999999999994	0\\
0.859999999999999	0\\
0.870000000000005	10\\
0.879999999999995	12.6571428571429\\
0.890000000000001	0\\
0.900000000000006	8.8\\
0.909999999999997	8.22857142857143\\
0.920000000000002	1.57142857142857\\
};
\addlegendentry{$p^\textrm{des}_{2\textrm{a}}$}

\addplot[area legend, draw=none, fill=mycolor2, fill opacity=0.1, forget plot]
table[row sep=crcr] {%
x	y\\
0.6	-20\\
2	-20\\
2	120\\
0.6	120\\
}--cycle;

\addplot[area legend, draw=none, fill=mycolor3, fill opacity=0.1, forget plot]
table[row sep=crcr] {%
x	y\\
0	-20\\
0.6	-20\\
0.6	120\\
0	120\\
}--cycle;
\end{axis}
\end{tikzpicture}%

%% file: figures/appendix/pantago2.tex
%
%
\definecolor{mycolor1}{rgb}{0.00000,0.44700,0.74100}%
\definecolor{mycolor2}{rgb}{0.85000,0.32500,0.09800}%
\definecolor{mycolor3}{rgb}{0.46600,0.67400,0.18800}%
\begin{tikzpicture}

\begin{axis}[%
width=0.951\figwidth,
height=\figheight,
at={(0\figwidth,0\figheight)},
scale only axis,
xmin=0,
xmax=0.89,
ymin=-20,
ymax=120,
ylabel style={font=\color{white!15!black}},
ylabel={$[\%]$},
axis background/.style={fill=white},
ylabel near ticks,
xlabel near ticks,
legend style={at={(.8,.9)},anchor=north west,legend cell align=left,align=left,fill=none,draw=none}
]
\addplot [color=mycolor1, forget plot]
  table[row sep=crcr]{%
0.0100000000000051	61.28\\
0.019999999999996	49.64\\
0.0300000000000011	49.92\\
0.0400000000000063	52.0133333333333\\
0.0499999999999972	64.1066666666667\\
0.0600000000000023	78.5333333333333\\
0.0699999999999932	91.8933333333333\\
0.0799999999999983	101.733333333333\\
0.0900000000000034	108.373333333333\\
0.0999999999999943	114.093333333333\\
0.109999999999999	115.76\\
0.120000000000005	112.453333333333\\
0.129999999999995	111.88\\
0.140000000000001	104.533333333333\\
0.150000000000006	78.9866666666667\\
0.159999999999997	71.1066666666667\\
0.170000000000002	78.04\\
0.180000000000007	90.4933333333333\\
0.189999999999998	100.626666666667\\
0.200000000000003	106.2\\
0.209999999999994	107.92\\
0.219999999999999	105.68\\
0.230000000000004	103.84\\
0.239999999999995	103.493333333333\\
0.25	103.626666666667\\
0.260000000000005	102.333333333333\\
0.269999999999996	101.426666666667\\
0.280000000000001	100.96\\
0.290000000000006	97.16\\
0.299999999999997	90.1466666666667\\
0.310000000000002	90.1866666666667\\
0.319999999999993	90.48\\
0.329999999999998	90.04\\
0.340000000000003	91.8666666666667\\
0.359999999999999	98.4933333333333\\
0.379999999999995	104.346666666667\\
0.390000000000001	102.106666666667\\
0.400000000000006	98.2933333333333\\
0.409999999999997	93.8266666666667\\
0.420000000000002	88\\
0.430000000000007	81.8933333333333\\
0.439999999999998	79.0933333333333\\
0.450000000000003	77.64\\
0.459999999999994	71.6933333333333\\
0.469999999999999	64.96\\
0.489999999999995	52.6666666666667\\
0.510000000000005	41.7466666666667\\
0.519999999999996	35.48\\
0.530000000000001	30.9066666666667\\
0.540000000000006	27.04\\
0.549999999999997	24.2266666666667\\
0.560000000000002	21.9733333333333\\
0.569999999999993	21.0666666666667\\
0.579999999999998	19.8933333333333\\
0.590000000000003	18.3333333333333\\
0.599999999999994	17.2\\
0.609999999999999	16.56\\
0.620000000000005	16.2\\
0.629999999999995	16.5866666666667\\
0.640000000000001	17.1333333333333\\
0.650000000000006	18.9466666666667\\
0.659999999999997	20.1466666666667\\
0.670000000000002	21.6266666666667\\
0.680000000000007	21.4266666666667\\
0.689999999999998	22.8\\
0.700000000000003	22.1866666666667\\
0.709999999999994	15.2666666666667\\
0.719999999999999	12.3466666666667\\
0.730000000000004	14.88\\
0.739999999999995	14.16\\
0.75	12.9866666666667\\
0.760000000000005	14.24\\
0.769999999999996	17.6266666666667\\
0.780000000000001	19.3733333333333\\
0.790000000000006	23.84\\
0.799999999999997	29.4666666666667\\
0.810000000000002	34.12\\
0.819999999999993	37.16\\
0.829999999999998	39.12\\
0.840000000000003	25.0933333333333\\
0.849999999999994	12.9733333333333\\
0.859999999999999	9.21333333333334\\
0.870000000000005	4.66666666666667\\
0.879999999999995	0.306666666666672\\
0.890000000000001	7.13333333333334\\
};
\addplot [color=mycolor2, dashed, mark=o, mark options={solid, mycolor2}, forget plot]
  table[row sep=crcr]{%
0.0100000000000051	50\\
0.019999999999996	61.8266666666667\\
0.0300000000000011	67.8\\
0.0400000000000063	82.64\\
0.0499999999999972	92.4533333333333\\
0.0600000000000023	100\\
0.0699999999999932	100\\
0.0799999999999983	100\\
0.0900000000000034	100\\
0.0999999999999943	100\\
0.109999999999999	100\\
0.120000000000005	94.7066666666667\\
0.129999999999995	80.1066666666667\\
0.140000000000001	90.0666666666667\\
0.150000000000006	100\\
0.159999999999997	100\\
0.170000000000002	100\\
0.180000000000007	100\\
0.189999999999998	100\\
0.200000000000003	100\\
0.209999999999994	100\\
0.219999999999999	100\\
0.230000000000004	100\\
0.239999999999995	100\\
0.25	100\\
0.260000000000005	100\\
0.269999999999996	89.48\\
0.280000000000001	97.2266666666667\\
0.290000000000006	100\\
0.299999999999997	100\\
0.310000000000002	100\\
0.319999999999993	100\\
0.329999999999998	100\\
0.340000000000003	100\\
0.349999999999994	100\\
0.359999999999999	100\\
0.370000000000005	100\\
0.379999999999995	100\\
0.390000000000001	100\\
0.400000000000006	97.8666666666667\\
0.409999999999997	100\\
0.420000000000002	84.6666666666667\\
0.430000000000007	80.8533333333333\\
0.439999999999998	86.3866666666667\\
0.450000000000003	84.1333333333333\\
0.459999999999994	84.9866666666667\\
0.469999999999999	100\\
0.480000000000004	89.2666666666667\\
0.489999999999995	100\\
0.5	92.8666666666667\\
0.510000000000005	100\\
0.519999999999996	100\\
0.530000000000001	100\\
0.540000000000006	100\\
0.549999999999997	93.1066666666667\\
0.560000000000002	100\\
0.569999999999993	98.7333333333333\\
0.579999999999998	100\\
0.590000000000003	81.1866666666667\\
0.599999999999994	94.3066666666667\\
0.609999999999999	92.6533333333333\\
0.620000000000005	69.5333333333333\\
0.629999999999995	70.76\\
0.640000000000001	63.5066666666667\\
0.650000000000006	35.1066666666667\\
0.659999999999997	12.24\\
0.670000000000002	0\\
0.680000000000007	7.06666666666666\\
0.689999999999998	0\\
0.700000000000003	14.32\\
0.709999999999994	14.1066666666667\\
0.719999999999999	22.28\\
0.730000000000004	18.72\\
0.739999999999995	19.6266666666667\\
0.75	30.2666666666667\\
0.760000000000005	20.7866666666667\\
0.769999999999996	33.5066666666667\\
0.780000000000001	35.0266666666667\\
0.790000000000006	49.0266666666667\\
0.799999999999997	26.4533333333333\\
0.810000000000002	9.57333333333334\\
0.819999999999993	0\\
0.829999999999998	4.34666666666666\\
0.840000000000003	9.37333333333333\\
0.849999999999994	2\\
0.859999999999999	27.5733333333333\\
0.870000000000005	39.4133333333333\\
0.879999999999995	51.44\\
0.890000000000001	64.96\\
};

\addplot[area legend, draw=none, fill=mycolor2, fill opacity=0.1, forget plot]
table[row sep=crcr] {%
x	y\\
0.6	-20\\
2	-20\\
2	120\\
0.6	120\\
}--cycle;

\addplot[area legend, draw=none, fill=mycolor3, fill opacity=0.1, forget plot]
table[row sep=crcr] {%
x	y\\
0	-20\\
0.6	-20\\
0.6	120\\
0	120\\
}--cycle;
\end{axis}
\end{tikzpicture}%

%% file: figures/appendix/pantago2smash.tex
%
%
\definecolor{mycolor1}{rgb}{0.00000,0.44700,0.74100}%
\definecolor{mycolor2}{rgb}{0.85000,0.32500,0.09800}%
\definecolor{mycolor3}{rgb}{0.46600,0.67400,0.18800}%
\begin{tikzpicture}

\begin{axis}[%
width=0.951\figwidth,
height=\figheight,
at={(0\figwidth,0\figheight)},
scale only axis,
xmin=0,
xmax=0.92,
ymin=-20,
ymax=120,
axis background/.style={fill=white},
legend style={legend cell align=left, align=left, fill=none, draw=none},
ylabel near ticks,
xlabel near ticks,
legend style={at={(.8,.9)},anchor=north west,legend cell align=left,align=left,fill=none,draw=none}
]
\addplot [color=mycolor1]
  table[row sep=crcr]{%
0.0100000000000051	41.9866666666667\\
0.019999999999996	49.8666666666667\\
0.0300000000000011	49.9866666666667\\
0.0400000000000063	53.7066666666667\\
0.0499999999999972	71.6533333333333\\
0.0600000000000023	95.2666666666667\\
0.0699999999999932	114.013333333333\\
0.0799999999999983	124.306666666667\\
0.0900000000000034	124.786666666667\\
0.0999999999999943	118.893333333333\\
0.109999999999999	100.106666666667\\
0.120000000000005	94.0533333333333\\
0.129999999999995	98.2\\
0.140000000000001	99.3733333333333\\
0.150000000000006	96.2933333333333\\
0.159999999999997	88.92\\
0.170000000000002	87.8533333333333\\
0.189999999999998	98.8133333333333\\
0.200000000000003	80.84\\
0.209999999999994	71.7466666666667\\
0.219999999999999	72.2666666666667\\
0.230000000000004	63.7333333333333\\
0.239999999999995	43.4666666666667\\
0.260000000000005	11.84\\
0.269999999999996	-1.31999999999999\\
0.280000000000001	-9.57333333333334\\
0.290000000000006	-16.0266666666667\\
0.299999999999997	-20.48\\
0.310000000000002	-21.2266666666667\\
0.319999999999993	-20\\
0.329999999999998	-14.2666666666667\\
0.340000000000003	-5.40000000000001\\
0.349999999999994	-0.986666666666665\\
0.359999999999999	4.28\\
0.370000000000005	15.5866666666667\\
0.379999999999995	22.2533333333333\\
0.390000000000001	24.3066666666667\\
0.400000000000006	17.24\\
0.409999999999997	7.86666666666666\\
0.420000000000002	2.29333333333334\\
0.430000000000007	-0.359999999999999\\
0.439999999999998	-1.72\\
0.450000000000003	-2.23999999999999\\
0.459999999999994	-2.70666666666666\\
0.469999999999999	-1.44\\
0.480000000000004	-1.42666666666666\\
0.489999999999995	-1.22666666666667\\
0.5	8.61333333333333\\
0.510000000000005	19.0133333333333\\
0.519999999999996	30.08\\
0.530000000000001	39.6933333333333\\
0.540000000000006	45.76\\
0.549999999999997	47.4533333333333\\
0.560000000000002	44.6666666666667\\
0.569999999999993	44.6133333333333\\
0.579999999999998	45.4666666666667\\
0.590000000000003	47.1066666666667\\
0.599999999999994	47.9333333333333\\
0.609999999999999	46.84\\
0.620000000000005	46.7066666666667\\
0.629999999999995	48.2133333333333\\
0.640000000000001	50.08\\
0.650000000000006	51.6133333333333\\
0.659999999999997	51.1333333333333\\
0.670000000000002	49.8\\
0.680000000000007	49\\
0.689999999999998	40.7733333333333\\
0.700000000000003	21.8533333333333\\
0.709999999999994	7.33333333333333\\
0.719999999999999	-2.53333333333333\\
0.730000000000004	-9.37333333333333\\
0.739999999999995	-15.0533333333333\\
0.75	-17.4666666666667\\
0.760000000000005	-15.56\\
0.769999999999996	-14.3733333333333\\
0.780000000000001	-12.5066666666667\\
0.790000000000006	-6.62666666666667\\
0.799999999999997	-1.82666666666667\\
0.810000000000002	2.49333333333334\\
0.819999999999993	4.66666666666667\\
0.829999999999998	4.94666666666667\\
0.840000000000003	4.73333333333333\\
0.849999999999994	4.97333333333333\\
0.859999999999999	5.14666666666666\\
0.870000000000005	5.23999999999999\\
0.879999999999995	4.49333333333334\\
0.900000000000006	2.09333333333333\\
0.920000000000002	-1.28\\
};
\addlegendentry{$p_{2\textrm{b}}$}

\addplot [color=mycolor2, dashed, mark=o, mark options={solid, mycolor2}]
  table[row sep=crcr]{%
0.0100000000000051	50\\
0.019999999999996	64.1733333333333\\
0.0300000000000011	88.4666666666667\\
0.0400000000000063	100\\
0.0499999999999972	100\\
0.0600000000000023	100\\
0.0699999999999932	100\\
0.0799999999999983	100\\
0.0900000000000034	100\\
0.0999999999999943	100\\
0.109999999999999	100\\
0.120000000000005	100\\
0.129999999999995	92.44\\
0.140000000000001	93.3466666666667\\
0.150000000000006	100\\
0.159999999999997	100\\
0.170000000000002	86.0666666666667\\
0.180000000000007	75.8666666666667\\
0.189999999999998	87.0933333333333\\
0.200000000000003	67.0933333333333\\
0.209999999999994	67.4266666666667\\
0.219999999999999	41.1466666666667\\
0.230000000000004	28.8133333333333\\
0.239999999999995	7.69333333333333\\
0.25	1.76000000000001\\
0.260000000000005	0\\
0.269999999999996	0\\
0.280000000000001	0\\
0.290000000000006	0\\
0.299999999999997	0\\
0.310000000000002	0\\
0.319999999999993	5.16\\
0.329999999999998	0\\
0.340000000000003	8.14666666666666\\
0.349999999999994	25.6533333333333\\
0.359999999999999	10.1333333333333\\
0.370000000000005	0\\
0.379999999999995	0\\
0.390000000000001	0\\
0.400000000000006	0\\
0.409999999999997	0\\
0.420000000000002	4.45333333333333\\
0.430000000000007	0\\
0.439999999999998	4.14666666666666\\
0.450000000000003	6.8\\
0.459999999999994	4.02666666666667\\
0.469999999999999	4.59999999999999\\
0.480000000000004	15.8\\
0.489999999999995	35.4\\
0.5	51.0133333333333\\
0.510000000000005	63.2\\
0.519999999999996	74.5866666666667\\
0.530000000000001	74.2\\
0.540000000000006	93.1333333333333\\
0.549999999999997	100\\
0.560000000000002	100\\
0.569999999999993	100\\
0.579999999999998	95.0666666666667\\
0.590000000000003	100\\
0.599999999999994	100\\
0.609999999999999	86.36\\
0.620000000000005	75.68\\
0.629999999999995	74.1066666666667\\
0.640000000000001	48.7733333333333\\
0.650000000000006	23.9733333333333\\
0.659999999999997	26.32\\
0.670000000000002	24.6666666666667\\
0.680000000000007	3.85333333333334\\
0.689999999999998	0\\
0.700000000000003	0\\
0.709999999999994	0\\
0.719999999999999	0.640000000000001\\
0.730000000000004	0\\
0.739999999999995	0\\
0.75	0\\
0.760000000000005	0\\
0.769999999999996	0\\
0.780000000000001	0\\
0.790000000000006	0\\
0.799999999999997	0\\
0.810000000000002	0\\
0.819999999999993	0\\
0.829999999999998	0\\
0.840000000000003	0\\
0.849999999999994	0\\
0.859999999999999	0\\
0.870000000000005	0\\
0.879999999999995	0\\
0.890000000000001	0\\
0.900000000000006	0\\
0.909999999999997	7.33333333333333\\
0.920000000000002	0\\
};
\addlegendentry{$p^\textrm{des}_{2\textrm{b}}$}

\addplot[area legend, draw=none, fill=mycolor2, fill opacity=0.1, forget plot]
table[row sep=crcr] {%
x	y\\
0.6	-20\\
2	-20\\
2	120\\
0.6	120\\
}--cycle;

\addplot[area legend, draw=none, fill=mycolor3, fill opacity=0.1, forget plot]
table[row sep=crcr] {%
x	y\\
0	-20\\
0.6	-20\\
0.6	120\\
0	120\\
}--cycle;
\end{axis}
\end{tikzpicture}%

%% file: figures/appendix/pago3.tex
%
%
\definecolor{mycolor1}{rgb}{0.00000,0.44700,0.74100}%
\definecolor{mycolor2}{rgb}{0.85000,0.32500,0.09800}%
\definecolor{mycolor3}{rgb}{0.46600,0.67400,0.18800}%
\begin{tikzpicture}

\begin{axis}[%
width=0.951\figwidth,
height=\figheight,
at={(0\figwidth,0\figheight)},
scale only axis,
xmin=0,
xmax=0.89,
ymin=-20,
ymax=120,
ylabel style={font=\color{white!15!black}},
ylabel={$[\%]$},
axis background/.style={fill=white},
ylabel near ticks,
xlabel near ticks,
legend style={at={(.8,.9)},anchor=north west,legend cell align=left,align=left,fill=none,draw=none}
]
\addplot [color=mycolor1, forget plot]
  table[row sep=crcr]{%
0.0100000000000051	36.62\\
0.019999999999996	49.96\\
0.0300000000000011	49.98\\
0.0499999999999972	35.03\\
0.0600000000000023	28.87\\
0.0699999999999932	24.4\\
0.0900000000000034	12.67\\
0.129999999999995	-3.01000000000001\\
0.140000000000001	-5.84\\
0.150000000000006	-8.78\\
0.159999999999997	-11.15\\
0.170000000000002	-12.51\\
0.180000000000007	-13.03\\
0.189999999999998	-12.79\\
0.200000000000003	-10.55\\
0.209999999999994	-6.48999999999999\\
0.219999999999999	-1.31999999999999\\
0.230000000000004	1.01000000000001\\
0.239999999999995	1.93000000000001\\
0.25	1.78\\
0.260000000000005	1.88\\
0.280000000000001	17.96\\
0.290000000000006	28.02\\
0.310000000000002	44.77\\
0.319999999999993	49.8\\
0.329999999999998	48.23\\
0.340000000000003	45.35\\
0.349999999999994	39.57\\
0.359999999999999	28.77\\
0.370000000000005	19.61\\
0.379999999999995	11.65\\
0.390000000000001	6.78\\
0.400000000000006	3.29000000000001\\
0.420000000000002	-5.63\\
0.430000000000007	-8.7\\
0.439999999999998	-11.2\\
0.450000000000003	-13.47\\
0.459999999999994	-13.45\\
0.469999999999999	-13.29\\
0.480000000000004	-7.68000000000001\\
0.489999999999995	-2.44\\
0.5	-0.769999999999996\\
0.510000000000005	2.8\\
0.519999999999996	2.37\\
0.530000000000001	2.08\\
0.540000000000006	2.95\\
0.549999999999997	2.23999999999999\\
0.560000000000002	2.12\\
0.569999999999993	3.14\\
0.579999999999998	2.5\\
0.599999999999994	14.25\\
0.609999999999999	21.55\\
0.629999999999995	34.89\\
0.640000000000001	40.56\\
0.650000000000006	45.85\\
0.659999999999997	50.21\\
0.680000000000007	56.8\\
0.689999999999998	59.56\\
0.700000000000003	61.5\\
0.709999999999994	62.96\\
0.719999999999999	64.28\\
0.730000000000004	65.4\\
0.739999999999995	66.71\\
0.75	68.25\\
0.760000000000005	69.35\\
0.769999999999996	70.68\\
0.780000000000001	71.35\\
0.790000000000006	73.15\\
0.799999999999997	73.61\\
0.810000000000002	73.85\\
0.819999999999993	68.93\\
0.829999999999998	62.37\\
0.840000000000003	60.35\\
0.849999999999994	61.65\\
0.859999999999999	64.47\\
0.870000000000005	67.65\\
0.879999999999995	69.96\\
0.890000000000001	71.54\\
};
\addplot [color=mycolor2, dashed, mark=o, mark options={solid, mycolor2}, forget plot]
  table[row sep=crcr]{%
0.0100000000000051	38.4615384615385\\
0.019999999999996	27.8076923076923\\
0.0300000000000011	24.4\\
0.0400000000000063	9.1076923076923\\
0.0499999999999972	5.23076923076923\\
0.0600000000000023	0\\
0.0699999999999932	2.38461538461539\\
0.0799999999999983	1.63846153846154\\
0.0900000000000034	1.78461538461538\\
0.0999999999999943	0.41538461538461\\
0.109999999999999	3.86153846153846\\
0.120000000000005	0\\
0.129999999999995	0\\
0.140000000000001	0\\
0.150000000000006	0\\
0.159999999999997	0\\
0.170000000000002	0\\
0.180000000000007	0\\
0.189999999999998	0\\
0.200000000000003	0\\
0.209999999999994	0\\
0.219999999999999	0\\
0.230000000000004	0\\
0.239999999999995	3.04615384615384\\
0.25	8.23846153846154\\
0.260000000000005	16.2153846153846\\
0.269999999999996	19.1769230769231\\
0.280000000000001	20.3307692307692\\
0.290000000000006	25.9692307692308\\
0.299999999999997	25.2076923076923\\
0.310000000000002	31.2538461538462\\
0.319999999999993	28.2615384615385\\
0.329999999999998	25.9615384615385\\
0.340000000000003	13.8846153846154\\
0.349999999999994	8.49230769230769\\
0.359999999999999	0.161538461538456\\
0.370000000000005	0\\
0.379999999999995	0\\
0.390000000000001	0\\
0.400000000000006	0\\
0.409999999999997	0.330769230769235\\
0.420000000000002	0\\
0.430000000000007	0\\
0.439999999999998	3.66153846153846\\
0.450000000000003	2.44615384615385\\
0.459999999999994	3.4076923076923\\
0.469999999999999	0\\
0.480000000000004	0\\
0.489999999999995	0\\
0.5	0\\
0.510000000000005	0\\
0.519999999999996	0\\
0.530000000000001	0\\
0.540000000000006	0\\
0.549999999999997	0\\
0.560000000000002	6.94615384615385\\
0.569999999999993	14.2307692307692\\
0.579999999999998	17.6\\
0.590000000000003	26.9230769230769\\
0.599999999999994	46.4692307692308\\
0.609999999999999	61.0461538461539\\
0.620000000000005	76.9230769230769\\
0.629999999999995	76.9230769230769\\
0.640000000000001	76.9230769230769\\
0.650000000000006	76.9230769230769\\
0.659999999999997	76.9230769230769\\
0.670000000000002	76.9230769230769\\
0.680000000000007	76.9230769230769\\
0.689999999999998	76.9230769230769\\
0.700000000000003	76.9230769230769\\
0.709999999999994	76.9230769230769\\
0.719999999999999	76.9230769230769\\
0.730000000000004	76.9230769230769\\
0.739999999999995	75.2538461538462\\
0.75	65.6307692307692\\
0.760000000000005	60.2923076923077\\
0.769999999999996	52.3076923076923\\
0.780000000000001	45.8230769230769\\
0.790000000000006	42.0692307692308\\
0.799999999999997	43.3307692307692\\
0.810000000000002	45.3692307692308\\
0.819999999999993	53.0769230769231\\
0.829999999999998	54.5307692307692\\
0.840000000000003	60.3846153846154\\
0.849999999999994	56.7923076923077\\
0.859999999999999	59.2461538461538\\
0.870000000000005	58.4923076923077\\
0.879999999999995	61.2846153846154\\
0.890000000000001	56.0923076923077\\
};

\addplot[area legend, draw=none, fill=mycolor2, fill opacity=0.1, forget plot]
table[row sep=crcr] {%
x	y\\
0.6	-20\\
2	-20\\
2	120\\
0.6	120\\
}--cycle;

\addplot[area legend, draw=none, fill=mycolor3, fill opacity=0.1, forget plot]
table[row sep=crcr] {%
x	y\\
0	-20\\
0.6	-20\\
0.6	120\\
0	120\\
}--cycle;
\end{axis}
\end{tikzpicture}%

%% file: figures/appendix/pago3smash.tex
%
%
\definecolor{mycolor1}{rgb}{0.00000,0.44700,0.74100}%
\definecolor{mycolor2}{rgb}{0.85000,0.32500,0.09800}%
\definecolor{mycolor3}{rgb}{0.46600,0.67400,0.18800}%
\begin{tikzpicture}

\begin{axis}[%
width=0.951\figwidth,
height=\figheight,
at={(0\figwidth,0\figheight)},
scale only axis,
xmin=0,
xmax=0.92,
ymin=-20,
ymax=120,
axis background/.style={fill=white},
legend style={legend cell align=left, align=left, fill=none, draw=none},
ylabel near ticks,
xlabel near ticks,
legend style={at={(.8,.9)},anchor=north west,legend cell align=left,align=left,fill=none,draw=none}
]
\addplot [color=mycolor1]
  table[row sep=crcr]{%
0.00999999999999801	27.0230769230769\\
0.0200000000000031	38.5538461538462\\
0.0300000000000011	38.5923076923077\\
0.0399999999999991	35.2923076923077\\
0.0499999999999972	27.5076923076923\\
0.0600000000000023	21.8461538461538\\
0.0799999999999983	14.4615384615385\\
0.0900000000000034	9.67692307692307\\
0.100000000000001	6.69230769230769\\
0.109999999999999	4.18461538461538\\
0.130000000000003	-2.07692307692308\\
0.149999999999999	-5.63076923076923\\
0.159999999999997	-7.67692307692307\\
0.170000000000002	-7.84615384615385\\
0.18	-7.64615384615384\\
0.189999999999998	-6.5\\
0.200000000000003	-1.30769230769231\\
0.210000000000001	0.884615384615387\\
0.219999999999999	2.43076923076923\\
0.229999999999997	3.09230769230769\\
0.240000000000002	2.33846153846154\\
0.25	2.11538461538461\\
0.259999999999998	3.01538461538462\\
0.270000000000003	2.49230769230769\\
0.280000000000001	1.75384615384615\\
0.289999999999999	1.68461538461538\\
0.299999999999997	0.853846153846156\\
0.310000000000002	-0.0615384615384613\\
0.32	0.0769230769230802\\
0.329999999999998	-0.738461538461536\\
0.340000000000003	-1.03846153846154\\
0.350000000000001	-0.607692307692311\\
0.359999999999999	-0.730769230769234\\
0.369999999999997	-0.638461538461542\\
0.380000000000003	-0.492307692307691\\
0.390000000000001	-0.630769230769232\\
0.399999999999999	-0.5\\
0.409999999999997	0.338461538461537\\
0.420000000000002	-0.376923076923077\\
0.43	0.876923076923077\\
0.439999999999998	1.55384615384615\\
0.450000000000003	1.13846153846154\\
0.460000000000001	1.33846153846154\\
0.469999999999999	1.43846153846154\\
0.479999999999997	1.45384615384615\\
0.490000000000002	1.57692307692308\\
0.5	1.76153846153846\\
0.509999999999998	1.09230769230769\\
0.520000000000003	0.846153846153847\\
0.530000000000001	0.746153846153845\\
0.539999999999999	0.215384615384615\\
0.549999999999997	3.76923076923077\\
0.560000000000002	2.69230769230769\\
0.57	3.36923076923077\\
0.579999999999998	4.94615384615385\\
0.590000000000003	2.25384615384615\\
0.600000000000001	0.45384615384615\\
0.609999999999999	0.338461538461537\\
0.619999999999997	-1.23076923076923\\
0.630000000000003	-1.90769230769231\\
0.640000000000001	-0.938461538461539\\
0.649999999999999	-1.49230769230769\\
0.659999999999997	-1.48461538461539\\
0.670000000000002	-0.776923076923076\\
0.68	-1.38461538461539\\
0.689999999999998	-1.18461538461538\\
0.700000000000003	0.669230769230772\\
0.710000000000001	6.85384615384616\\
0.719999999999999	11.1692307692308\\
0.729999999999997	10.1615384615385\\
0.740000000000002	8.38461538461539\\
0.759999999999998	2.08461538461538\\
0.770000000000003	0.746153846153845\\
0.780000000000001	-1.6\\
0.789999999999999	-3.39230769230769\\
0.799999999999997	-3.39230769230769\\
0.810000000000002	-3.62307692307692\\
0.82	-3.60769230769231\\
0.829999999999998	-2.68461538461538\\
0.840000000000003	-2.95384615384616\\
0.850000000000001	0.623076923076923\\
0.859999999999999	1.81538461538462\\
0.869999999999997	0.792307692307695\\
0.880000000000003	1.63846153846154\\
0.890000000000001	3.22307692307692\\
0.899999999999999	5.10769230769231\\
0.909999999999997	4.6\\
0.920000000000002	13.0076923076923\\
};
\addlegendentry{$p_{3\textrm{a}}$}

\addplot [color=mycolor2, dashed, mark=o, mark options={solid, mycolor2}]
  table[row sep=crcr]{%
0.00999999999999801	38.4615384615385\\
0.0200000000000031	32.8076923076923\\
0.0300000000000011	17.5538461538462\\
0.0399999999999991	3.61538461538461\\
0.0499999999999972	0\\
0.0600000000000023	0\\
0.0700000000000003	0\\
0.0799999999999983	0\\
0.0900000000000034	3.62307692307692\\
0.100000000000001	6.15384615384615\\
0.109999999999999	0\\
0.119999999999997	4.70769230769231\\
0.130000000000003	0\\
0.140000000000001	0\\
0.149999999999999	2.97692307692308\\
0.159999999999997	2.71538461538461\\
0.170000000000002	3.7\\
0.18	0\\
0.189999999999998	2.12307692307692\\
0.200000000000003	0\\
0.210000000000001	0\\
0.219999999999999	0\\
0.229999999999997	0\\
0.240000000000002	2.06153846153846\\
0.25	0\\
0.259999999999998	0\\
0.270000000000003	1.26153846153846\\
0.280000000000001	0\\
0.289999999999999	0\\
0.299999999999997	0\\
0.310000000000002	0\\
0.32	0\\
0.329999999999998	0\\
0.340000000000003	0\\
0.350000000000001	0\\
0.359999999999999	0\\
0.369999999999997	0\\
0.380000000000003	0\\
0.390000000000001	5.84615384615385\\
0.399999999999999	3.33076923076923\\
0.409999999999997	0\\
0.420000000000002	0\\
0.43	0\\
0.439999999999998	0\\
0.450000000000003	0\\
0.460000000000001	0\\
0.469999999999999	0\\
0.479999999999997	0\\
0.490000000000002	0\\
0.5	0\\
0.509999999999998	1.03846153846154\\
0.520000000000003	0\\
0.530000000000001	12.4307692307692\\
0.539999999999999	1.61538461538461\\
0.549999999999997	0\\
0.560000000000002	0\\
0.57	0\\
0.579999999999998	0\\
0.590000000000003	0\\
0.600000000000001	0\\
0.609999999999999	0\\
0.619999999999997	0\\
0.630000000000003	0\\
0.640000000000001	0\\
0.649999999999999	0\\
0.659999999999997	0\\
0.670000000000002	0\\
0.68	8.2\\
0.689999999999998	12.0692307692308\\
0.700000000000003	5.79230769230769\\
0.710000000000001	0\\
0.719999999999999	0\\
0.729999999999997	1.20769230769231\\
0.740000000000002	0\\
0.75	0\\
0.759999999999998	0\\
0.770000000000003	0\\
0.780000000000001	0\\
0.789999999999999	0\\
0.799999999999997	0.723076923076924\\
0.810000000000002	0\\
0.82	1.06923076923077\\
0.829999999999998	4.21538461538461\\
0.840000000000003	0\\
0.850000000000001	0\\
0.859999999999999	0\\
0.869999999999997	7.10769230769231\\
0.880000000000003	0.223076923076924\\
0.890000000000001	1.61538461538461\\
0.899999999999999	15.4230769230769\\
0.909999999999997	22.9076923076923\\
0.920000000000002	23.0461538461538\\
};
\addlegendentry{$p^\textrm{des}_{3\textrm{a}}$}

\addplot[area legend, draw=none, fill=mycolor2, fill opacity=0.1, forget plot]
table[row sep=crcr] {%
x	y\\
0.6	-20\\
2	-20\\
2	120\\
0.6	120\\
}--cycle;

\addplot[area legend, draw=none, fill=mycolor3, fill opacity=0.1, forget plot]
table[row sep=crcr] {%
x	y\\
0	-20\\
0.6	-20\\
0.6	120\\
0	120\\
}--cycle;
\end{axis}
\end{tikzpicture}%

%% file: figures/appendix/pantago3.tex
%
%
\definecolor{mycolor1}{rgb}{0.00000,0.44700,0.74100}%
\definecolor{mycolor2}{rgb}{0.85000,0.32500,0.09800}%
\definecolor{mycolor3}{rgb}{0.46600,0.67400,0.18800}%
\begin{tikzpicture}

\begin{axis}[%
width=0.951\figwidth,
height=\figheight,
at={(0\figwidth,0\figheight)},
scale only axis,
xmin=0,
xmax=0.89,
ymin=-20,
ymax=120,
ylabel style={font=\color{white!15!black}},
ylabel={$[\%]$},
axis background/.style={fill=white},
ylabel near ticks,
xlabel near ticks,
legend style={at={(.8,.9)},anchor=north west,legend cell align=left,align=left,fill=none,draw=none}
]
\addplot [color=mycolor1, forget plot]
  table[row sep=crcr]{%
0.0100000000000051	52.58\\
0.019999999999996	49.81\\
0.0300000000000011	49.85\\
0.0400000000000063	64.43\\
0.0499999999999972	82.99\\
0.0600000000000023	97.58\\
0.0699999999999932	107.09\\
0.0799999999999983	112.57\\
0.0900000000000034	114.2\\
0.0999999999999943	110.97\\
0.109999999999999	108.48\\
0.129999999999995	99.54\\
0.140000000000001	97.24\\
0.150000000000006	98.85\\
0.159999999999997	97.91\\
0.170000000000002	95.3\\
0.180000000000007	94.56\\
0.189999999999998	94.35\\
0.200000000000003	93.32\\
0.209999999999994	92.91\\
0.219999999999999	91.74\\
0.230000000000004	90.73\\
0.239999999999995	84.59\\
0.25	77.49\\
0.260000000000005	73.44\\
0.269999999999996	74.84\\
0.280000000000001	82.51\\
0.290000000000006	83.23\\
0.299999999999997	81.82\\
0.310000000000002	65.49\\
0.319999999999993	55.43\\
0.329999999999998	49.41\\
0.340000000000003	40.86\\
0.349999999999994	39.08\\
0.359999999999999	30.93\\
0.370000000000005	25.75\\
0.379999999999995	21.18\\
0.390000000000001	16.08\\
0.400000000000006	10.59\\
0.409999999999997	5.98999999999999\\
0.420000000000002	3.42\\
0.430000000000007	-1.17\\
0.439999999999998	-3.69\\
0.450000000000003	4.84999999999999\\
0.459999999999994	9.09\\
0.469999999999999	14.94\\
0.480000000000004	23.35\\
0.489999999999995	30.78\\
0.5	34.21\\
0.510000000000005	36.74\\
0.519999999999996	41.52\\
0.530000000000001	45.01\\
0.540000000000006	46.51\\
0.549999999999997	48.38\\
0.560000000000002	51.64\\
0.569999999999993	54.09\\
0.579999999999998	54.97\\
0.590000000000003	55.8\\
0.599999999999994	56.94\\
0.609999999999999	57.29\\
0.620000000000005	57.25\\
0.629999999999995	57.91\\
0.640000000000001	58.44\\
0.650000000000006	59.62\\
0.659999999999997	60.19\\
0.670000000000002	61.19\\
0.680000000000007	62.81\\
0.689999999999998	63.7\\
0.700000000000003	64.1\\
0.709999999999994	64.68\\
0.719999999999999	65.32\\
0.730000000000004	66.32\\
0.739999999999995	67.44\\
0.75	68.77\\
0.760000000000005	70.36\\
0.769999999999996	71.78\\
0.780000000000001	73.44\\
0.790000000000006	74.38\\
0.799999999999997	75.98\\
0.810000000000002	66.58\\
0.819999999999993	55.74\\
0.829999999999998	46.78\\
0.840000000000003	40.13\\
0.849999999999994	33.17\\
0.859999999999999	26.83\\
0.870000000000005	21.46\\
0.879999999999995	18.87\\
0.890000000000001	13.38\\
};
\addplot [color=mycolor2, dashed, mark=o, mark options={solid, mycolor2}, forget plot]
  table[row sep=crcr]{%
0.0100000000000051	50\\
0.019999999999996	85.64\\
0.0300000000000011	100\\
0.0400000000000063	100\\
0.0499999999999972	98.23\\
0.0600000000000023	100\\
0.0699999999999932	100\\
0.0799999999999983	100\\
0.0900000000000034	100\\
0.0999999999999943	100\\
0.109999999999999	100\\
0.120000000000005	100\\
0.129999999999995	100\\
0.140000000000001	100\\
0.150000000000006	100\\
0.159999999999997	100\\
0.170000000000002	93.66\\
0.180000000000007	100\\
0.189999999999998	94.26\\
0.200000000000003	87.48\\
0.209999999999994	88.51\\
0.219999999999999	78.6\\
0.230000000000004	81.99\\
0.239999999999995	75.65\\
0.25	99.21\\
0.260000000000005	100\\
0.269999999999996	73.18\\
0.280000000000001	63.42\\
0.290000000000006	36.17\\
0.299999999999997	26.73\\
0.310000000000002	0\\
0.319999999999993	0\\
0.329999999999998	34.58\\
0.340000000000003	8.92\\
0.349999999999994	19.64\\
0.359999999999999	0\\
0.370000000000005	0\\
0.379999999999995	0\\
0.390000000000001	0\\
0.400000000000006	5.65000000000001\\
0.409999999999997	0\\
0.420000000000002	14.3\\
0.430000000000007	34.32\\
0.439999999999998	49.26\\
0.450000000000003	50.44\\
0.459999999999994	51.68\\
0.469999999999999	39.1\\
0.480000000000004	44.2\\
0.489999999999995	61.45\\
0.5	90.2\\
0.510000000000005	72.45\\
0.519999999999996	65.56\\
0.530000000000001	63.49\\
0.540000000000006	60.31\\
0.549999999999997	68.35\\
0.560000000000002	73\\
0.569999999999993	79.2\\
0.579999999999998	77.48\\
0.590000000000003	100\\
0.599999999999994	100\\
0.609999999999999	100\\
0.620000000000005	100\\
0.629999999999995	100\\
0.640000000000001	100\\
0.650000000000006	100\\
0.659999999999997	100\\
0.670000000000002	100\\
0.680000000000007	100\\
0.689999999999998	100\\
0.700000000000003	100\\
0.709999999999994	100\\
0.719999999999999	100\\
0.730000000000004	81.88\\
0.739999999999995	100\\
0.75	100\\
0.760000000000005	74.23\\
0.769999999999996	74.7\\
0.780000000000001	59.42\\
0.790000000000006	47.52\\
0.799999999999997	37.98\\
0.810000000000002	39.5\\
0.819999999999993	29.9\\
0.829999999999998	31.56\\
0.840000000000003	21\\
0.849999999999994	36.07\\
0.859999999999999	22.97\\
0.870000000000005	18.66\\
0.879999999999995	35.87\\
0.890000000000001	29.05\\
};

\addplot[area legend, draw=none, fill=mycolor2, fill opacity=0.1, forget plot]
table[row sep=crcr] {%
x	y\\
0.6	-20\\
2	-20\\
2	120\\
0.6	120\\
}--cycle;

\addplot[area legend, draw=none, fill=mycolor3, fill opacity=0.1, forget plot]
table[row sep=crcr] {%
x	y\\
0	-20\\
0.6	-20\\
0.6	120\\
0	120\\
}--cycle;
\end{axis}
\end{tikzpicture}%

%% file: figures/appendix/pantago3smash.tex
%
%
\definecolor{mycolor1}{rgb}{0.00000,0.44700,0.74100}%
\definecolor{mycolor2}{rgb}{0.85000,0.32500,0.09800}%
\definecolor{mycolor3}{rgb}{0.46600,0.67400,0.18800}%
\begin{tikzpicture}

\begin{axis}[%
width=0.951\figwidth,
height=\figheight,
at={(0\figwidth,0\figheight)},
scale only axis,
xmin=0,
xmax=0.92,
ymin=-20,
ymax=120,
axis background/.style={fill=white},
legend style={legend cell align=left, align=left, fill=none, draw=none},
ylabel near ticks,
xlabel near ticks,
legend style={at={(.8,.9)},anchor=north west,legend cell align=left,align=left,fill=none,draw=none}
]
\addplot [color=mycolor1]
  table[row sep=crcr]{%
0.0100000000000051	56.4393509588108\\
0.019999999999996	56.9159196641325\\
0.0300000000000011	56.8818790423238\\
0.0499999999999972	36.9227277884943\\
0.0600000000000023	31.9641438783615\\
0.0699999999999932	25.7347100873709\\
0.0799999999999983	18.8925451038239\\
0.0900000000000034	12.3907863383638\\
0.0999999999999943	8.99807103143084\\
0.109999999999999	3.52887779416771\\
0.120000000000005	-1.46374673777375\\
0.129999999999995	-5.00397140587768\\
0.140000000000001	-8.07897424259616\\
0.150000000000006	-11.7213207761262\\
0.159999999999997	-13.5368206059231\\
0.170000000000002	-13.4687393623057\\
0.180000000000007	-13.0829456484739\\
0.189999999999998	-12.6063769431522\\
0.200000000000003	-3.96005900374446\\
0.209999999999994	-3.66504028140247\\
0.219999999999999	9.61080222398729\\
0.230000000000004	12.254623851129\\
0.239999999999995	8.52150232610916\\
0.25	14.5013048905027\\
0.260000000000005	11.9255645069783\\
0.269999999999996	2.32610915692727\\
0.280000000000001	2.16725292182004\\
0.290000000000006	1.74741858617951\\
0.299999999999997	-4.69760580959945\\
0.310000000000002	-5.09474639736753\\
0.319999999999993	-3.48349029842278\\
0.329999999999998	-5.49188698513559\\
0.340000000000003	-4.89050266651537\\
0.349999999999994	-3.44944967661409\\
0.359999999999999	-4.69760580959945\\
0.370000000000005	-3.57426528991263\\
0.379999999999995	4.93589016226029\\
0.390000000000001	4.40258708725746\\
0.400000000000006	3.32463406331556\\
0.409999999999997	7.82934301599909\\
0.420000000000002	5.12878701917622\\
0.430000000000007	4.43662770906616\\
0.439999999999998	7.97685237717009\\
0.450000000000003	7.14853057982526\\
0.459999999999994	2.93884034948371\\
0.469999999999999	5.6053557244979\\
0.480000000000004	14.0587768069897\\
0.489999999999995	6.17269942130943\\
0.5	8.98672415749461\\
0.510000000000005	4.7543401792806\\
0.519999999999996	-0.601384318620219\\
0.530000000000001	1.53182798139113\\
0.540000000000006	0.181549982979689\\
0.549999999999997	2.39419040054464\\
0.560000000000002	10.0533303075003\\
0.569999999999993	12.1638488596392\\
0.579999999999998	14.9438329740157\\
0.590000000000003	16.9635765346647\\
0.599999999999994	22.1717916713945\\
0.620000000000005	33.7796437081584\\
0.629999999999995	42.4826960172472\\
0.640000000000001	48.2355611029162\\
0.650000000000006	52.4906388290026\\
0.659999999999997	57.8350164529672\\
0.670000000000002	64.7679564280041\\
0.680000000000007	53.3530012481561\\
0.689999999999998	45.1945988880064\\
0.700000000000003	38.6587995007375\\
0.709999999999994	32.9172812890049\\
0.719999999999999	22.8072166118234\\
0.730000000000004	17.1564733915806\\
0.739999999999995	13.2077612617724\\
0.75	7.40950868035856\\
0.760000000000005	0.896403040962213\\
0.769999999999996	-2.53035288777942\\
0.780000000000001	-6.09327130375581\\
0.790000000000006	-11.0972427096335\\
0.799999999999997	-13.0942925224101\\
0.810000000000002	-14.4786111426302\\
0.819999999999993	-15.5452172926359\\
0.829999999999998	-14.0928174287984\\
0.840000000000003	-13.287189379326\\
0.849999999999994	-6.05923068194713\\
0.859999999999999	-2.85941223193011\\
0.870000000000005	2.1559060478838\\
0.879999999999995	4.05083399523431\\
0.890000000000001	3.42675592874163\\
0.900000000000006	3.1544309542721\\
0.909999999999997	4.43662770906616\\
0.920000000000002	8.47611483036424\\
};
\addlegendentry{$p_{3\textrm{b}}$}

\addplot [color=mycolor2, dashed, mark=o, mark options={solid, mycolor2}]
  table[row sep=crcr]{%
0.0100000000000051	56.7343696811528\\
0.019999999999996	29.7628503347328\\
0.0300000000000011	0\\
0.0400000000000063	0\\
0.0499999999999972	0\\
0.0600000000000023	0\\
0.0699999999999932	0\\
0.0799999999999983	0\\
0.0900000000000034	0\\
0.0999999999999943	0\\
0.109999999999999	0\\
0.120000000000005	0\\
0.129999999999995	0\\
0.140000000000001	3.5629184159764\\
0.150000000000006	0\\
0.159999999999997	0\\
0.170000000000002	0\\
0.180000000000007	2.92749347554749\\
0.189999999999998	0\\
0.200000000000003	18.5975263814819\\
0.209999999999994	0\\
0.219999999999999	0\\
0.230000000000004	0\\
0.239999999999995	0\\
0.25	0\\
0.260000000000005	0\\
0.269999999999996	0\\
0.280000000000001	0\\
0.290000000000006	0\\
0.299999999999997	3.46079655055033\\
0.310000000000002	0\\
0.319999999999993	5.41245886758198\\
0.329999999999998	0\\
0.340000000000003	0.0340406218086855\\
0.349999999999994	0\\
0.359999999999999	9.8717803245206\\
0.370000000000005	0\\
0.379999999999995	0\\
0.390000000000001	0\\
0.400000000000006	10.8816521048451\\
0.409999999999997	0\\
0.420000000000002	4.10756836491547\\
0.430000000000007	0\\
0.439999999999998	0\\
0.450000000000003	18.5407920118007\\
0.459999999999994	0\\
0.469999999999999	5.23090888460229\\
0.480000000000004	0\\
0.489999999999995	0\\
0.5	0\\
0.510000000000005	13.8658799500738\\
0.519999999999996	1.20276863724044\\
0.530000000000001	26.3928287756723\\
0.540000000000006	6.78543061386588\\
0.549999999999997	16.191989107001\\
0.560000000000002	0\\
0.569999999999993	21.751957335754\\
0.579999999999998	32.3953250879383\\
0.590000000000003	59.4803131737206\\
0.599999999999994	85.6688982185408\\
0.609999999999999	100\\
0.620000000000005	87.302848065358\\
0.629999999999995	76.2169522296607\\
0.640000000000001	67.9677748780211\\
0.650000000000006	46.4541018949279\\
0.659999999999997	21.309429252241\\
0.670000000000002	7.28469306706002\\
0.680000000000007	0\\
0.689999999999998	0\\
0.700000000000003	0\\
0.709999999999994	0\\
0.719999999999999	0\\
0.730000000000004	0\\
0.739999999999995	0\\
0.75	0\\
0.760000000000005	0\\
0.769999999999996	0\\
0.780000000000001	0\\
0.790000000000006	0\\
0.799999999999997	0\\
0.810000000000002	0\\
0.819999999999993	0\\
0.829999999999998	0\\
0.840000000000003	0\\
0.849999999999994	0\\
0.859999999999999	0\\
0.870000000000005	0\\
0.879999999999995	0\\
0.890000000000001	8.72574605696131\\
0.900000000000006	0\\
0.909999999999997	6.55849313514126\\
0.920000000000002	32.4066719618745\\
};
\addlegendentry{$p^\textrm{des}_{3\textrm{b}}$}

\addplot[area legend, draw=none, fill=mycolor2, fill opacity=0.1, forget plot]
table[row sep=crcr] {%
x	y\\
0.6	-20\\
2	-20\\
2	120\\
0.6	120\\
}--cycle;

\addplot[area legend, draw=none, fill=mycolor3, fill opacity=0.1, forget plot]
table[row sep=crcr] {%
x	y\\
0	-20\\
0.6	-20\\
0.6	120\\
0	120\\
}--cycle;
\end{axis}
\end{tikzpicture}%

%% file: figures/appendix/pago4.tex
%
%
\definecolor{mycolor1}{rgb}{0.00000,0.44700,0.74100}%
\definecolor{mycolor2}{rgb}{0.85000,0.32500,0.09800}%
\definecolor{mycolor3}{rgb}{0.46600,0.67400,0.18800}%
\begin{tikzpicture}

\begin{axis}[%
width=0.951\figwidth,
height=\figheight,
at={(0\figwidth,0\figheight)},
scale only axis,
xmin=0,
xmax=0.89,
ymin=-20,
ymax=120,
ylabel style={font=\color{white!15!black}},
ylabel={$[\%]$},
axis background/.style={fill=white},
ylabel near ticks,
xlabel near ticks,
legend style={at={(.8,.9)},anchor=north west,legend cell align=left,align=left,fill=none,draw=none}
]
\addplot [color=mycolor1, forget plot]
  table[row sep=crcr]{%
0.00999999999999801	40.6551827143284\\
0.0200000000000031	54.9736134621129\\
0.0300000000000011	54.8342128845962\\
0.0399999999999991	54.4060539679379\\
0.0499999999999972	44.6679279099871\\
0.0600000000000023	36.204321417903\\
0.0700000000000003	30.3992830827442\\
0.0900000000000034	19.7749676391516\\
0.100000000000001	13.8404859105845\\
0.109999999999999	9.89744100368416\\
0.119999999999997	6.27302598825052\\
0.130000000000003	1.41392014338345\\
0.140000000000001	-1.90182216469182\\
0.149999999999999	-4.73961963556706\\
0.159999999999997	-8.11510504829234\\
0.170000000000002	-10.6342726277009\\
0.189999999999998	-13.1932689435428\\
0.200000000000003	-12.9841680772677\\
0.210000000000001	-12.1178930598427\\
0.219999999999999	-7.96574728666733\\
0.229999999999997	-1.29443393408344\\
0.240000000000002	1.33426267051678\\
0.25	2.53908194762521\\
0.259999999999998	3.87334461814199\\
0.270000000000003	2.83779747087524\\
0.280000000000001	2.31006671313352\\
0.289999999999999	4.13223140495868\\
0.310000000000002	2.31006671313352\\
0.32	3.15642736234193\\
0.329999999999998	1.6031066414418\\
0.340000000000003	-0.199143682166685\\
0.350000000000001	-0.119486209300007\\
0.359999999999999	-0.945932490291746\\
0.369999999999997	-1.02558996315842\\
0.380000000000003	3.97291645922533\\
0.390000000000001	6.45225530220054\\
0.399999999999999	3.05685552125859\\
0.409999999999997	5.35696505028378\\
0.420000000000002	1.54336353679179\\
0.43	0.288758339141694\\
0.439999999999998	2.43951010654187\\
0.450000000000003	1.74250721895848\\
0.460000000000001	8.20471970526735\\
0.469999999999999	11.2416608583093\\
0.479999999999997	6.89037140296724\\
0.490000000000002	9.9970128447675\\
0.5	16.9072986159514\\
0.509999999999998	17.6043015035348\\
0.520000000000003	24.6340734840187\\
0.530000000000001	31.1560290749776\\
0.539999999999999	33.1275515284278\\
0.549999999999997	34.9696305884696\\
0.560000000000002	39.74907896047\\
0.57	43.712038235587\\
0.579999999999998	44.7575425669621\\
0.590000000000003	47.645125958379\\
0.600000000000001	51.0703972916459\\
0.609999999999999	52.8527332470377\\
0.619999999999997	54.8242557004879\\
0.630000000000003	56.347704869063\\
0.640000000000001	57.5226525938465\\
0.649999999999999	58.8370008961466\\
0.659999999999997	60.0318629891467\\
0.670000000000002	60.8184805337051\\
0.68	59.4643034949716\\
0.689999999999998	47.5355969331873\\
0.700000000000003	37.5784128248531\\
0.710000000000001	31.554316439311\\
0.719999999999999	26.7748680673106\\
0.729999999999997	20.1632978193767\\
0.740000000000002	14.3084735636762\\
0.75	9.86756945135916\\
0.759999999999998	6.55182714328388\\
0.770000000000003	1.83212187593349\\
0.780000000000001	-1.32430548640844\\
0.789999999999999	-0.726874439908393\\
0.799999999999997	-1.38404859105845\\
0.810000000000002	3.41531414915862\\
0.82	9.9970128447675\\
0.829999999999998	16.240167280693\\
0.840000000000003	21.7863188290352\\
0.850000000000001	28.9853629393607\\
0.859999999999999	34.4219854625112\\
0.869999999999997	37.3991835109031\\
0.880000000000003	42.4076471173952\\
0.890000000000001	46.9481230707956\\
};
\addplot [color=mycolor2, dashed, mark=o, mark options={solid, mycolor2}, forget plot]
  table[row sep=crcr]{%
0.0100000000000051	54.7645125958379\\
0.019999999999996	52.8626904311461\\
0.0300000000000011	38.5343024992532\\
0.0400000000000063	25.1618042417604\\
0.0499999999999972	19.9442397689933\\
0.0600000000000023	0\\
0.0699999999999932	0\\
0.0799999999999983	0\\
0.0900000000000034	6.59165587971722\\
0.0999999999999943	0\\
0.109999999999999	0\\
0.120000000000005	0\\
0.129999999999995	0\\
0.140000000000001	0\\
0.150000000000006	0\\
0.159999999999997	0\\
0.170000000000002	0\\
0.180000000000007	0\\
0.189999999999998	0\\
0.200000000000003	0\\
0.209999999999994	2.39968137010854\\
0.219999999999999	0\\
0.230000000000004	0\\
0.239999999999995	0\\
0.25	0\\
0.260000000000005	0\\
0.269999999999996	0\\
0.280000000000001	0\\
0.290000000000006	0\\
0.299999999999997	0\\
0.310000000000002	0\\
0.319999999999993	0\\
0.329999999999998	0\\
0.340000000000003	0\\
0.349999999999994	3.72398685651697\\
0.359999999999999	12.9045106044011\\
0.370000000000005	0\\
0.379999999999995	0\\
0.390000000000001	0\\
0.400000000000006	0\\
0.409999999999997	4.73961963556707\\
0.420000000000002	0\\
0.430000000000007	10.6044010753759\\
0.439999999999998	18.2714328387932\\
0.450000000000003	3.83351588170865\\
0.459999999999994	0\\
0.469999999999999	18.6896345713432\\
0.480000000000004	32.3807627203027\\
0.489999999999995	29.4633077765608\\
0.5	38.1459723190282\\
0.510000000000005	56.9550930996714\\
0.519999999999996	34.7007866175446\\
0.530000000000001	38.0165289256198\\
0.540000000000006	41.4418002588868\\
0.549999999999997	56.8057353380464\\
0.560000000000002	76.072886587673\\
0.569999999999993	77.3274917853231\\
0.579999999999998	81.3004082445484\\
0.590000000000003	90.7896046997909\\
0.599999999999994	97.4808324205915\\
0.609999999999999	97.9986059942248\\
0.620000000000005	92.4823259982077\\
0.629999999999995	84.0087623220153\\
0.640000000000001	61.6648411829135\\
0.650000000000006	47.2468385940456\\
0.659999999999997	40.2668525341034\\
0.670000000000002	34.5813004082446\\
0.680000000000007	23.2599820770686\\
0.689999999999998	2.83779747087523\\
0.700000000000003	0\\
0.709999999999994	0\\
0.719999999999999	0\\
0.730000000000004	4.57034750572538\\
0.739999999999995	5.16777855222543\\
0.75	9.14069501145076\\
0.760000000000005	16.6782833814597\\
0.769999999999996	10.7338444687842\\
0.780000000000001	16.7878124066514\\
0.790000000000006	15.8120083640347\\
0.799999999999997	41.9197450960868\\
0.810000000000002	39.838693617445\\
0.819999999999993	34.3024992532112\\
0.829999999999998	43.1544359255203\\
0.840000000000003	59.2153738922633\\
0.849999999999994	43.0648212685453\\
0.859999999999999	59.47426067908\\
0.870000000000005	66.5339042118889\\
0.879999999999995	68.7543562680474\\
0.890000000000001	100\\
};

\addplot[area legend, draw=none, fill=mycolor2, fill opacity=0.1, forget plot]
table[row sep=crcr] {%
x	y\\
0.6	-20\\
2	-20\\
2	120\\
0.6	120\\
}--cycle;

\addplot[area legend, draw=none, fill=mycolor3, fill opacity=0.1, forget plot]
table[row sep=crcr] {%
x	y\\
0	-20\\
0.6	-20\\
0.6	120\\
0	120\\
}--cycle;
\end{axis}
\end{tikzpicture}%

%% file: figures/appendix/pago4smash.tex
%
%
\definecolor{mycolor1}{rgb}{0.00000,0.44700,0.74100}%
\definecolor{mycolor2}{rgb}{0.85000,0.32500,0.09800}%
\definecolor{mycolor3}{rgb}{0.46600,0.67400,0.18800}%
\begin{tikzpicture}

\begin{axis}[%
width=0.951\figwidth,
height=\figheight,
at={(0\figwidth,0\figheight)},
scale only axis,
xmin=0,
xmax=0.92,
ymin=-20,
ymax=120,
axis background/.style={fill=white},
legend style={legend cell align=left, align=left, fill=none, draw=none},
ylabel near ticks,
xlabel near ticks,
legend style={at={(.8,.7)},anchor=north west,legend cell align=left,align=left,fill=none,draw=none}
]
\addplot [color=mycolor1]
  table[row sep=crcr]{%
0.0100000000000051	51.1272727272727\\
0.019999999999996	50.0454545454545\\
0.0300000000000011	50.0636363636364\\
0.0400000000000063	50.2272727272727\\
0.0499999999999972	43.0181818181818\\
0.0600000000000023	34.7727272727273\\
0.0699999999999932	30\\
0.0799999999999983	30.5363636363636\\
0.0900000000000034	31.5636363636364\\
0.0999999999999943	30.0727272727273\\
0.109999999999999	33.2545454545455\\
0.120000000000005	36.8909090909091\\
0.129999999999995	28.3\\
0.140000000000001	22.2181818181818\\
0.150000000000006	19.2545454545455\\
0.159999999999997	14.2909090909091\\
0.170000000000002	8.7\\
0.180000000000007	4.93636363636364\\
0.189999999999998	2.8\\
0.200000000000003	-1.51818181818182\\
0.209999999999994	-5.43636363636364\\
0.219999999999999	-6.92727272727272\\
0.239999999999995	-11.7363636363636\\
0.25	-12.2272727272727\\
0.260000000000005	-11.8\\
0.269999999999996	-11.9090909090909\\
0.280000000000001	-5.43636363636364\\
0.290000000000006	-3.83636363636364\\
0.299999999999997	-0.472727272727269\\
0.310000000000002	2.60909090909091\\
0.319999999999993	2.14545454545454\\
0.329999999999998	1.86363636363636\\
0.340000000000003	2.11818181818182\\
0.349999999999994	2.48181818181818\\
0.359999999999999	2.53636363636363\\
0.370000000000005	2.82727272727273\\
0.379999999999995	2.60909090909091\\
0.390000000000001	1.80909090909091\\
0.400000000000006	1.75454545454545\\
0.409999999999997	0.809090909090912\\
0.420000000000002	-0.263636363636365\\
0.430000000000007	-0.518181818181816\\
0.439999999999998	-0.854545454545459\\
0.450000000000003	-0.690909090909088\\
0.459999999999994	-0.690909090909088\\
0.469999999999999	-0.481818181818184\\
0.480000000000004	-0.318181818181813\\
0.489999999999995	3.8\\
0.5	17.9272727272727\\
0.510000000000005	19.9181818181818\\
0.519999999999996	26.7909090909091\\
0.530000000000001	35.9090909090909\\
0.540000000000006	39.8363636363636\\
0.549999999999997	43.4727272727273\\
0.560000000000002	48.6545454545455\\
0.569999999999993	53.2727272727273\\
0.579999999999998	58.2272727272727\\
0.599999999999994	64.8272727272727\\
0.609999999999999	66.7909090909091\\
0.620000000000005	69.2090909090909\\
0.629999999999995	71.0818181818182\\
0.640000000000001	72.7181818181818\\
0.650000000000006	74.0545454545454\\
0.659999999999997	74.6545454545455\\
0.670000000000002	74.8727272727273\\
0.680000000000007	76.6181818181818\\
0.689999999999998	77.3272727272727\\
0.700000000000003	80.3363636363636\\
0.709999999999994	80.2727272727273\\
0.719999999999999	79.8363636363636\\
0.730000000000004	81.5\\
0.739999999999995	83.8727272727273\\
0.760000000000005	91.0090909090909\\
0.769999999999996	93.7909090909091\\
0.780000000000001	96.2909090909091\\
0.790000000000006	97.1181818181818\\
0.799999999999997	96.2636363636364\\
0.810000000000002	96.5090909090909\\
0.819999999999993	97.8636363636364\\
0.829999999999998	100.872727272727\\
0.840000000000003	104.354545454545\\
0.849999999999994	105.181818181818\\
0.859999999999999	104.154545454545\\
0.870000000000005	104.281818181818\\
0.879999999999995	105.209090909091\\
0.890000000000001	102.363636363636\\
0.900000000000006	86.7363636363636\\
0.909999999999997	73.2818181818182\\
0.920000000000002	61.2\\
};
\addlegendentry{$p_{4\textrm{a}}$}

\addplot [color=mycolor2, dashed, mark=o, mark options={solid, mycolor2}]
  table[row sep=crcr]{%
0.0100000000000051	50\\
0.019999999999996	49.1181818181818\\
0.0300000000000011	37.4272727272727\\
0.0400000000000063	31.3181818181818\\
0.0499999999999972	33.6\\
0.0600000000000023	40.8181818181818\\
0.0699999999999932	38.2818181818182\\
0.0799999999999983	40.1181818181818\\
0.0900000000000034	39.3727272727273\\
0.0999999999999943	30.6181818181818\\
0.109999999999999	21.5636363636364\\
0.120000000000005	10.9818181818182\\
0.129999999999995	0\\
0.140000000000001	0\\
0.150000000000006	0\\
0.159999999999997	0\\
0.170000000000002	0\\
0.180000000000007	0\\
0.189999999999998	0\\
0.200000000000003	0\\
0.209999999999994	0\\
0.219999999999999	0\\
0.230000000000004	0\\
0.239999999999995	0\\
0.25	0\\
0.260000000000005	0\\
0.269999999999996	0\\
0.280000000000001	0\\
0.290000000000006	0\\
0.299999999999997	0\\
0.310000000000002	0\\
0.319999999999993	0\\
0.329999999999998	0\\
0.340000000000003	0\\
0.349999999999994	0\\
0.359999999999999	0\\
0.370000000000005	0\\
0.379999999999995	0\\
0.390000000000001	0\\
0.400000000000006	0\\
0.409999999999997	0\\
0.420000000000002	0\\
0.430000000000007	0\\
0.439999999999998	0\\
0.450000000000003	0\\
0.459999999999994	0\\
0.469999999999999	15.4909090909091\\
0.480000000000004	20.0909090909091\\
0.489999999999995	20.4\\
0.5	22.1363636363636\\
0.510000000000005	38.3\\
0.519999999999996	47.0818181818182\\
0.530000000000001	57.4909090909091\\
0.540000000000006	75.0454545454545\\
0.549999999999997	83.7818181818182\\
0.560000000000002	89.7454545454545\\
0.569999999999993	100\\
0.579999999999998	100\\
0.590000000000003	100\\
0.599999999999994	100\\
0.609999999999999	100\\
0.620000000000005	100\\
0.629999999999995	100\\
0.640000000000001	93.6090909090909\\
0.650000000000006	99.0818181818182\\
0.659999999999997	100\\
0.670000000000002	100\\
0.680000000000007	100\\
0.689999999999998	100\\
0.700000000000003	100\\
0.709999999999994	100\\
0.719999999999999	100\\
0.730000000000004	100\\
0.739999999999995	100\\
0.75	100\\
0.760000000000005	100\\
0.769999999999996	100\\
0.780000000000001	88.8454545454545\\
0.790000000000006	96.5181818181818\\
0.799999999999997	100\\
0.810000000000002	97.7363636363636\\
0.819999999999993	95.7\\
0.829999999999998	94.3636363636364\\
0.840000000000003	91.9090909090909\\
0.849999999999994	100\\
0.859999999999999	96.8909090909091\\
0.870000000000005	87.6\\
0.879999999999995	82.1727272727273\\
0.890000000000001	71.6363636363636\\
0.900000000000006	64.9363636363636\\
0.909999999999997	44.7363636363636\\
0.920000000000002	22.6090909090909\\
};
\addlegendentry{$p^\textrm{des}_{4\textrm{a}}$}

\addplot[area legend, draw=none, fill=mycolor2, fill opacity=0.1, forget plot]
table[row sep=crcr] {%
x	y\\
0.6	-20\\
2	-20\\
2	120\\
0.6	120\\
}--cycle;

\addplot[area legend, draw=none, fill=mycolor3, fill opacity=0.1, forget plot]
table[row sep=crcr] {%
x	y\\
0	-20\\
0.6	-20\\
0.6	120\\
0	120\\
}--cycle;
\end{axis}
\end{tikzpicture}%

%% file: figures/appendix/pantago4.tex
%
%
\definecolor{mycolor1}{rgb}{0.00000,0.44700,0.74100}%
\definecolor{mycolor2}{rgb}{0.85000,0.32500,0.09800}%
\definecolor{mycolor3}{rgb}{0.46600,0.67400,0.18800}%
\begin{tikzpicture}

\begin{axis}[%
width=0.951\figwidth,
height=\figheight,
at={(0\figwidth,0\figheight)},
scale only axis,
xmin=0,
xmax=0.89,
xlabel style={font=\color{white!15!black}},
xlabel={$t~[\textrm{s}]$},
ymin=-20,
ymax=120,
ylabel style={font=\color{white!15!black}},
ylabel={$[\%]$},
axis background/.style={fill=white},
ylabel near ticks,
xlabel near ticks,
legend style={at={(.8,.9)},anchor=north west,legend cell align=left,align=left,fill=none,draw=none}
]
\addplot [color=mycolor1, forget plot]
  table[row sep=crcr]{%
0.0100000000000051	49.9272727272727\\
0.019999999999996	50\\
0.0300000000000011	49.9181818181818\\
0.0400000000000063	44.0545454545455\\
0.0499999999999972	36.6818181818182\\
0.0600000000000023	35.3545454545455\\
0.0699999999999932	34.7363636363636\\
0.0799999999999983	31.4363636363636\\
0.0900000000000034	32.1636363636364\\
0.0999999999999943	31.6727272727273\\
0.109999999999999	23.9636363636364\\
0.120000000000005	19.8727272727273\\
0.129999999999995	17\\
0.140000000000001	11.9818181818182\\
0.150000000000006	6.15454545454546\\
0.159999999999997	4.10909090909091\\
0.170000000000002	1.15454545454546\\
0.180000000000007	-2\\
0.189999999999998	-4.32727272727273\\
0.200000000000003	-4.82727272727273\\
0.209999999999994	-6.90000000000001\\
0.219999999999999	-7.58181818181818\\
0.230000000000004	-6.83636363636364\\
0.239999999999995	-6.51818181818182\\
0.25	-3.88181818181818\\
0.260000000000005	-2.57272727272728\\
0.280000000000001	1.77272727272727\\
0.290000000000006	2.19090909090909\\
0.299999999999997	1.77272727272727\\
0.310000000000002	1.74545454545455\\
0.319999999999993	1.59090909090909\\
0.329999999999998	1.84545454545454\\
0.340000000000003	1.91818181818182\\
0.349999999999994	2.00909090909092\\
0.359999999999999	4.18181818181819\\
0.370000000000005	2.71818181818182\\
0.379999999999995	3.23636363636363\\
0.390000000000001	4.28181818181818\\
0.400000000000006	10.4636363636364\\
0.409999999999997	13.8909090909091\\
0.420000000000002	19.8363636363636\\
0.430000000000007	26.0272727272727\\
0.439999999999998	32.7909090909091\\
0.450000000000003	37.9272727272727\\
0.459999999999994	43.5363636363636\\
0.469999999999999	50.9090909090909\\
0.480000000000004	55.4\\
0.489999999999995	59.4909090909091\\
0.5	61.4272727272727\\
0.510000000000005	61.2454545454545\\
0.519999999999996	58.7272727272727\\
0.530000000000001	56.4727272727273\\
0.540000000000006	55.0818181818182\\
0.549999999999997	54.3\\
0.560000000000002	53.8818181818182\\
0.569999999999993	54.0545454545455\\
0.579999999999998	54.4181818181818\\
0.590000000000003	53.8636363636364\\
0.599999999999994	52.7636363636364\\
0.609999999999999	52.0181818181818\\
0.620000000000005	51.7818181818182\\
0.629999999999995	52.2727272727273\\
0.650000000000006	53.8636363636364\\
0.659999999999997	55.1\\
0.670000000000002	56.0909090909091\\
0.680000000000007	57.4181818181818\\
0.689999999999998	57.8636363636364\\
0.700000000000003	58.9636363636364\\
0.709999999999994	59.2\\
0.719999999999999	59.4545454545455\\
0.730000000000004	60.7272727272727\\
0.739999999999995	61.9181818181818\\
0.75	63.2818181818182\\
0.760000000000005	64.9909090909091\\
0.769999999999996	66.1181818181818\\
0.780000000000001	66.9818181818182\\
0.790000000000006	61.1090909090909\\
0.810000000000002	39.0363636363636\\
0.819999999999993	31.4545454545455\\
0.829999999999998	25.7090909090909\\
0.840000000000003	22.0909090909091\\
0.849999999999994	16.1818181818182\\
0.859999999999999	11.2272727272727\\
0.870000000000005	7.25454545454545\\
0.879999999999995	3.80909090909091\\
0.890000000000001	-0.618181818181824\\
};
\addplot [color=mycolor2, dashed, mark=o, mark options={solid, mycolor2}, forget plot]
  table[row sep=crcr]{%
0.0100000000000051	50\\
0.019999999999996	37.3818181818182\\
0.0300000000000011	37.8090909090909\\
0.0400000000000063	44.1909090909091\\
0.0499999999999972	33.9272727272727\\
0.0600000000000023	36.7363636363636\\
0.0699999999999932	35.5909090909091\\
0.0799999999999983	31.6272727272727\\
0.0900000000000034	13.8909090909091\\
0.0999999999999943	16.2363636363636\\
0.109999999999999	11.3545454545455\\
0.120000000000005	0.900000000000006\\
0.129999999999995	0\\
0.140000000000001	9.38181818181818\\
0.150000000000006	7.57272727272728\\
0.159999999999997	12.1454545454545\\
0.170000000000002	0\\
0.180000000000007	0\\
0.189999999999998	0\\
0.200000000000003	6.47272727272727\\
0.209999999999994	0\\
0.219999999999999	0\\
0.230000000000004	0\\
0.239999999999995	0\\
0.25	0\\
0.260000000000005	0\\
0.269999999999996	0\\
0.280000000000001	0\\
0.290000000000006	0\\
0.299999999999997	0\\
0.310000000000002	0\\
0.319999999999993	0\\
0.329999999999998	0\\
0.340000000000003	8.19090909090909\\
0.349999999999994	0\\
0.359999999999999	6.24545454545455\\
0.370000000000005	2.33636363636364\\
0.379999999999995	14.1363636363636\\
0.390000000000001	17.2545454545455\\
0.400000000000006	20.5818181818182\\
0.409999999999997	28.8454545454545\\
0.420000000000002	46.0818181818182\\
0.430000000000007	63.9454545454546\\
0.439999999999998	62.7818181818182\\
0.450000000000003	80.3\\
0.459999999999994	81.9272727272727\\
0.469999999999999	89.7363636363636\\
0.480000000000004	100\\
0.489999999999995	100\\
0.5	100\\
0.510000000000005	100\\
0.519999999999996	100\\
0.530000000000001	100\\
0.540000000000006	100\\
0.549999999999997	100\\
0.560000000000002	88.0363636363636\\
0.569999999999993	75.7\\
0.579999999999998	69.1272727272727\\
0.590000000000003	68.4818181818182\\
0.599999999999994	65.1454545454545\\
0.609999999999999	66.2636363636364\\
0.620000000000005	58.6272727272727\\
0.629999999999995	56.3363636363636\\
0.640000000000001	63.4181818181818\\
0.650000000000006	64.2363636363636\\
0.659999999999997	58.5909090909091\\
0.670000000000002	60.5636363636364\\
0.680000000000007	60.6363636363636\\
0.689999999999998	67.7636363636364\\
0.700000000000003	52.0909090909091\\
0.709999999999994	50.3363636363636\\
0.719999999999999	70.9272727272727\\
0.730000000000004	72.7090909090909\\
0.739999999999995	61.9090909090909\\
0.75	62.9909090909091\\
0.760000000000005	53.3636363636364\\
0.769999999999996	40.4090909090909\\
0.780000000000001	29.7363636363636\\
0.790000000000006	5.61818181818182\\
0.799999999999997	0\\
0.810000000000002	0\\
0.819999999999993	0\\
0.829999999999998	0\\
0.840000000000003	0\\
0.849999999999994	0\\
0.859999999999999	0\\
0.870000000000005	0\\
0.879999999999995	0\\
0.890000000000001	20.6363636363636\\
};

\addplot[area legend, draw=none, fill=mycolor2, fill opacity=0.1, forget plot]
table[row sep=crcr] {%
x	y\\
0.6	-20\\
2	-20\\
2	120\\
0.6	120\\
}--cycle;

\addplot[area legend, draw=none, fill=mycolor3, fill opacity=0.1, forget plot]
table[row sep=crcr] {%
x	y\\
0	-20\\
0.6	-20\\
0.6	120\\
0	120\\
}--cycle;
\end{axis}
\end{tikzpicture}%

%% file: figures/appendix/pantago4smash.tex
%
%
\definecolor{mycolor1}{rgb}{0.00000,0.44700,0.74100}%
\definecolor{mycolor2}{rgb}{0.85000,0.32500,0.09800}%
\definecolor{mycolor3}{rgb}{0.46600,0.67400,0.18800}%
\begin{tikzpicture}

\begin{axis}[%
width=0.951\figwidth,
height=\figheight,
at={(0\figwidth,0\figheight)},
scale only axis,
unbounded coords=jump,
xmin=0,
xmax=0.92,
xlabel style={font=\color{white!15!black}},
xlabel={$t~[\textrm{s}]$},
ymin=-20,
ymax=120,
axis background/.style={fill=white},
legend style={legend cell align=left, align=left, fill=none, draw=none},
ylabel near ticks,
xlabel near ticks,
legend style={at={(.8,1)},anchor=north west,legend cell align=left,align=left,fill=none,draw=none}
]
\addplot [color=mycolor1]
  table[row sep=crcr]{%
0.0100000000000051	71.4\\
0.019999999999996	100\\
0.0300000000000011	100.109090909091\\
0.0400000000000063	99.4363636363636\\
0.0600000000000023	84.9272727272727\\
0.0699999999999932	68.0909090909091\\
0.0799999999999983	55.5636363636364\\
0.0999999999999943	38.6909090909091\\
0.109999999999999	26.3090909090909\\
0.120000000000005	18.5090909090909\\
0.129999999999995	11.9818181818182\\
0.140000000000001	4.83636363636364\\
0.150000000000006	-3.47272727272727\\
0.159999999999997	-8.36363636363636\\
0.170000000000002	-14.1272727272727\\
0.180000000000007	-19.6545454545455\\
0.189999999999998	-22.3090909090909\\
nan	nan\\
0.219999999999999	-23.1636363636364\\
0.230000000000004	-18.4363636363636\\
0.239999999999995	-1.90909090909091\\
0.25	4.09090909090909\\
0.260000000000005	3.01818181818182\\
0.269999999999996	8.27272727272728\\
0.280000000000001	2.96363636363637\\
0.290000000000006	2.69090909090909\\
0.299999999999997	6.49090909090908\\
0.310000000000002	4.07272727272728\\
0.319999999999993	3.92727272727272\\
0.329999999999998	6.21818181818182\\
0.340000000000003	2.23636363636363\\
0.349999999999994	0.709090909090904\\
0.359999999999999	0.654545454545456\\
0.370000000000005	-1.52727272727273\\
0.379999999999995	-2.45454545454545\\
0.390000000000001	-1.18181818181819\\
0.400000000000006	-2.2909090909091\\
0.409999999999997	-1.54545454545455\\
0.420000000000002	-1.07272727272728\\
0.430000000000007	-1.23636363636363\\
0.439999999999998	-0.890909090909091\\
0.450000000000003	-0.309090909090912\\
0.459999999999994	-0.599999999999994\\
0.469999999999999	-0.25454545454545\\
0.480000000000004	0.490909090909085\\
0.489999999999995	0.109090909090909\\
0.5	0.436363636363637\\
0.510000000000005	1.10909090909091\\
0.519999999999996	0.618181818181824\\
0.530000000000001	1.01818181818182\\
0.540000000000006	1.12727272727273\\
0.549999999999997	0.490909090909085\\
0.560000000000002	1.01818181818182\\
0.569999999999993	0.799999999999997\\
0.579999999999998	0.654545454545456\\
0.590000000000003	1.01818181818182\\
0.599999999999994	0.672727272727272\\
0.609999999999999	0.618181818181824\\
0.620000000000005	0.836363636363643\\
0.629999999999995	1.07272727272728\\
0.640000000000001	0.854545454545459\\
0.650000000000006	1.30909090909091\\
0.659999999999997	0.818181818181813\\
0.670000000000002	1.32727272727273\\
0.680000000000007	0.690909090909088\\
0.689999999999998	0.909090909090907\\
0.700000000000003	0.890909090909091\\
0.709999999999994	0.654545454545456\\
0.719999999999999	-0.0363636363636317\\
0.730000000000004	-0.327272727272728\\
0.739999999999995	-0.454545454545453\\
0.75	-0.563636363636363\\
0.760000000000005	-0.63636363636364\\
0.769999999999996	-0.400000000000006\\
0.780000000000001	-0.36363636363636\\
0.790000000000006	-0.290909090909096\\
0.799999999999997	-0.181818181818187\\
0.810000000000002	-0.272727272727266\\
0.819999999999993	-0.145454545454541\\
0.829999999999998	-0.290909090909096\\
0.840000000000003	0.0545454545454476\\
0.849999999999994	0.0909090909090935\\
0.859999999999999	-0.0727272727272776\\
0.870000000000005	-0.0181818181818159\\
0.879999999999995	0.418181818181822\\
0.890000000000001	2.47272727272727\\
0.900000000000006	15.8\\
0.909999999999997	9.98181818181818\\
0.920000000000002	17.2545454545455\\
};
\addlegendentry{$p_{4\textrm{b}}$}

\addplot [color=mycolor2, dashed, mark=o, mark options={solid, mycolor2}]
  table[row sep=crcr]{%
0.0100000000000051	100\\
0.019999999999996	97.6727272727273\\
0.0300000000000011	87.9818181818182\\
0.0400000000000063	90.6363636363636\\
0.0499999999999972	52.6\\
0.0600000000000023	30.9090909090909\\
0.0699999999999932	19.3272727272727\\
0.0799999999999983	0\\
0.0900000000000034	0\\
0.0999999999999943	0\\
0.109999999999999	0\\
0.120000000000005	0\\
0.129999999999995	0\\
0.140000000000001	0\\
0.150000000000006	0\\
0.159999999999997	0\\
0.170000000000002	0\\
0.180000000000007	0\\
0.189999999999998	0\\
0.200000000000003	0\\
0.209999999999994	0\\
0.219999999999999	0\\
0.230000000000004	0\\
0.239999999999995	0\\
0.25	0\\
0.260000000000005	0\\
0.269999999999996	0\\
0.280000000000001	0\\
0.290000000000006	0\\
0.299999999999997	0\\
0.310000000000002	0\\
0.319999999999993	0\\
0.329999999999998	0\\
0.340000000000003	0\\
0.349999999999994	0\\
0.359999999999999	0\\
0.370000000000005	0\\
0.379999999999995	0\\
0.390000000000001	0\\
0.400000000000006	0\\
0.409999999999997	0\\
0.420000000000002	0\\
0.430000000000007	0\\
0.439999999999998	0\\
0.450000000000003	0\\
0.459999999999994	0\\
0.469999999999999	0\\
0.480000000000004	0\\
0.489999999999995	1.61818181818182\\
0.5	0\\
0.510000000000005	0\\
0.519999999999996	0\\
0.530000000000001	0\\
0.540000000000006	0\\
0.549999999999997	0\\
0.560000000000002	0\\
0.569999999999993	2.40000000000001\\
0.579999999999998	0\\
0.590000000000003	2.23636363636363\\
0.599999999999994	5.81818181818181\\
0.609999999999999	0\\
0.620000000000005	0\\
0.629999999999995	0\\
0.640000000000001	0\\
0.650000000000006	0\\
0.659999999999997	0\\
0.670000000000002	0\\
0.680000000000007	0\\
0.689999999999998	0\\
0.700000000000003	0\\
0.709999999999994	0\\
0.719999999999999	0\\
0.730000000000004	0\\
0.739999999999995	0\\
0.75	0\\
0.760000000000005	0\\
0.769999999999996	0\\
0.780000000000001	0\\
0.790000000000006	0\\
0.799999999999997	0\\
0.810000000000002	0\\
0.819999999999993	0\\
0.829999999999998	0\\
0.840000000000003	0\\
0.849999999999994	0\\
0.859999999999999	3.61818181818182\\
0.870000000000005	24.4909090909091\\
0.879999999999995	8.36363636363636\\
0.890000000000001	0\\
0.900000000000006	28.6727272727273\\
0.909999999999997	38.6363636363636\\
0.920000000000002	94.7090909090909\\
};
\addlegendentry{$p^\textrm{des}_{4\textrm{b}}$}

\addplot[area legend, draw=none, fill=mycolor2, fill opacity=0.1, forget plot]
table[row sep=crcr] {%
x	y\\
0.6	-20\\
2	-20\\
2	120\\
0.6	120\\
}--cycle;

\addplot[area legend, draw=none, fill=mycolor3, fill opacity=0.1, forget plot]
table[row sep=crcr] {%
x	y\\
0	-20\\
0.6	-20\\
0.6	120\\
0	120\\
}--cycle;
\end{axis}
\end{tikzpicture}%